\RequirePackage{fix-cm}
\documentclass[twocolumn]{svjour3}          
\smartqed  
\usepackage{graphicx}
\usepackage{epstopdf}
\usepackage{pifont}
\usepackage{subfigure}
\usepackage{algorithmic}
\usepackage{algorithm}
\usepackage{bm}
\usepackage{booktabs}
\usepackage{multirow}
\usepackage{colortbl}
\usepackage{url}
\usepackage{amsmath}
\usepackage[colorlinks,linkcolor=blue]{hyperref}
\definecolor{mygray}{gray}{.9}
\definecolor{newcolor}{rgb}{.8,.349,.1}
\begin{document}
\title{Superpixel-guided Two-view Deterministic Geometric Model Fitting}
\author{Guobao Xiao\and
            Hanzi Wang\and
            Yan Yan\and
        David Suter
}
\institute{\ding{41}
 Hanzi Wang \at
  \email{hanzi.wang@xmu.edu.cn}  \and
 Guobao Xiao
 \and
  Hanzi Wang
  \and
  Yan Yan \at
  Fujian Key Laboratory of Sensing and Computing for Smart City, School of Information Science and Engineering, Xiamen University, China
  \and
  David Suter\at
  School of Science, Edith Cowan University, Australia
}
\date{Received: date / Accepted: date}

\maketitle

\begin{abstract}
Geometric model fitting is a fundamental research topic in computer vision and it aims to fit and segment multiple-structure data. In this paper, we propose a novel superpixel-guided two-view geometric model fitting method (called SDF), which can obtain reliable and consistent results for real images. Specifically, SDF includes three main parts: a deterministic sampling algorithm, a model hypothesis updating strategy and a novel model selection algorithm. The proposed deterministic sampling algorithm generates a set of initial model hypotheses according to the prior information of superpixels. Then the proposed updating strategy further improves the quality of model hypotheses. After that, by analyzing the properties of the updated model hypotheses, the proposed model selection algorithm extends the conventional ``fit-and-remove" framework to estimate model instances in multiple-structure data. The three parts are tightly coupled to boost the performance of SDF in both speed and accuracy, and SDF has the deterministic nature. Experimental results show that the proposed SDF has significant advantages over several state-of-the-art fitting methods when it is applied to real images with single-structure and multiple-structure data.
\keywords{Model fitting \and superpixel \and deterministic algorithm \and multiple-structure data}
\end{abstract}

\section{Introduction}
\label{intro}
\begin{figure}
\centering
\subfigure{
\begin{minipage}{.48\textwidth}
\centerline{\includegraphics[width=0.9\textwidth]{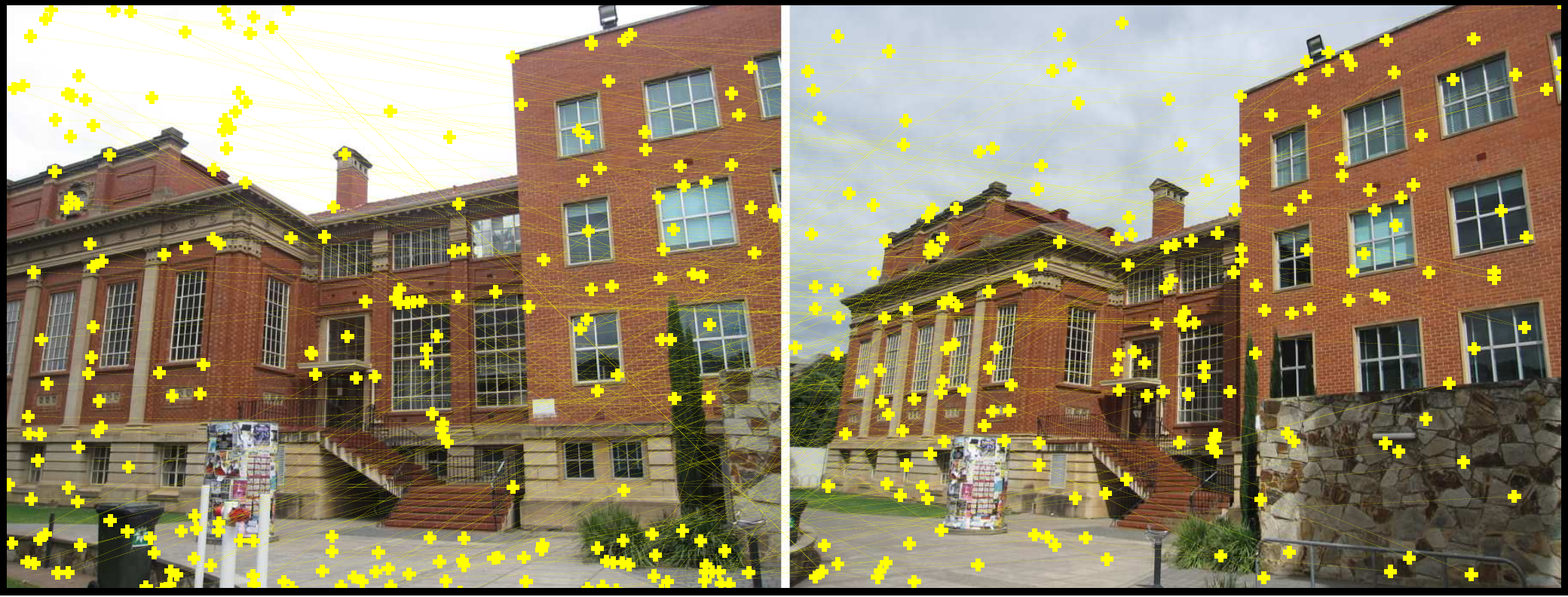}}
  \centerline{(a) The input data}
\end{minipage}
}
\subfigure{
\begin{minipage}{.48\textwidth}
\centerline{\includegraphics[width=0.9\textwidth]{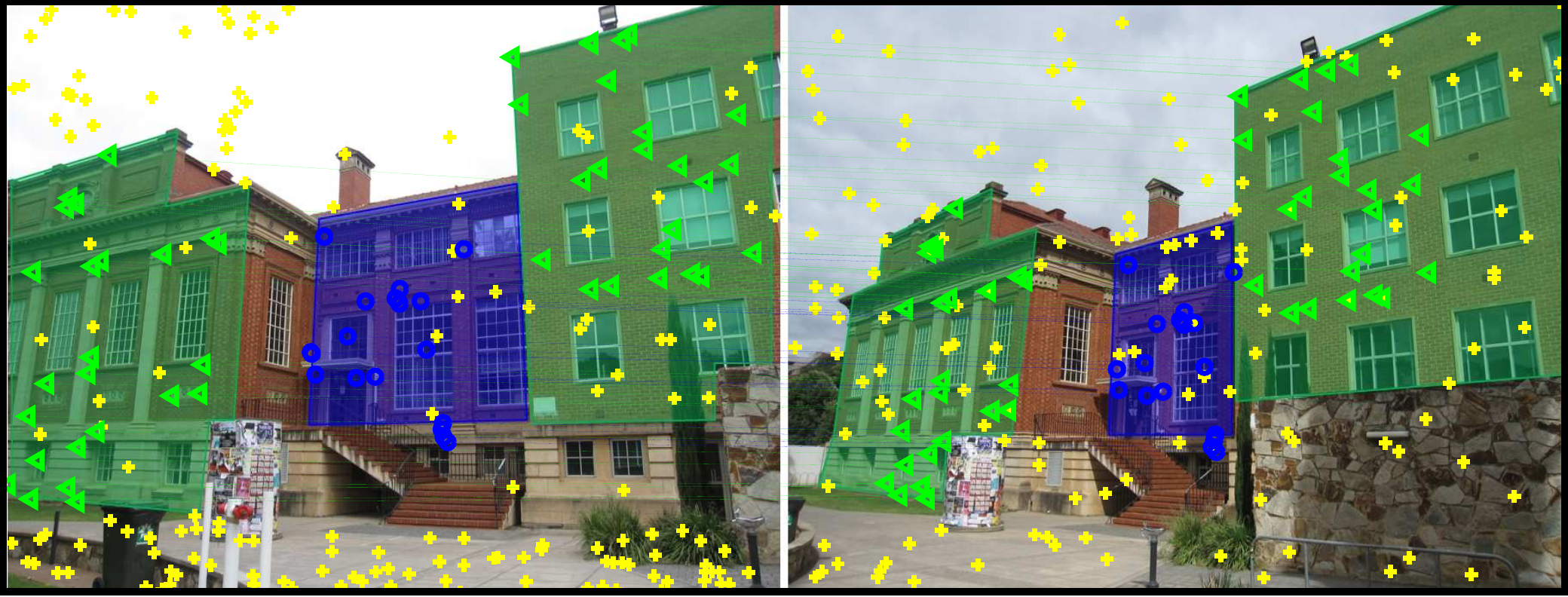}}
  \centerline{(b) The obtained fitting and segmentation results}
\end{minipage}
}
\caption{An example of the fitting and segmentation problem for multiple-structure data. }
\label{fig:showproblems}
\end{figure}
This paper addresses the challenging problem of fitting multiple structures in data {in} the presence of severe outliers. The goal of geometric model fitting is to select a small subset of model hypotheses that best explains input data \cite{pham2014interacting}, and it has been widely used in many applications, such as motion segmentation \cite{Poling2014,Magri_2016_CVPR}, 3D reconstruction \cite{isack2014energy,Woodford2014} and homography/fundamental matrix estimation \cite{Wang_2015_ICCV,tennakoon2016robust}.

To illustrate the problem addressed in this paper, we show a standard homography matrix estimation problem in Fig. \ref{fig:showproblems}. As shown in Fig. \ref{fig:showproblems}(a), we deal with the data (i.e., the keypoint correspondences in two-view images), which include two homographies. A model fitting method aims to estimate the parameters of model instances (also called ``structures", which are the homographies in this example) and identify inliers (i.e., the points marked in blue and green in Fig. \ref{fig:showproblems}(b)) and outliers (i.e., the points marked in yellow in Fig. \ref{fig:showproblems}(b)). Of course, model instances are not restricted to homographies and they may include fundamental matrices or other geometric models. 

A model fitting method generally includes two steps, i.e., sampling a number of minimum subsets to generate some model hypothesis candidates and estimating model instances in data. A minimum subset is the minimum number of data points required to generate a model hypothesis (e.g., $4$ data points for the homography matrix estimation in Fig. \ref{fig:showproblems}). To sample an effective minimum subset, whose elements are all the inliers of a model instance, a model fitting method often increases the number of sampled subsets. As one of the most popular fitting methods, RANdom SAmple Consensus (RANSAC) \cite{fischler1981random} is simple and effective for fitting single-structure data. Some guided-sampling methods (e.g., Multi-GS \cite{chin2012accelerated,Tran2014}, LO-RANSAC \cite{chum2003locally}, PROSAC \cite{chum2005matching} and Proximity \cite{kanazawa2004detection}) have been proposed to improve RANSAC. However, these guided-sampling methods cannot guarantee the consistency of their solutions due to their randomness. Thus, the fitting results obtained by some fitting methods (e.g.,  \cite{xiao2016HF,Wang_2015_ICCV,mittal2012generalized,tennakoon2016robust} may vary if they are not provided a sufficient number of sampled minimum subsets.

Recently, some researchers have proposed deterministic fitting methods (e.g., \cite{lee2013deterministic,li2009consensus,litman2015inverting,fredriksson2015practical,chin2015efficient}), which are able to provide consistent solutions. The fitting results obtained by deterministic fitting methods are also more reliable than those obtained by the previous guided-sampling based fitting methods. However, these deterministic fitting methods also suffer from some major problems, e.g., most of them only work for single-structure data, and their computational efficiency is very low, especially for data with few inliers.

In this paper, we focus on a deterministic fitting method, which can efficiently sample effective minimum subsets and estimate model instances for multiple-structure data. Note that feature appearances contain important prior information, and some methods (e.g., \cite{isack2014energy,chum2005matching,serradell2010combining}) have been proposed to introduce ``feature appearances" to model fitting. However, few deterministic fitting methods fully take advantage of ``feature appearances" of keypoint correspondences.
Thus, we propose to utilize the ``feature appearances" to derive more effective sampled minimum subsets and reduce the computational cost of the deterministic fitting method. Specifically, we firstly analyze the relationship between keypoint correspondences and superpixels (that characterize the prior information of feature appearances), and obtain the grouping cues of keypoint correspondences. Then, we propose a deterministic sampling algorithm to initialize a set of model hypotheses according to the grouping cues. After that, we further improve the quality of model hypotheses using an effective updating strategy. Finally, we propose a novel model selection algorithm to estimate all model instances in data.

Overall, the main contributions of this paper are summarized as follows:
\begin{itemize}
\item We firstly utilize the information of superpixels to deal with the deterministic model fitting problem. Superpixels involve spatial homogeneity, and they can provide powerful grouping cues of keypoint correspondences. This will largely reduce the number of all combinations of keypoint correspondences (a minimal subset is a combination). Note that sampling all the minimal subset is usually an \emph{NP-hard} problem. However, we provide a consistent solution for deterministically fitting multiple-structure data within a reasonable time by utilizing the information of superpixels.

\item We introduce an effective model hypothesis updating strategy to improve the quality of model hypotheses. The updating strategy is an important part of the proposed fitting method, especially for dealing with multiple-structure data. This is because the updating strategy is able to effectively increase the probability of generating an effective model hypothesis closer to the true model instance.

\item We propose a novel model selection algorithm, which extends the conventional ``fit-and-remove'' framework by sequentially removing redundant model hypotheses instead of data points. Compared to the conventional framework, the proposed algorithm is more efficient since it does not need to generate a number of new model hypotheses at each iteration.
\end{itemize}

The proposed SDF is able to provide consistent solutions efficiently and deterministically, for both single-structure and multiple-structure data. It is worth pointing out that, most conventional model fitting methods are based on randomness, and most existing deterministic fitting methods suffer from high computational complexity. Therefore, the proposed SDF method is a good alternative for solving the model fitting problem. Experimental results show that SDF achieves substantial improvements over several state-of-the-art fitting methods.

This paper is an extension of our previous work in \cite{ECCVXiao2016}. We have made several crucial additions:
\begin{itemize}
\item A novel model hypothesis updating strategy is employed to improve the quality of model hypotheses generated by the original proposed method (Sec. \ref{sec:hypothesesupdating});
\item The weighing scores of model hypotheses are utilized to select ``significant" model hypotheses in the model selection algorithm (Sec. \ref{sec:modelselection});
\item A more effective measurement of fitting errors is used to evaluate the performance of all the competing fitting methods (Eq. (\ref{equ:fittingerror}));
\item More data and more competing fitting methods are added to evaluate the performance of the proposed fitting method (Sec. \ref{sec:experiments});
\item More analyses and discussions are added (Sec. \ref{sec:limitations}).
\end{itemize}

The rest of the paper is organized as follows: In Sec. \ref{sec:relatework} we review the related literatures. In Sec. \ref{sec:samplingalgorithm} and Sec. \ref{sec:modelselection}, we propose the deterministic sampling algorithm and the model selection algorithm, respectively. We summarize the complete SDF method in Sec. \ref{sec:completealgorithm}. In Sec. \ref{sec:experiments}, we present the experimental results on both single-structure and multiple-structure data. We further discuss and analyze the proposed SDF method in Sec. \ref{sec:limitations}, and draw conclusions in Sec. \ref{sec:conclusion}.
\section{Related work}
\label{sec:relatework}
In this section, we will introduce the model fitting methods highly related to this paper, including random sampling based fitting methods, matching score based fitting methods and deterministic sampling based fitting methods.
\subsection{Random Sampling Based Fitting Methods}
As one of the most popular fitting methods, RANSAC has been used in a wide range of applications and improved by many fitting methods. The original RANSAC algorithm is proposed by Fischler and Bolles \cite{fischler1981random} as a general framework for the task of model fitting. RANSAC consists of two steps, i.e., randomly sampling some minimal subsets of data points to generate a number of model hypothesis candidates, and selecting a model hypothesis with the largest number of inliers from the model hypothesis candidates
as the estimated model instance. Many model fitting methods, e.g., \cite{mittal2012generalized,isack2014energy,tennakoon2016robust,Magri_2016_CVPR,xiao2016HF}, employ the random sampling strategy (due to its simplicity) to sample minimal subsets, and they mainly focus on model selection.

Some works, e.g., \cite{Magri_2014_CVPR,Tordoff2005,Lai2017152}, were proposed to guide the minimal subset sampling process.
These model fitting methods have achieved better performance than RANSAC on either speed or accuracy.  However, they cannot achieve consistent and {tractable} fitting results. In contrast, the proposed method in this paper focuses on the deterministic fitting, and it can obtain more consistent results.
\subsection{Matching Score Based Fitting Methods}
Generally, a good keypoint correspondence between an image pair has a high matching score, while a bad keypoint correspondence has a low matching score. Based on this observation, some fitting methods, e.g., \cite{Fragoso_2013_ICCV,Tordoff2005,Fragoso_2013_CVPR,chum2005matching,Lai2017152}, were proposed to guide sampling minimal subsets, by which they can effectively accelerate the process of promising model hypothesis generation. Most of these fitting methods can work well on single-structure data, but they may fail on multiple-structure data. This is because the keypoint correspondences with high matching scores may belong to different model instances in multiple-structure data, and thus it is hard to distinguish them from the global perspective.

In contrast, the proposed fitting method in this paper utilizes the information of matching scores from the local perspective (i.e., the information derived from superpixels), and it can work well on multiple-structure data.
\subsection{Deterministic Sampling Based Fitting Methods}
Deterministic sampling based fitting methods (e.g., \cite{lee2013deterministic,li2009consensus,litman2015inverting,fredriksson2015practical,Enqvist2015,chin2015efficient}) are able to provide consistent and reliable fitting results. For example, Li \cite{li2009consensus} used a tailored branch-and-bound scheme to deterministically solve the global optimization problem for model fitting. Lee et al. \cite{lee2013deterministic} deterministically generated model hypotheses based on the maximum feasible subsystem framework for model fitting. Litman et al. \cite{litman2015inverting} employed a method for inlier rate estimation to compute a globally optimal transformation for model fitting. Fredriksson et al. \cite{fredriksson2015practical} extended a branch-and-bound scheme to solve the two-view translation estimation problem. Enqvist et al. \cite{Enqvist2015} proposed to use loss functions to perform model estimation. Chin et al. \cite{chin2015efficient} used the Astar search algorithm \cite{hart1968formal} to provide a globally optimal solution for the model fitting problem.

However, most of these deterministic fitting methods assume that there only exists a single structure in data, and they cannot provide solutions for multiple-structure data within a reasonable time (most of them does not be applied sequentially for multiple-structure data since they cannot handle high outlier percentages). Note that IMaxFS-ISE \cite{lee2013deterministic} and MS \cite{Enqvist2015} can work for multiple-structure data, but they require to generate a number of model hypotheses repeatedly, which is computationally expensive. Compared to these deterministic sampling based fitting methods, the proposed SDF fitting method not only reserves the deterministic nature but also efficiently provides good solutions for fitting both single-structure and multiple-structure data.
\section{The proposed deterministic sampling algorithm}
\label{sec:samplingalgorithm}
In this section, we first introduce superpixels to model fitting (see Sec. \ref{sec:superpixel}), and then deterministically generate an initial model hypothesis set based on superpixels (see Sec. \ref{sec:hypothesesgeneration}). We also introduce a model hypothesis updating strategy to improve the quality of generated model hypotheses (see Sec. \ref{sec:hypothesesupdating}).
\subsection{Introducing Superpixels to Model Fitting}
\label{sec:superpixel}
We aim to accelerate the subset sampling process by analyzing the relationship between keypoint correspondences and superpixels. Note that superpixels obtained by superpixel segmentation methods (e.g.,  \cite{achanta2012slic,shen2014lazy}) adhere well to the object boundaries in an image. That is, the feature points in a superpixel usually belong to the same structure. For the two-view fitting problem, each keypoint correspondence consists of two feature points in two images. Thus, the keypoint correspondences (here we only consider inliers) have a high possibility of belonging to the inliers of the same structure, if their corresponding feature points come from the same superpixel.

\begin{figure}[h]
\centering
\centerline{\includegraphics[width=0.48\textwidth]{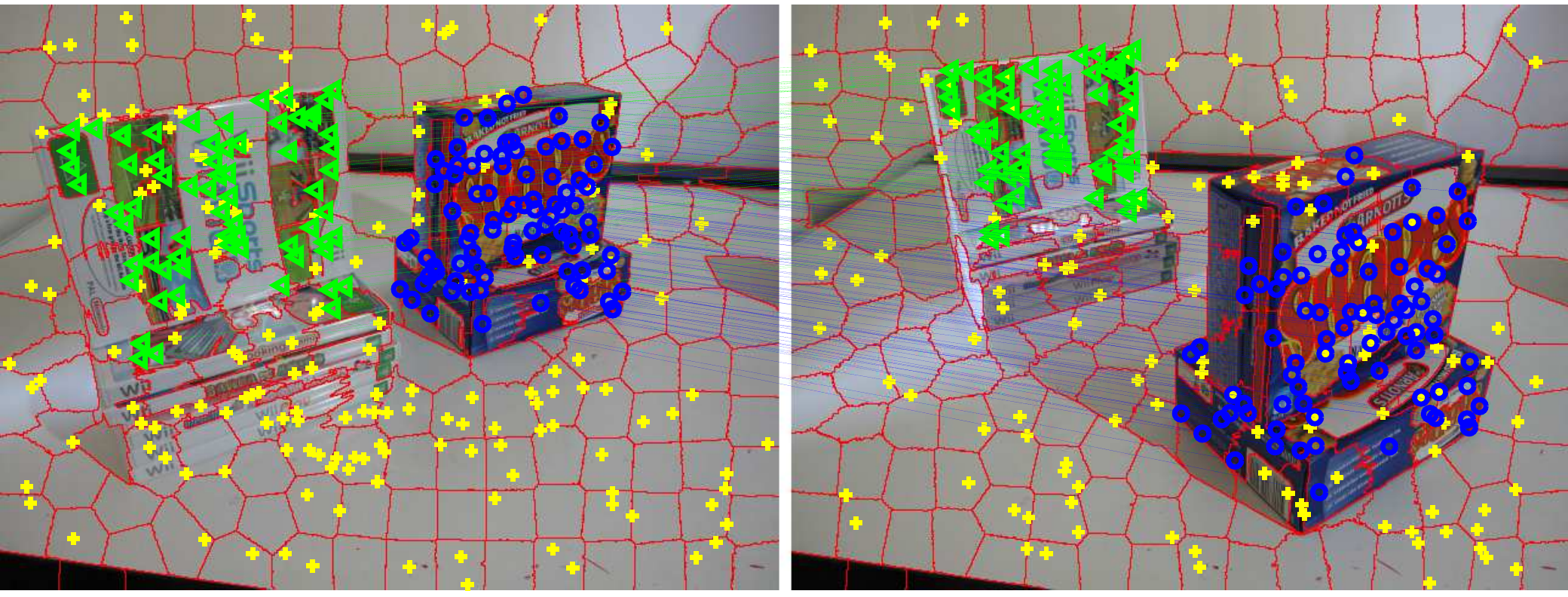}}
\caption{An example showing superpixels and keypoint correspondences based on the ground truth result of fundamental matrix for an image pair (``Gamebiscuit").}
\label{fig:supixel}
\end{figure}
To illustrate the the relationship between keypoint correspondences and superpixels more clearly, we show an example, i.e., the image pair ``Gamebiscuit" (from the AdelaideRMF dataset \cite{wong2011dynamic}), in Fig. \ref{fig:supixel}. We show both the segmented superpixels and the keypoint correspondences labeled according to the ground truth of fundamental matrix. From the figure, we can see that, the keypoint correspondences (marked by green or blue color) derived from the feature points in each superpixel, are the inliers of the same structure.

Based on the above analysis, we divide keypoint correspondences into a group set ${\bm{\mathcal{G}}}$, i.e., each group consists of the keypoint correspondences associated to the feature points in the same superpixel. This reduces the required number of sampled minimal subsets (i.e., the subset sampling process is accelerated).
\subsection{Initial Model Hypothesis Generation}
\label{sec:hypothesesgeneration}
In this subsection, we propose a deterministic sampling method to generate initial model hypotheses. We firstly combine some groups of the group set $\bm{\mathcal{G}}$. This is because the number of inliers in one of some groups may not be sufficient to yield a minimal subset. Generally, there are $C_{m_0}^2$ new groups if we combine any two groups in $\bm{\mathcal{G}}$, where $m_0$ is the number of groups in $\bm{\mathcal{G}}$. However, this combination strategy is not effective since it will generate a large number of ``insignificant" combined groups (i.e., the keypoint correspondences in a combined group may belong to different structures in data), and it is also very time-consuming. Therefore, we combine groups by using the rule, i.e., the neighboring constraint, to avoid the above-mentioned problem. Note that the keypoint correspondences of one group and the neighboring groups have a high possibility of belonging to the same structure. Thus, we propose to only combine the neighboring groups.

\begin{figure}[h]
\centering
\centerline{\includegraphics[width=0.48\textwidth]{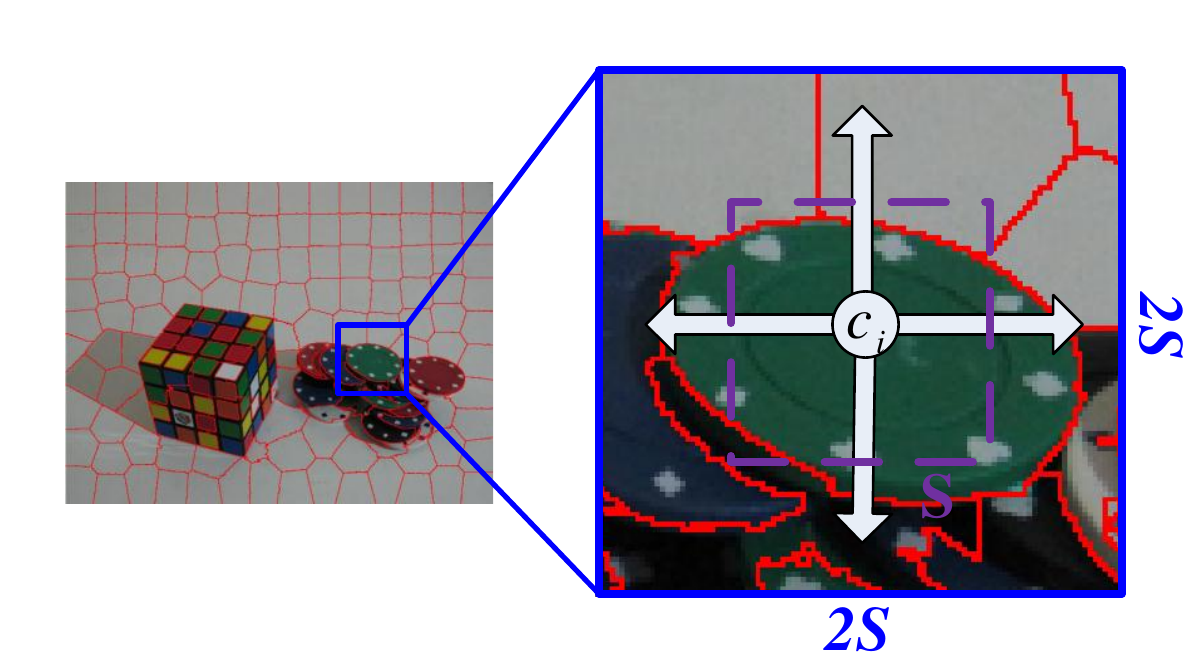}}
\caption{An example of group combination. $c_i$ is the center of the $i$-{th} group in $\bm{\mathcal{G}}$ and $S$ is the grid interval. The purple dashed box denotes a superpixel size and we perform the procedure of group combination within a $2S\times2S$ region (i.e., the blue solid box).}
\label{fig:limitedregion}
\end{figure}

More specifically, for a group $\mathcal{G}_i \in \bm{\mathcal{G}}$, we combine it with each of its neighboring group within a limited region to generate a new group $\hat{\mathcal{G}}_{i\cup j}$:
\begin{align}
 \label{equ:combine}
\hat{\mathcal{G}}_{i\cup j}&=\left\{ \begin{array}    {r@{\quad \quad \quad} l}
\mathcal{G}_i\cup \mathcal{G}_j, & if~\mathcal{G}_j\in \bm{\mathcal{N}(}\mathcal{G}_i\bm)\\
& ~~~and~R(l_i,l_j)\leq 2S\times2S,\\
\mathcal{G}_i~~~~,&otherwise,
\end{array}\right.
\end{align}
where $\bm{\mathcal{N}(}\mathcal{G}_i\bm)$ is the neighboring group of $\mathcal{G}_i$. $l_i$ and $l_j$ denote the corresponding superpixels of  $\mathcal{N(G}_i)$ and $\mathcal{N(G}_j)$, respectively. $R(.,.)$ denotes the combined region of two superpixels in the image. Here, according to the expected superpixel size ($S$$\times $$S$), we compute the grid interval $S$ as \cite{achanta2012slic}, i.e., $S$$=$$\sqrt{N/M}$, where $N$ and $M$ are the number of pixels and superpixels, respectively. An example of group combination is illustrated in Fig. \ref{fig:limitedregion}.

In this manner, only small-size groups (whose sizes are smaller than $2S$$\times$$2S$) are combined, by which it increases the number of inliers in a combined group while reducing the risk of generating ``insignificant" combined groups. Of course, we cannot guarantee that all new groups $\bm{\hat\mathcal{G}}$ are significant, but this manner effectively increases the number of groups that include sufficient inliers to generate a model hypothesis.

Moreover, we note that the feature points in the same superpixel may correspond to some bad keypoint correspondences (i.e., outliers), that is, the keypoint correspondences in a group may contain outliers. Obviously, we cannot generate an effective model hypothesis if we use all keypoint correspondences in such a group as a sampled subset. Therefore, we propose to only consider the most ``promising" keypoint correspondences in each group $\hat{\mathcal{G}}_{i}$$=$$\{x_{i}^j\}_{j=1}^{n_i}$. Matching scores (computed according to the SIFT correspondences \cite{lowe2004distinctive}) are used to judge if a data point is promising, because a data point is often assigned a high value if it is an inlier and a low value otherwise \cite{brahmachari2013hop}. Specifically, given a group $\hat{\mathcal{G}}_{i}$ and the corresponding matching scores $\bm{s_i}$$=$$[s_i^j~s_i^2~\cdots~s_i^{n_i}]$, we compute a permutation of each data point in $\hat{\mathcal{G}}_{i}$:
\begin{align}
\label{equ:sorting2}
\bm{a_i}=[a_i^1~a_i^2~\cdots~a_i^{n_i}],
\end{align}
where $a_i^j$ is the ranking index of the $j$-th data point in the group $\hat{\mathcal{G}}_{i}$. The keypoint correspondences in $\hat{\mathcal{G}}_{i}$ are sorted according to their matching scores in non-ascending order, i.e.,
\begin{align}
 \label{equ:sorting}
u<v \Longrightarrow s_i^{a_i^u} > s_i^{a_i^v},
\end{align}
where $u$ and $v$ respectively denote the indices of $x_{i}^u$ and $x_{i}^v$ in $\hat{\mathcal{G}}_{i}$.

\begin{algorithm}[t] 
\renewcommand{\algorithmicrequire}{\textbf{Input:}}
\renewcommand\algorithmicensure {\textbf{Output:} }
\caption{The proposed deterministic sampling algorithm} 
\label{alg:initisampling} 
\begin{algorithmic}[1] 
\REQUIRE 
An image pair, keypoint correspondences and the number of superpixels.
\STATE Perform the superpixel segmentation algorithm \cite{achanta2012slic} on the input image pair.
\STATE Partition keypoint correspondences into groups $\bm{\mathcal{G}}$ based on the segmented superpixels (described in Sec. \ref{sec:superpixel}).
\STATE Generate new groups $\bm{\hat\mathcal{G}}$ by combining groups in $\bm{\mathcal{G}}$ according to Eq. (\ref{equ:combine}).
\STATE Sort the keypoint correspondences in each new group by Eq. (\ref{equ:sorting}).
\STATE Select the top ($p+2$) sorted keypoint correspondences in each group as a sampled subset to generate a model hypothesis $\theta_i$. For $m$ groups in $\bm{\hat\mathcal{G}}$, a set of model hypotheses $\bm{\theta}$ with $m$ elements are generated.
\ENSURE The generated model hypothesis set {$\bm{\theta}$} (=$\{\theta_i\}_{i=1}^m$)
\end{algorithmic}
\end{algorithm}
In the permutation $\bm{a_i}$, we only select the top ($p+2$) sorted keypoint correspondences, i.e., $\{x_{i}^j\}_{j=a_i^1}^{a_i^{p+2}}$, as a sampled subset for $\hat{\mathcal{G}}_{i}$. Here, $p$ is the size of a minimum subset, and we select ($p+2$) keypoint correspondences since ($p+2$) keypoint correspondences are able to generate a more stable model hypothesis (which has been demonstrated in \cite{tennakoon2016robust}). We only sample the keypoint correspondences with high matching scores in a group, which can effectively reduce the influence of outliers.
\subsection{Model Hypothesis Updating}
\label{sec:hypothesesupdating}
\begin{figure*}
\centering
\subfigure{
\begin{minipage}{.25\textwidth}
\centerline{\includegraphics[width=0.9\textwidth]{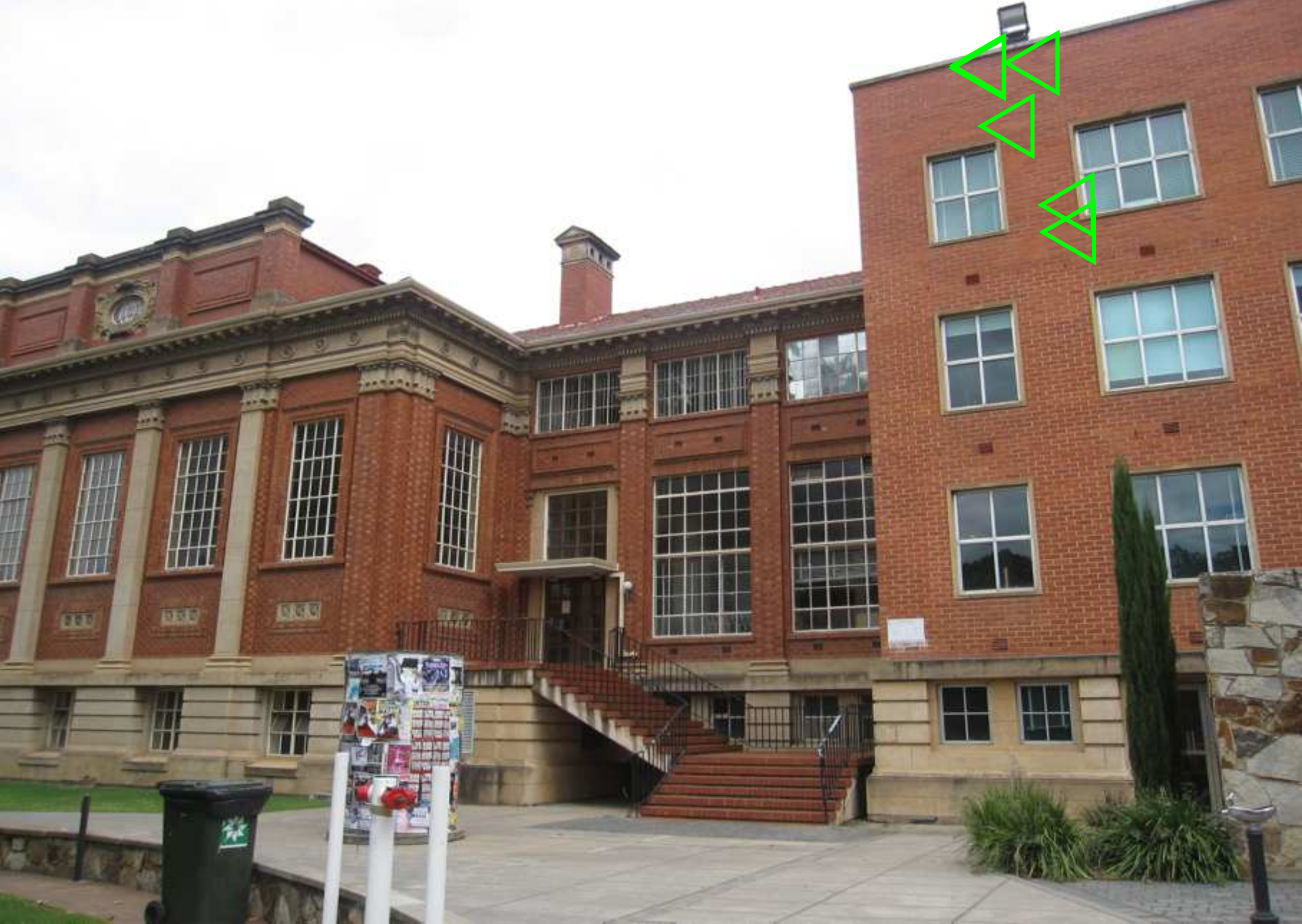}}
\end{minipage}
\begin{minipage}{.25\textwidth}
\centerline{\includegraphics[width=0.9\textwidth]{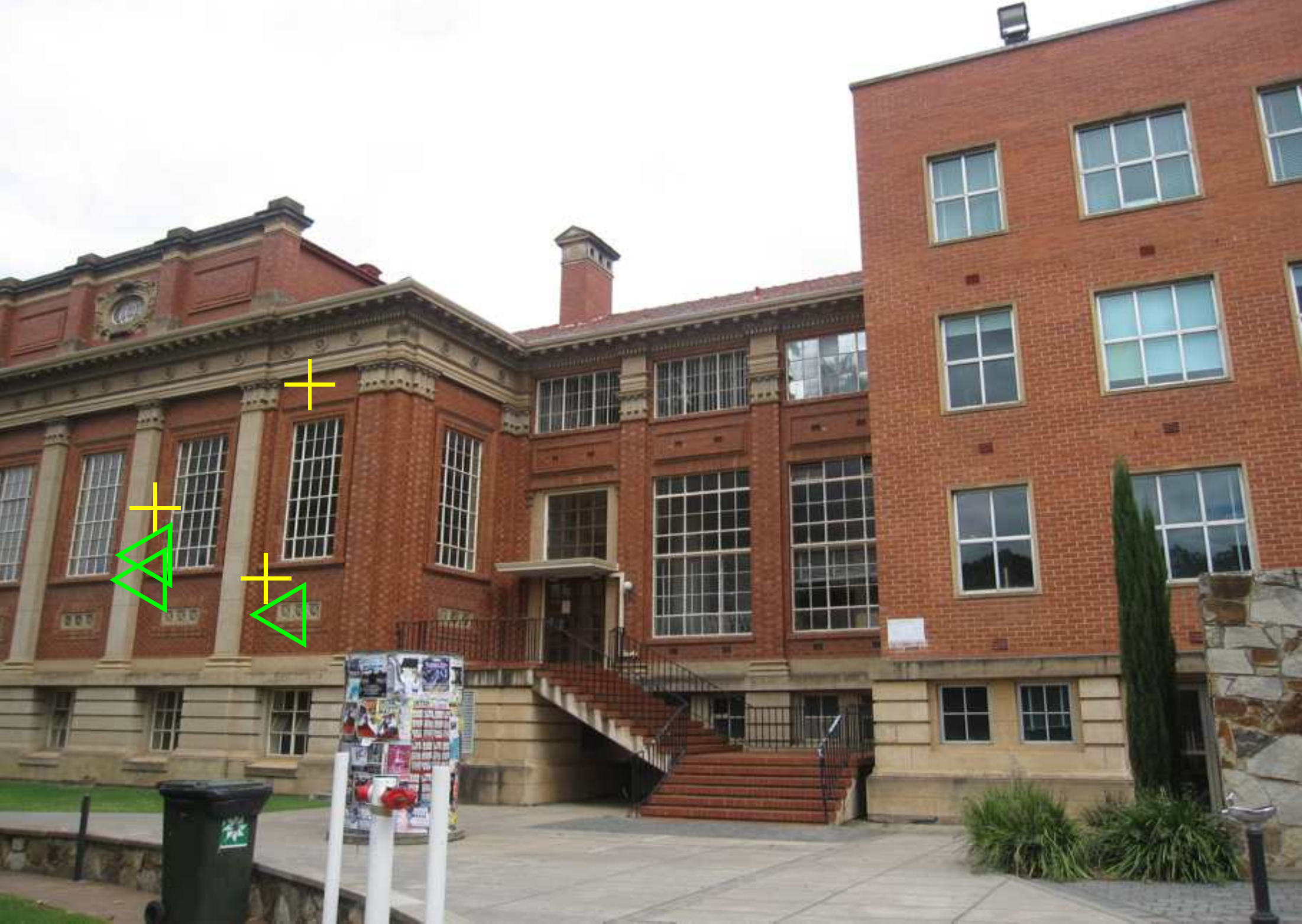}}
\end{minipage}
\begin{minipage}{.25\textwidth}
\centerline{\includegraphics[width=0.9\textwidth]{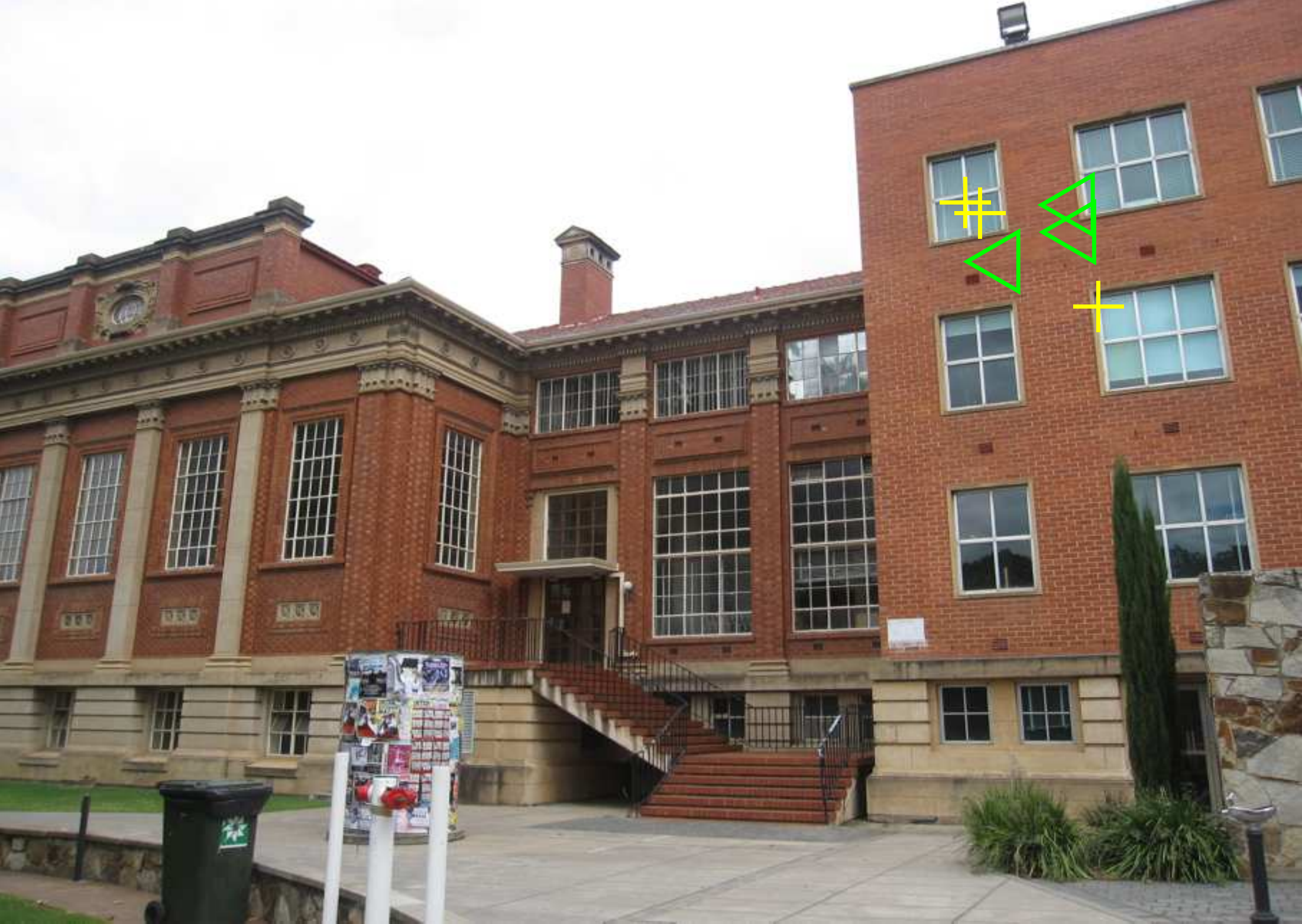}}
\end{minipage}
\begin{minipage}{.25\textwidth}
\centerline{\includegraphics[width=0.9\textwidth]{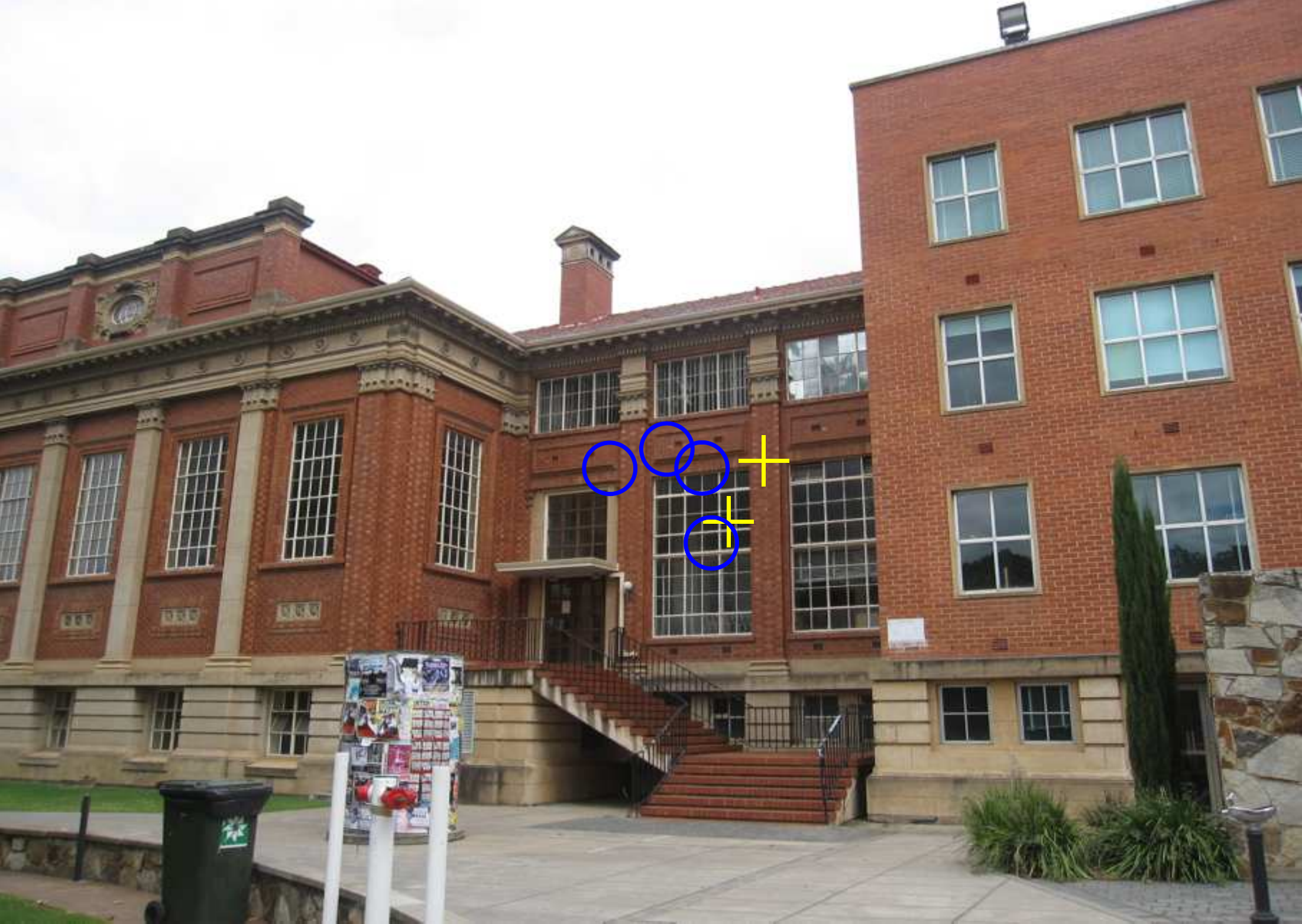}}
\end{minipage}

}
\subfigure{
\begin{minipage}{.25\textwidth}
\centerline{\includegraphics[width=0.9\textwidth]{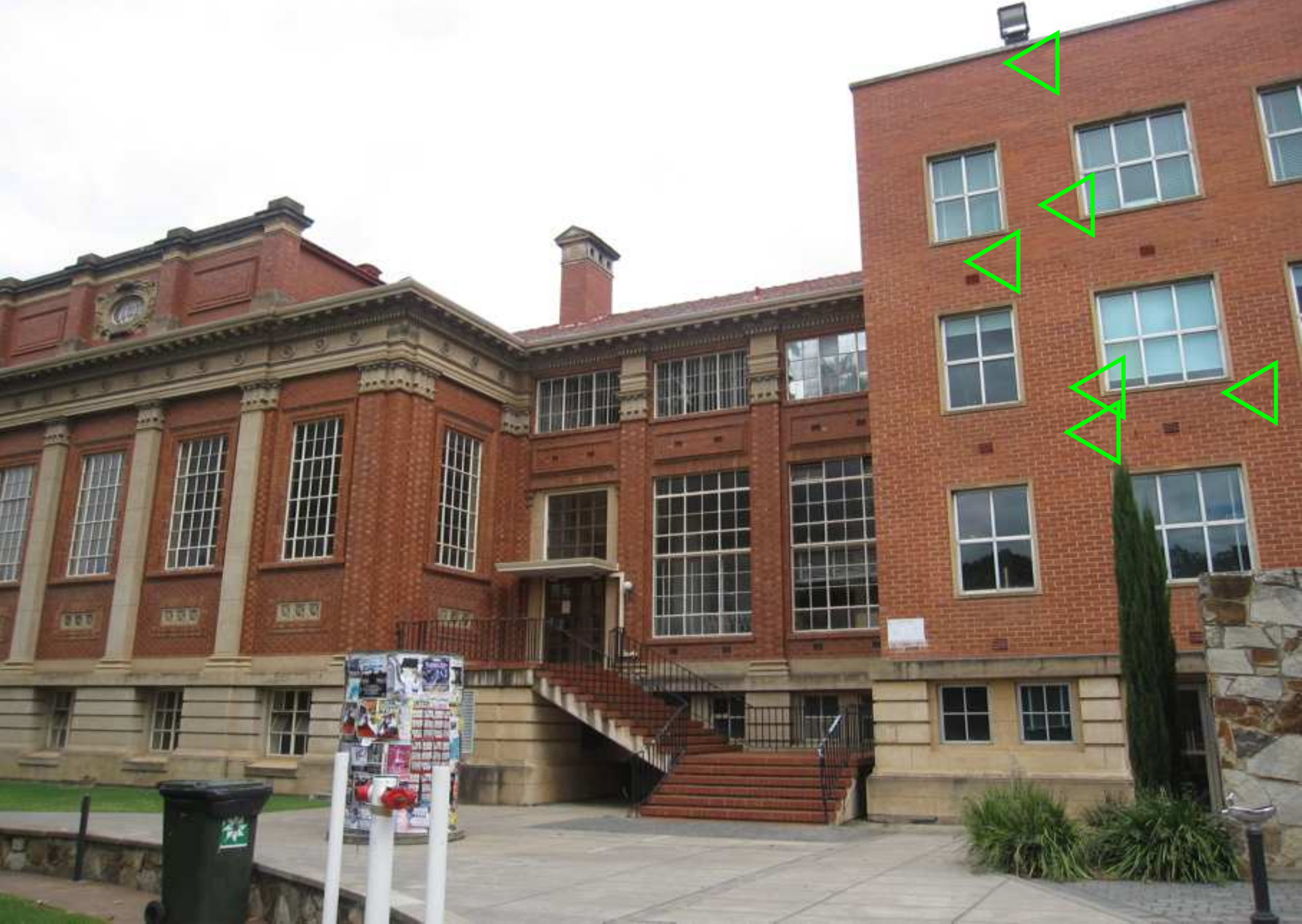}}
\end{minipage}
\begin{minipage}{.25\textwidth}
\centerline{\includegraphics[width=0.9\textwidth]{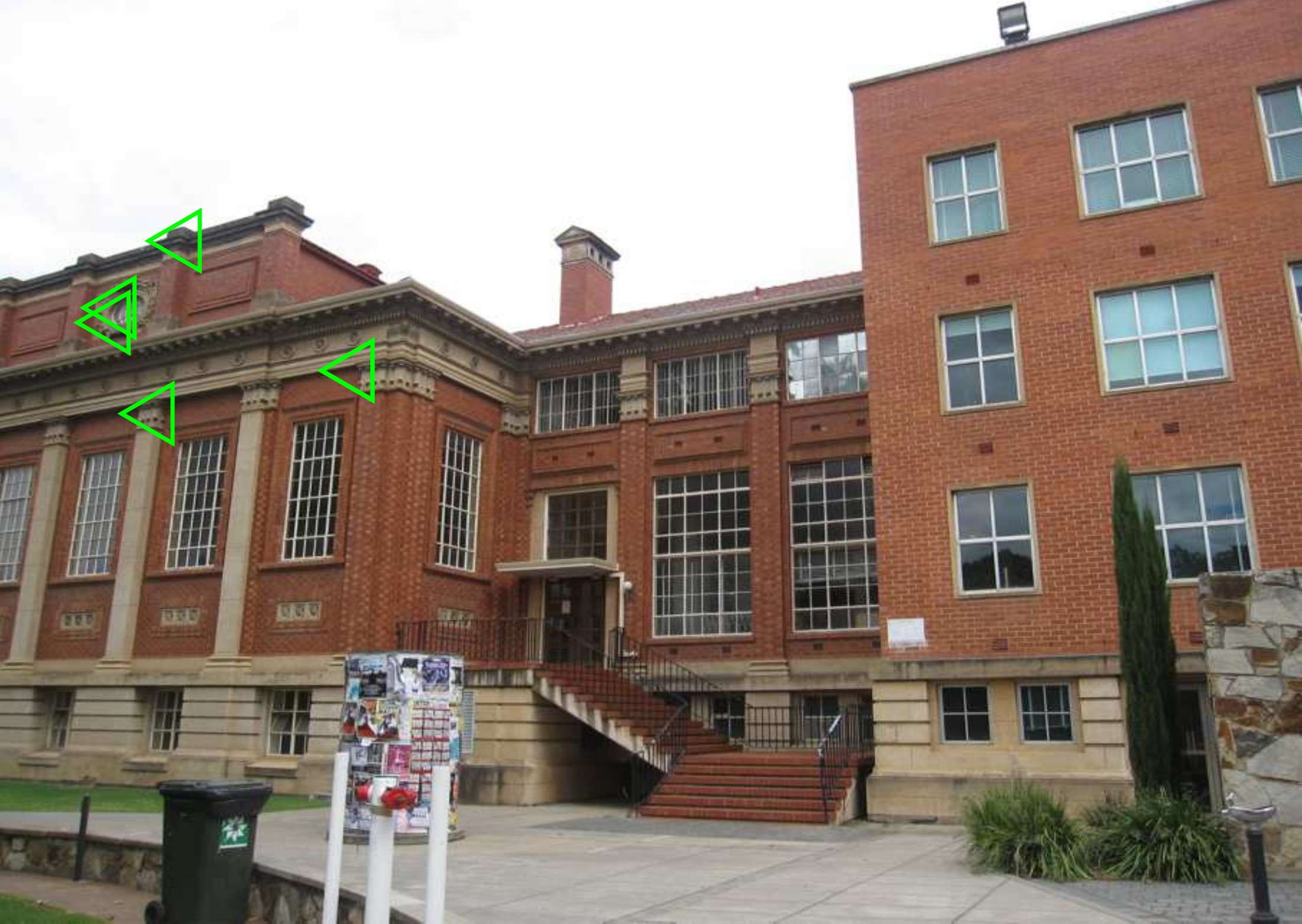}}
\end{minipage}
\begin{minipage}{.25\textwidth}
\centerline{\includegraphics[width=0.9\textwidth]{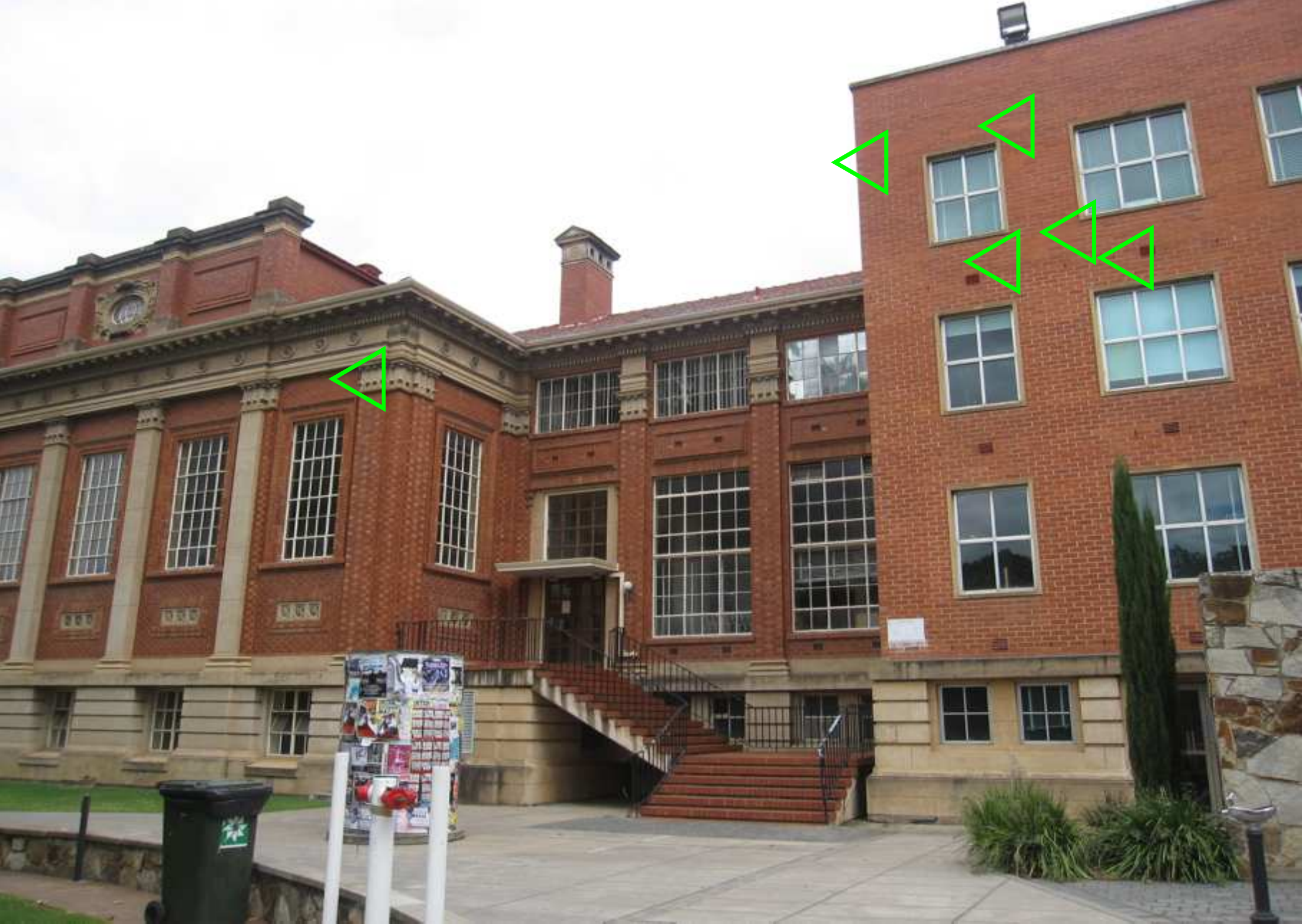}}
\end{minipage}
\begin{minipage}{.25\textwidth}
\centerline{\includegraphics[width=0.9\textwidth]{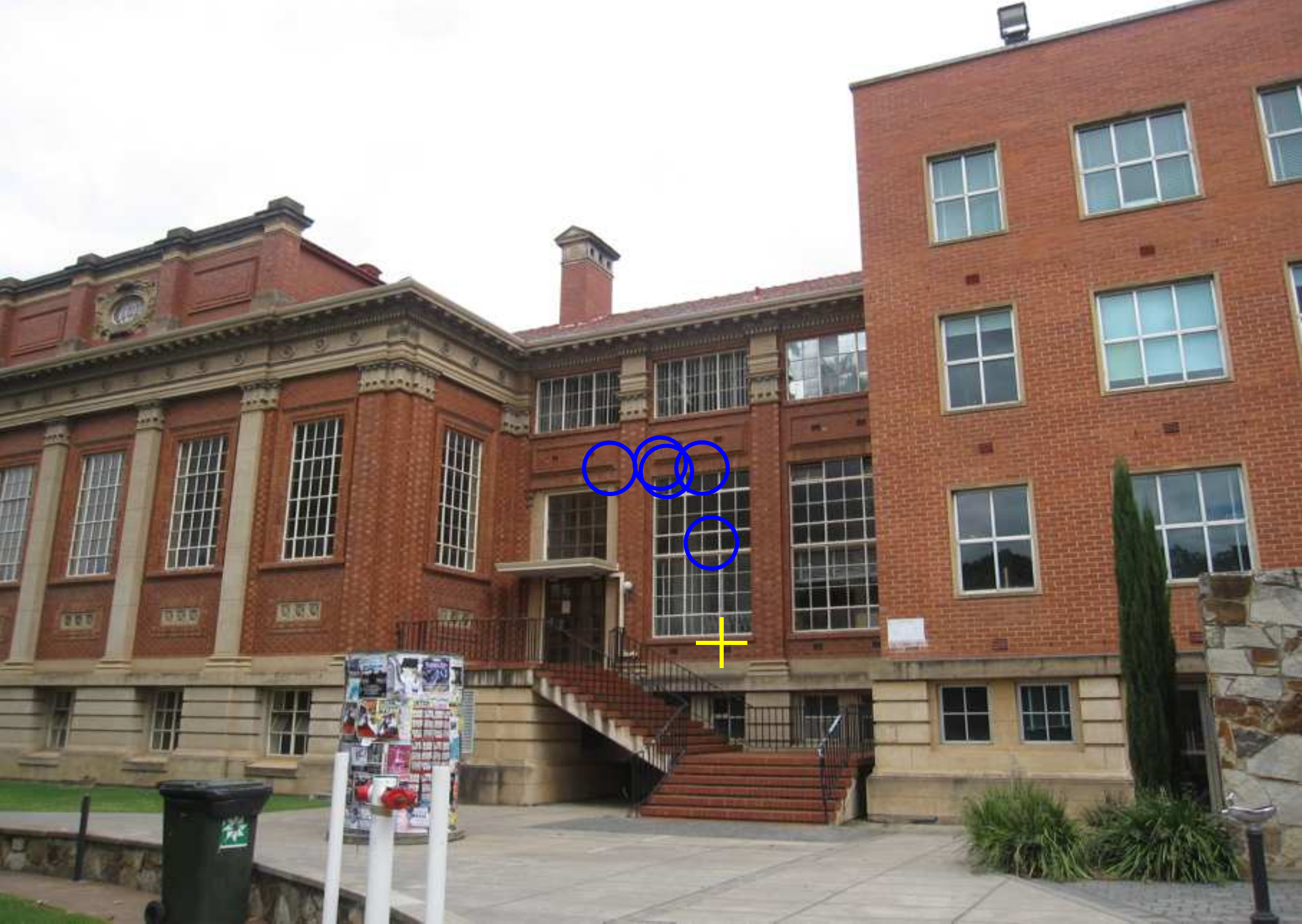}}
\end{minipage}
}
\subfigure{
\begin{minipage}{.25\textwidth}
\centerline{\includegraphics[width=0.9\textwidth]{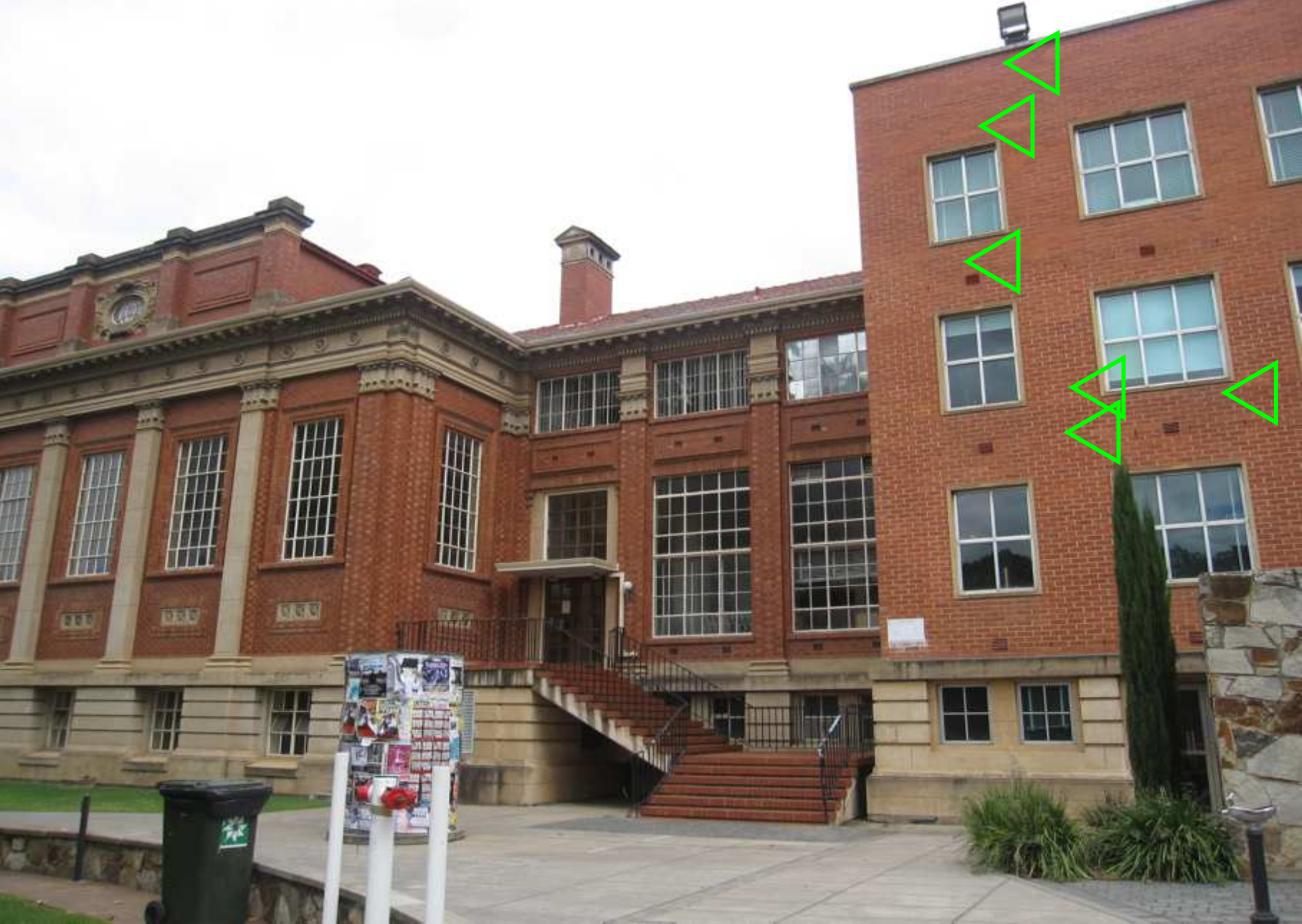}}
  \centerline{(a) }
\end{minipage}
\begin{minipage}{.25\textwidth}
\centerline{\includegraphics[width=0.9\textwidth]{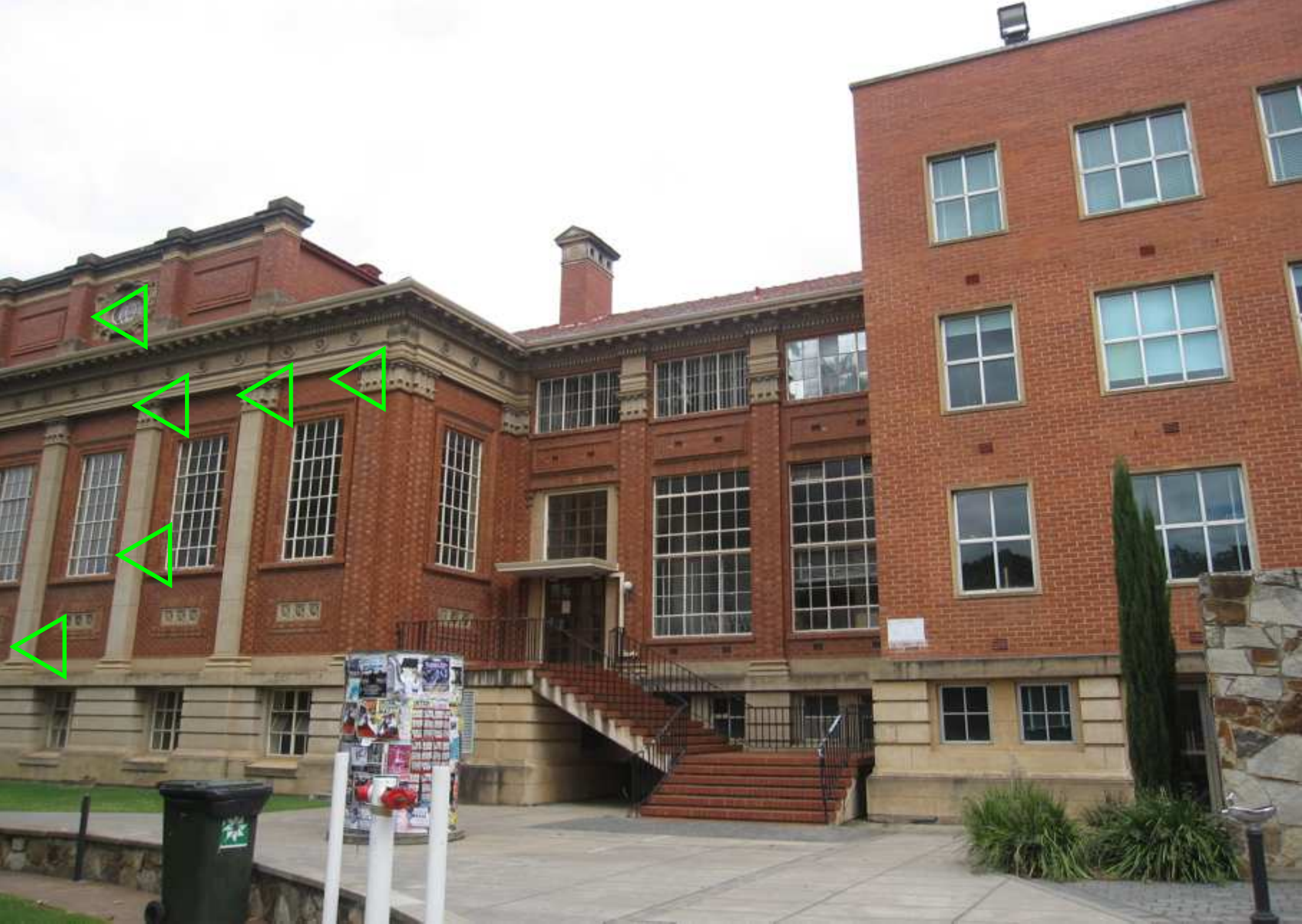}}
  \centerline{(b) }
\end{minipage}
\begin{minipage}{.25\textwidth}
\centerline{\includegraphics[width=0.9\textwidth]{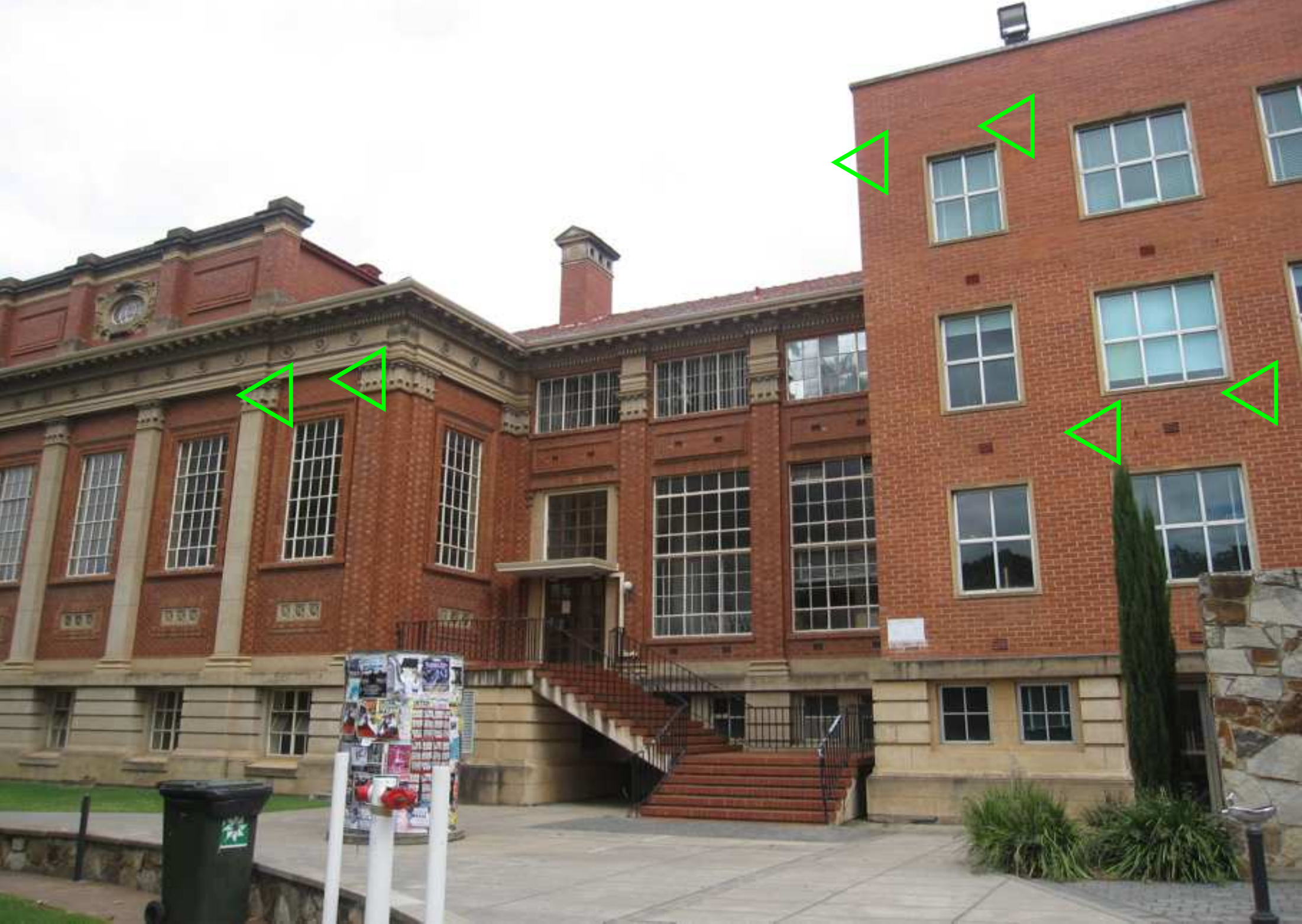}}
  \centerline{(c) }
\end{minipage}
\begin{minipage}{.25\textwidth}
\centerline{\includegraphics[width=0.9\textwidth]{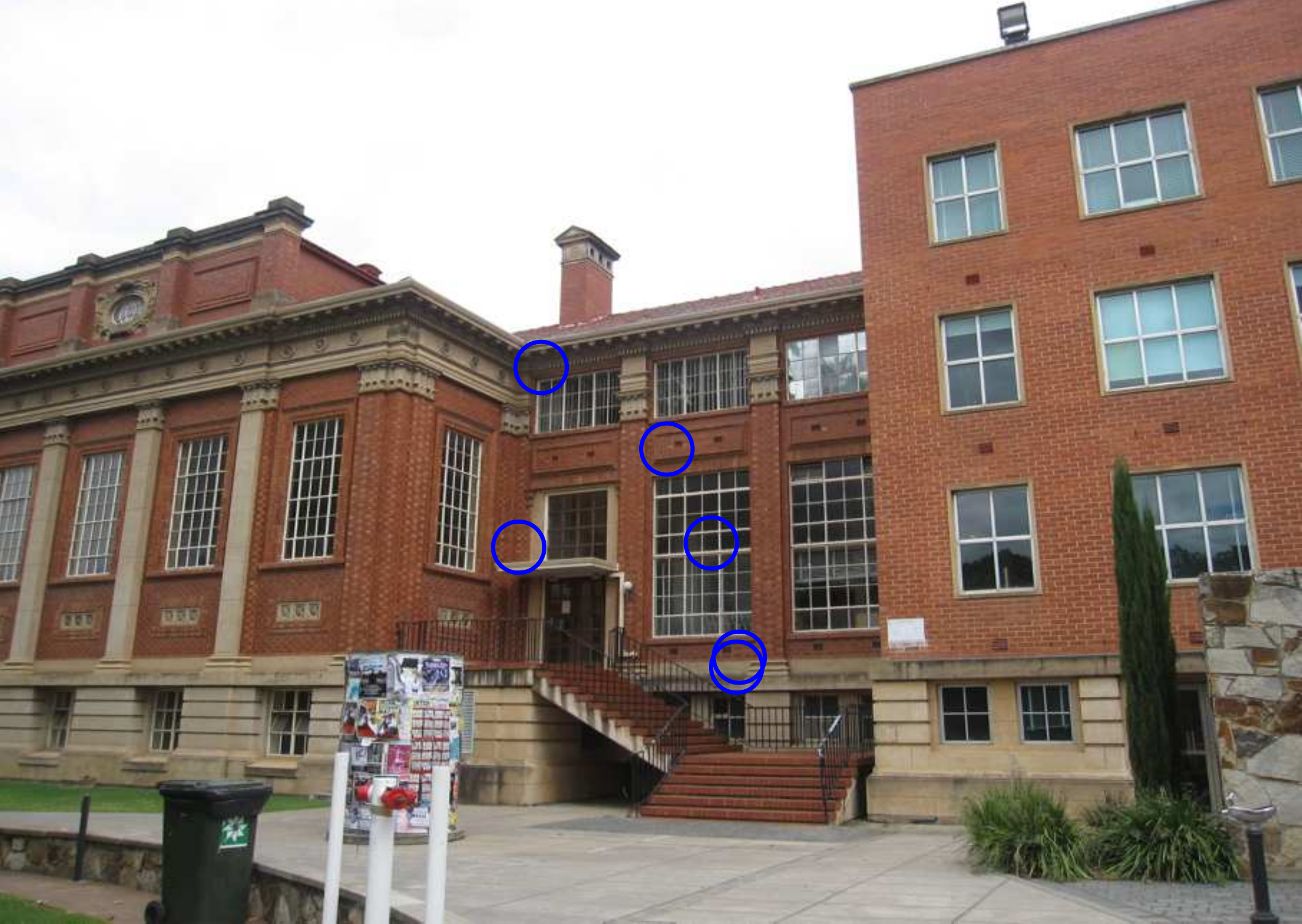}}
  \centerline{(d)}
\end{minipage}
}
\caption{Some updated results obtained by PM \cite{tennakoon2016robust} and the proposed MHU strategy on the ``Barrsmith" image pair (only one of the two views is shown) for homography estimation. $1^{st}$ to $3^{rd}$ rows respectively show the input data, the updated results by obtained PM and the updated results obtained by MHU. The keypoint correspondences belonging to different model instances are marked by different colors according to the ground truth results (see Fig. \ref{fig:showproblems}(b)). (a)$\sim$(d) show the updated results on four different examples. }
\label{fig:hypothesisupdating}
\end{figure*}

Based on the prior information of feature appearances and the two above-mentioned strategies (i.e., the group combination and the promising keypoint correspondences selection strategies), we can deterministically generate a set of model hypotheses (see Algorithm \ref{alg:initisampling}). In most cases, we can find some significant model hypotheses that are close to the true model instances from the generated model hypotheses. However, in the case that the spatial distribution of the inliers of a model instance relatively widely spreads, such as the plane marked in green in Fig. \ref{fig:showproblems}(b), it is hard to generate significant model hypotheses based on superpixels (even using the strategy of ``group combination"). For this problem, we introduce a model hypothesis updating strategy to improve the quality of the generated model hypotheses.

Note that the inliers of a significant model hypothesis usually have low residual values and high matching scores, based on which, we propose a Model Hypothesis Updating (called MHU) strategy in Algorithm \ref{alg:hypothesisupdating}. Specifically, for each generated model hypothesis $\theta_i$, we sort the residual values derived from $\theta_i$ and the input keypoint correspondences {$\bm{X}$} (=$\{x_i\}_{i=1}^n$); and then we assign $\theta_i$ a positive weighting value (see Eq. (\ref{equ:weighting})). After that, we terminate the iterations if the stopping criterion is satisfied (see Eq. (\ref{equ:stoping})); Conversely, we update the model hypothesis if the stopping criterion is not satisfied. For each iteration, we select the model hypothesis with the largest weighting value as the updated model hypothesis.

We assign a model hypothesis $\theta_i$ a positive weighting value, similar to \cite{wang2012simultaneously,xiao2016HF}:
\begin{equation}
\label{equ:weighting}
w(\theta_i)=\frac{1}{n}\sum_{j=1}^n\frac{\mathbf{EK}(r_i^j/{{b}({\theta_i})})}{\tilde{s}({\theta_i}){b}({\theta_i})},
\end{equation}
where $n$ is the number of keypoint correspondences; $\tilde{s}({\theta_i})$ is the estimated inlier noise scale of the $i$-th model hypothesis; ${r}_i^j$ is the residual derived from the $i$-th model hypothesis and the $j$-th keypoint correspondence; ${b}({\theta_i})$ is the bandwidth of the $i$-th model hypothesis, and it is defined as \cite{wand1994kernel}
\begin{align}
\label{equ:bandwith}
{b}(\theta_i)=\left[\frac{243\int_{-1}^1{\mathbf{EK}(\lambda)}^2d\lambda}{35n\int_{-1}^1{\lambda}^2\mathbf{EK}(\lambda)d\lambda}\right]^{0.2}\tilde{s}({\theta_i});
\end{align}
For the kernel function $\mathbf{EK}(\cdot)$, we employ the popular Epanechnikov kernel, which is written as follows:
\begin{align}
\label{equ:kernel}
\mathbf{EK}(\lambda) &=\left\{ \begin{array}    {r@{\quad \quad} l}
0.75(1-{\|\lambda\|}^2), & \|\lambda\|\leq 1,\\
0~~~~~~~~,&\|\lambda\|> 1.
\end{array}\right.\;
\end{align}
As discussed in \cite{wang2012simultaneously,xiao2016HF}, a model hypothesis with {a larger number of inliers, and with smaller residual values,} is assigned a higher weighting value according to Eq.~(\ref{equ:weighting}).

\begin{algorithm}[t] 
\renewcommand{\algorithmicrequire}{\textbf{Input:}}
\renewcommand\algorithmicensure {\textbf{Output:} }
\caption{The proposed model hypothesis updating strategy} 
\label{alg:hypothesisupdating} 
\begin{algorithmic}[1] 
\REQUIRE 
Keypoint correspondences {$\bm{X}$}, the generated model hypothesis set {$\bm{\theta}$} (=$\{\theta_i\}_{i=1}^m$),  the support size ($\hat{n}$), the maximum number of iterations $t_{max}$.
\FOR{$i=1$ to $m$}
\STATE $t\leftarrow1$; $\hat{\theta}_i^t\leftarrow\theta_i$;
\REPEAT
\STATE  $\bm{rank}^t \leftarrow Sort\big(Res(\bm{X},\hat{\theta}_i^t)\big)$; // $\bm{rank}^t=\{rank_j^t\}_{j=1}^n$.
\STATE Assign a weighting value $w(\hat{\theta}_i^t)$ to $\hat{\theta}_i^t$ by Eq.~(\ref{equ:weighting});
\STATE if $t_{stop}$ then //See Eq.~(\ref{equ:stoping}).
\STATE  ~~break;
\STATE end if
\STATE $\hat{\theta}_i^{t+1}\leftarrow Fit([x_{rank_j}]_{j={\hat{n}}-p-1}^{\hat{n}})$ //Fit a model hypothesis.
\STATE $t$$+$$+$;
\UNTIL $t<t_{max}$
\STATE $\theta_i^*\leftarrow argmax \{w(\hat{\theta}_i^j)\}_{j=1,2,\cdots}$
\ENDFOR
\ENSURE The updated model hypothesis set {$\bm{\theta}^*$} (=$\{\theta_i^*\}_{i=1}^m$)
\end{algorithmic}
\end{algorithm}
The stopping criterion is computed as follows:
\begin{equation}
\label{equ:stoping}
\begin{split}
t_{stop}=\Bigg( \frac{1}{\hat{n}}\Big|\{rank_j^t\}_{j=1}^{\hat{n}}\cap \{rank_j^{t-1}\}_{j=1}^{\hat{n}}\Big|>\epsilon \Bigg)\\
\wedge \Bigg(  \frac{1}{\hat{n}}\Big|\{rank_j^t\}_{j=1}^{\hat{n}}\cap \{rank_j^{t-2}\}_{j=1}^{\hat{n}}\Big|>\epsilon \Bigg),
 \end{split}
\end{equation}
where $\Big|\{rank_j^t\}_{j=1}^{\hat{n}}\cap \{rank_j^{t-1}\}_{j=1}^{\hat{n}}\Big|$ denotes the number of the common inliers of the model hypotheses generated in the $t$-th and $(t-1)$-th iterations. This criterion checks the model hypotheses generated at the $t$-th and $(t-1)$-th iterations, to judge if they share most of the common inliers (assuming that a model instance has at least $\hat{n}$ inliers). This indicates that the updated model hypotheses in the last three iterations have a high probability of belonging to the same structure and hence the iteration can be stopped. The value of $\epsilon$ is fixed to be $0.8$, which denotes that the two model hypotheses share $80\%$ common inliers.



The proposed model hypothesis updating (MHU) strategy is inspired by PM~\cite{tennakoon2016robust}. Although both the proposed MHU strategy and PM compute the least $\hat{n}$-th order statistics to improve each generated model hypothesis, they are significantly different: 1) MHU introduces weighting values to select the updated model hypotheses while PM directly selects the model hypothesis updated in the last iteration. The weighting values can measure the quality of model hypotheses, and thus the model hypotheses updated by MHU are more effective than the ones updated by PM. 2) MHU considers the information of $\hat{n}$ inliers in the stopping criterion, while PM only uses the information of sampled subsets in the stopping criterion. More keypoint correspondences can provide more information to judge if two model hypotheses belong to the same structure. 3) MHU is more efficient than PM, since that MHU can use parallel computation while PM has to iteratively update each generated model hypothesis. As a result, MHU is more effective and efficient than PM.

{To illustrate the differences between PM and MHU}, we show some results obtained by PM and MHU in Fig.~\ref{fig:hypothesisupdating}. We can see that, for the all-inlier sampled subset (Fig.~\ref{fig:hypothesisupdating}(a)), both PM and MHU can obtain good results. However, for the sampled subset consisting of both inliers and outliers from the ``wide" model instance  (Figs.~\ref{fig:hypothesisupdating}(b)$\sim$(c)), although both can obtain all-inlier subsets, the subsets updated by MHU are more evenly distributed with larger spans than those updated by PM. For the subset consisting of both inliers and outliers for the ``narrow" model instance (Fig.~\ref{fig:hypothesisupdating}(d)), PM cannot obtain an all-inlier subset while MHU can effectively update the subsets. Therefore, MHU is a significantly improved variant of PM.
\section{The proposed model selection algorithm}
\label{sec:modelselection}
\begin{figure}[t]
\centering
\begin{minipage}[t]{.23\textwidth}
  \centerline{\includegraphics[width=1.16\textwidth]{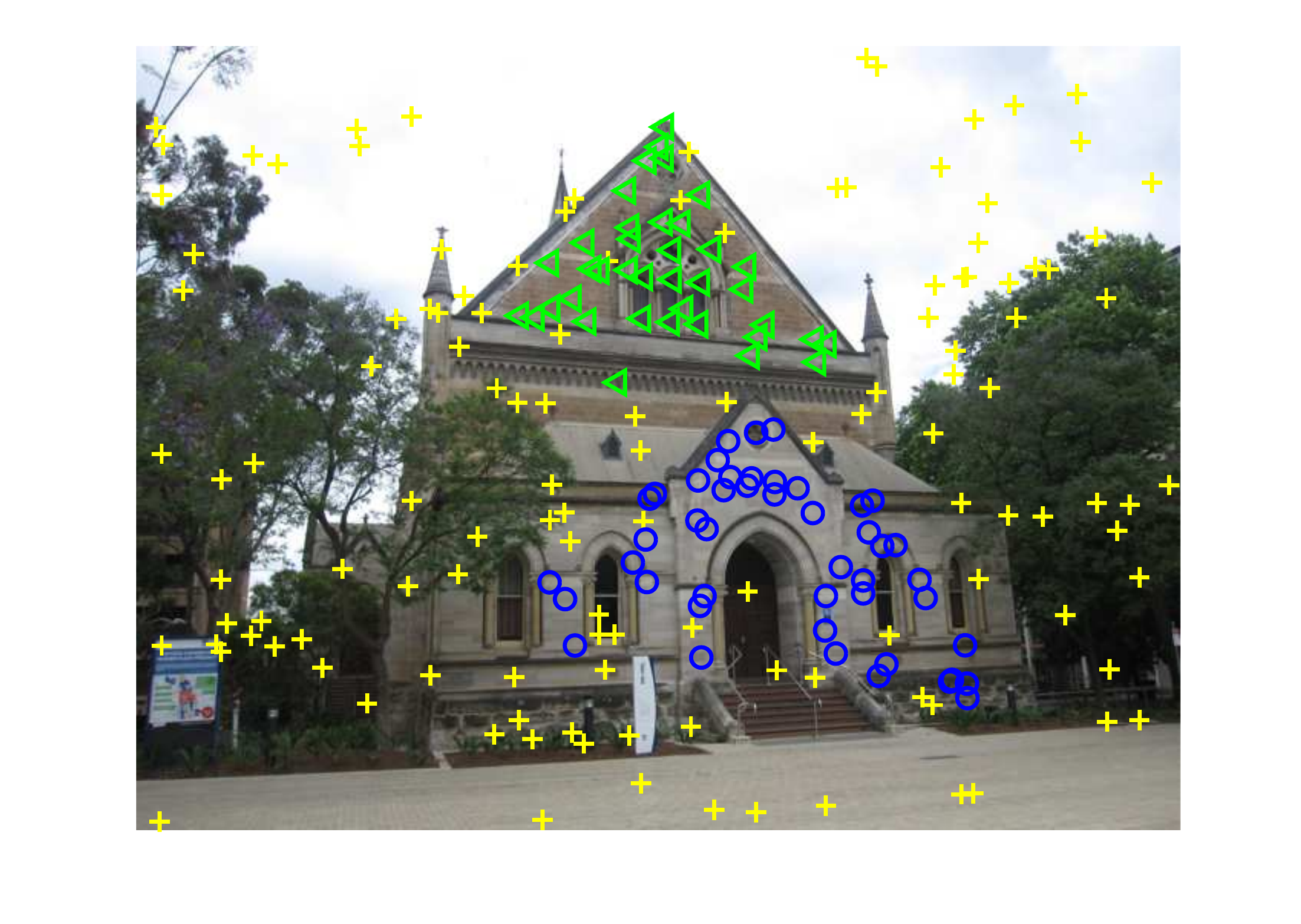}}
 \centerline{(a)}
  \centerline{\includegraphics[width=1.16\textwidth]{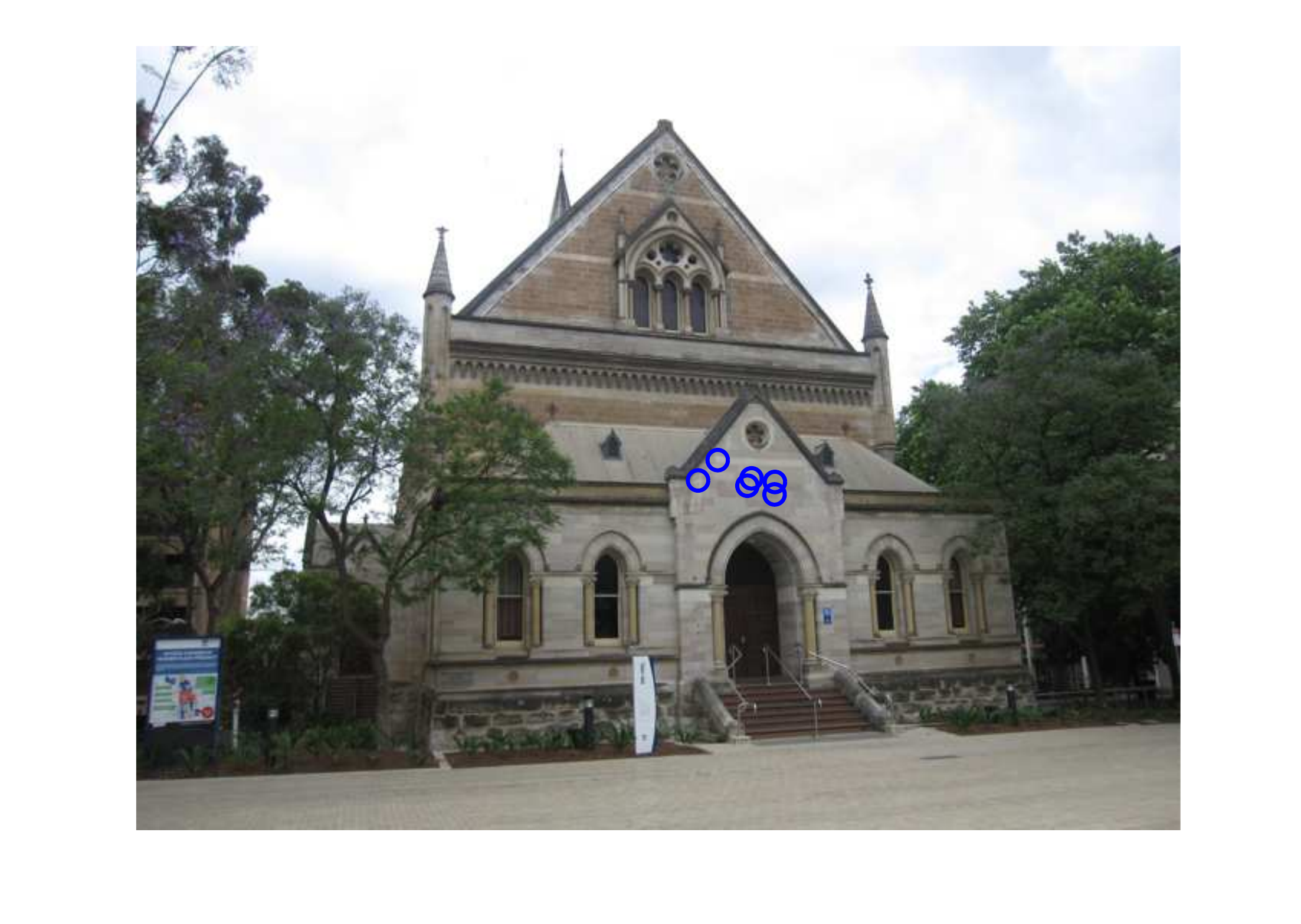}}
  \centerline {(c)}
      \centerline{\includegraphics[width=1.16\textwidth]{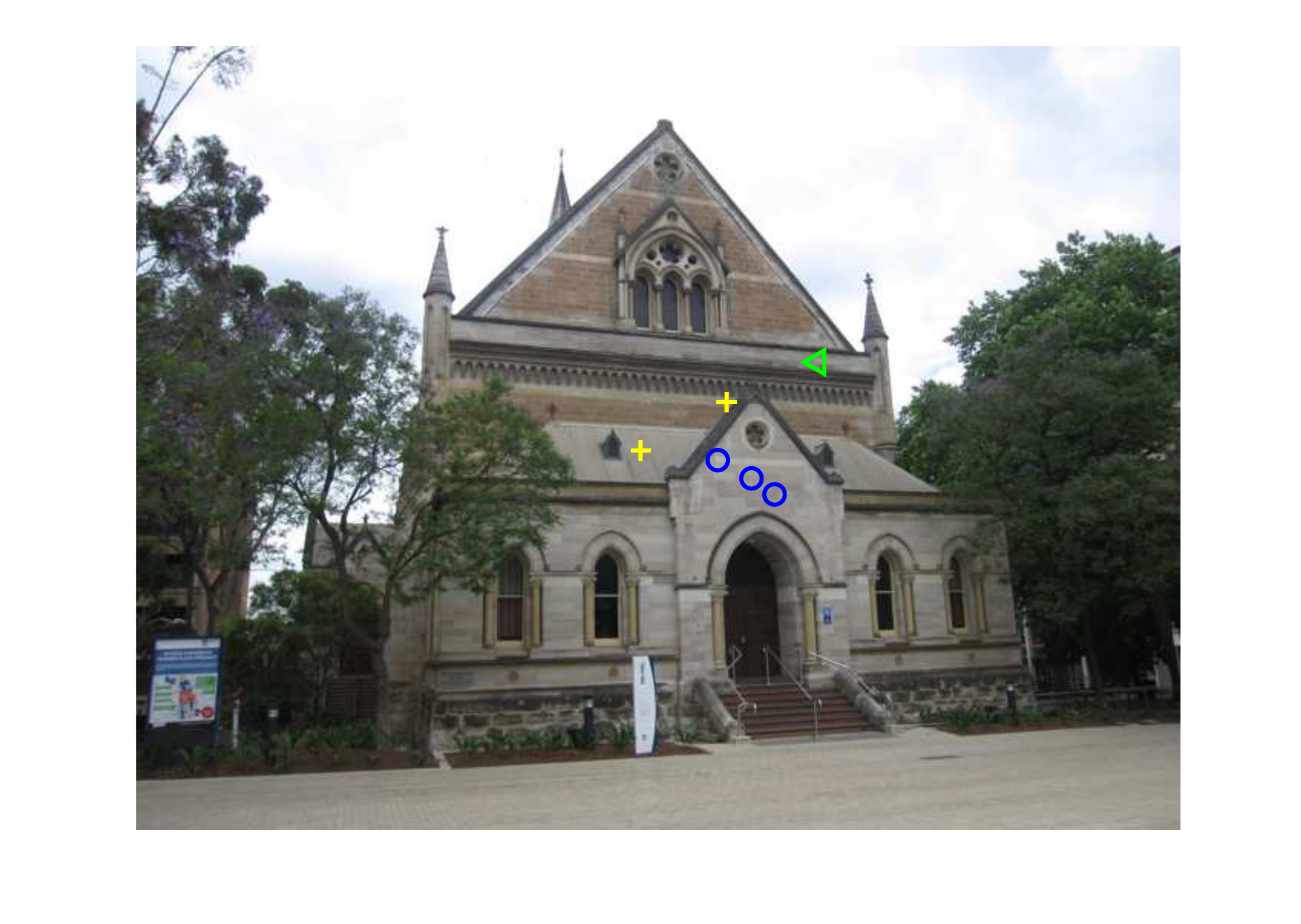}}
   \centerline {(e)}

\end{minipage}
\begin{minipage}[t]{.23\textwidth}
  \centerline{\includegraphics[width=1.16\textwidth]{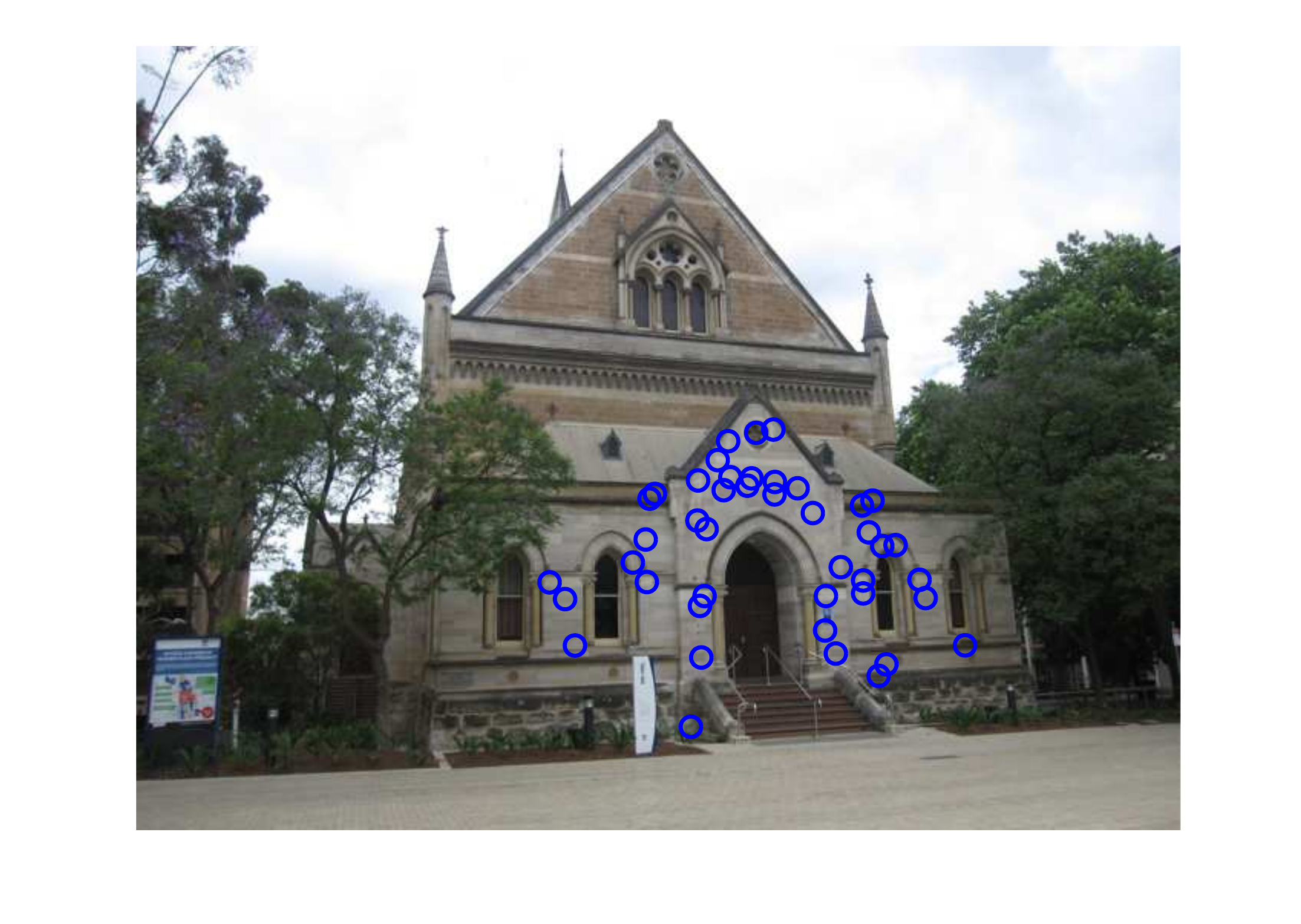}}
 \centerline{(b)}
  \centerline{\includegraphics[width=1.16\textwidth]{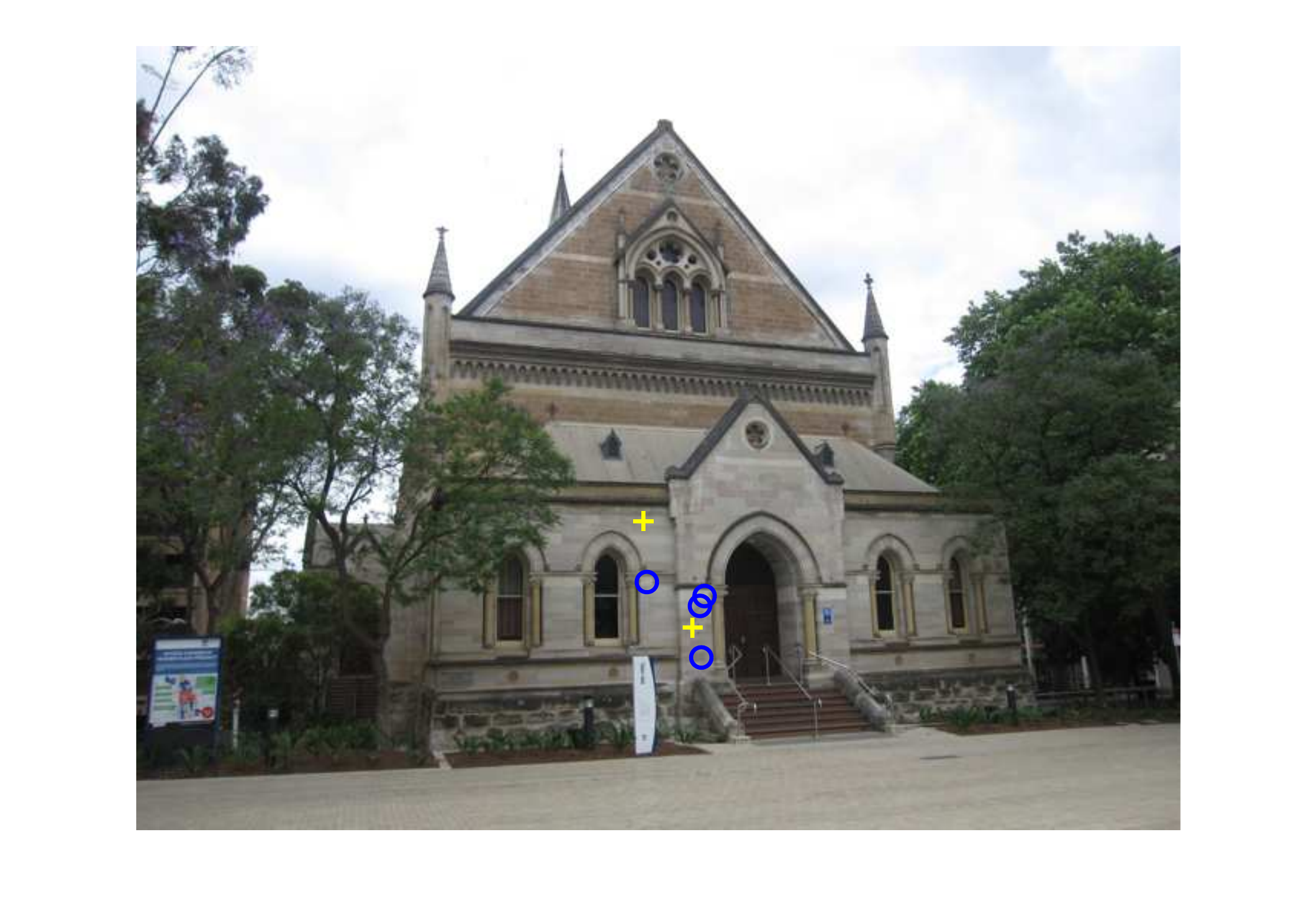}}
 \centerline {(d)}
   \centerline {\includegraphics[width=1.16\textwidth]{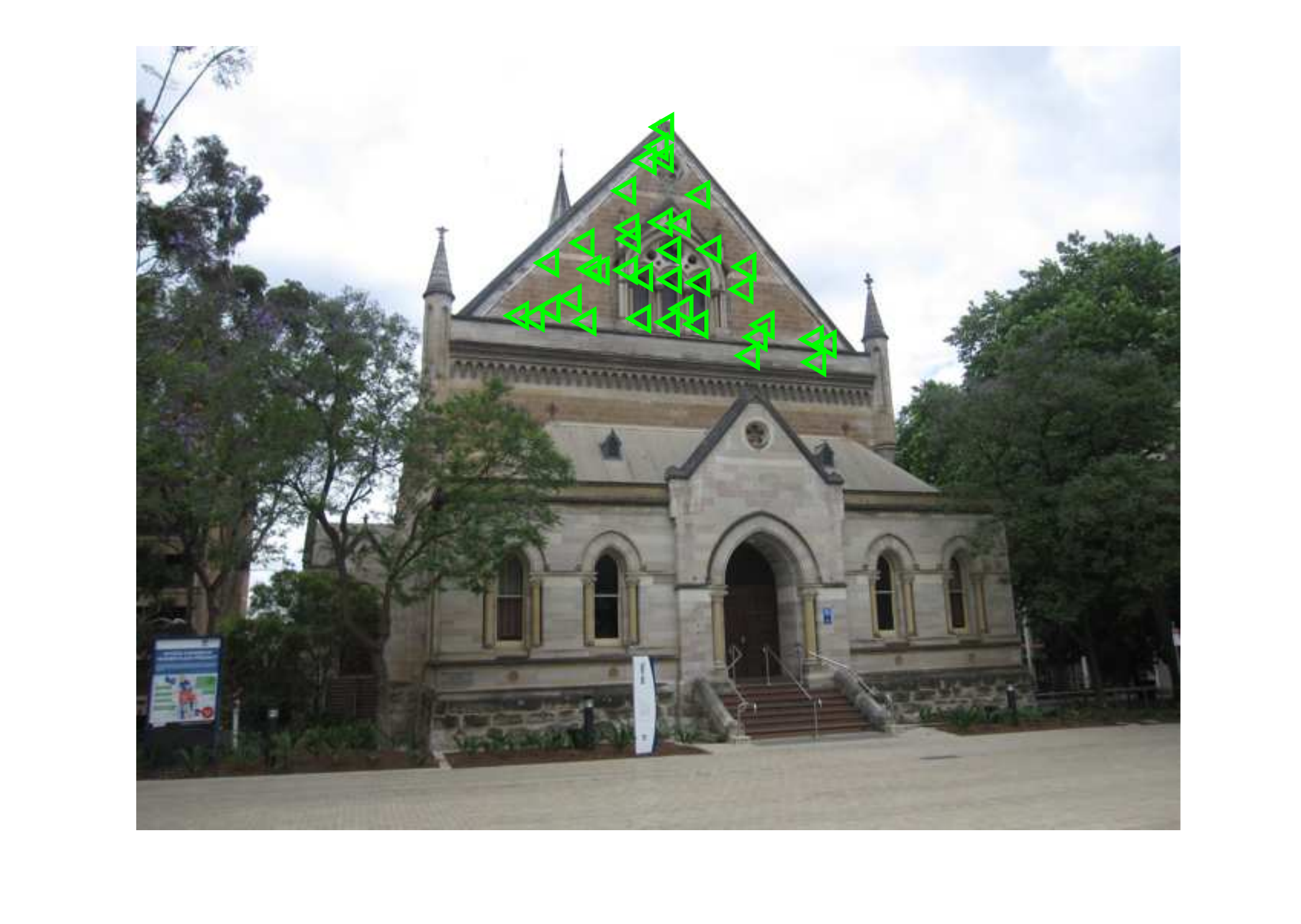}}
  \centerline{(f)}
\end{minipage}
\caption{An example of model selection for homography estimation (only one of the two views is shown): (a) An input image (``Elderhalla'') with the ground truth results. (b) The inliers of the first selected model hypothesis. (c)$\sim$(e) The sampled subsets of three redundant model hypotheses. (f) The inliers of the second selected model hypothesis.}
\label{fig:modelselection}
\end{figure}
In this section, we propose a novel model selection algorithm to estimate model instances in data. The main steps of the traditional ``fit-and-remove" framework (i.e., a popular model selection algorithm) include the following four steps: (1) Generate model hypotheses based on the remaining data points (which are the input data points at first); (2) Select the model hypothesis with the best evaluation score as an estimated model instance; (3) Stop if all model instances in data are estimated, and continue otherwise; (4) Remove the inliers of the estimated model instance from the remaining data points and go to Step (1).

The traditional ``fit-and-remove" framework can be used to extend some fitting methods (e.g.,~\cite{fischler1981random,chum2005matching}) that only work for single-structure data to handle multiple-structure data. However, this framework has some major limitations, e.g., inaccurate inlier/outlier dichotomy at the current iteration may lead to wrong estimation for the remaining model instances, and generating model hypotheses repeatedly at each iteration is time-consuming.

To relieve the limitations of the traditional ``fit-and-remove" framework, we propose a novel ``fit-and-remove" framework, whose main steps are described as follows: (1) Generate a set of model hypotheses based on the input data points; (2) Select the model hypothesis $\theta^*$ with the highest weighting value (computed by Eq.~(\ref{equ:weighting})) from the current model hypothesis set $\bm{\theta}^*$; (3) Stop if all model instances in data are estimated, and continue otherwise; (4) Remove redundant model hypotheses from $\bm{\theta}^*$ and go to Step (2).

Compared to the traditional ``fit-and-remove" framework, the proposed framework removes model hypotheses rather than data points, and it does not require to generate model hypotheses repeatedly at each iteration. Thus, the proposed framework significantly improves the efficiency of the traditional framework.

The judgement of redundant model hypotheses plays an important role for the proposed framework. We show an example of model selection for homography estimation in Fig.~\ref{fig:modelselection}. After selecting a model hypothesis $\theta_i^*$ (see Fig.~\ref{fig:modelselection}(b)), the redundant model hypotheses contain two parts: the model hypotheses, whose corresponding sampled subsets consist of outliers or inliers from different model instances (see Fig.~\ref{fig:modelselection}(d) and~\ref{fig:modelselection}(e)), and the model hypotheses,  whose corresponding sampled subsets consist of all inliers of $\theta_i^*$ (see Fig.~\ref{fig:modelselection}(c)). We define a function $h(i,j)$ to judge if a model hypothesis $\theta_j^*$  is redundant for $\theta_i^*$:
\begin{align}
 \label{equ:reduant}
h(i,j)&=\left\{ \begin{array}    {r@{\quad \quad} l}
1, & if~\mathbf{InS}(\theta_i^*) \cap \mathbf{SamS}(\theta_j^*) \neq \emptyset,\\
0,&otherwise,
\end{array}\right.
\end{align}
where $\mathbf{InS}(\theta_i^*)$ and $\mathbf{SamS}(\theta_j^*)$ are the inlier set of $\theta_i^*$ and the sampled subset of $\theta_j^*$, respectively. $\mathbf{InS}(\theta_i^*) \cap \mathbf{SamS}(\theta_j^*)$ denotes the same elements shared by the two sets. Thus, each model hypothesis $\theta_j^*$ with $h(i,j)=1$ is treated as a redundant one for $\theta_i^*$ and it will be removed from the model hypothesis set $\bm{\theta}^*$ at the next iteration.

Note that the performance of the proposed framework also depends on the quality of model hypotheses, that is, it may not work well for the model hypotheses generated by some conventional sampling algorithms (e.g., \cite{fischler1981random,Magri_2014_CVPR}), which may generate a high percentage of bad and duplicated model hypotheses. In contrast, the model hypotheses generated by the proposed sampling algorithm contain a high percentage of good model hypotheses and there are few duplicated model hypotheses. Thus, the proposed framework is able to effectively remove redundant model hypotheses at each iteration based on the model hypotheses generated by the proposed sampling algorithm. More specifically, after selecting a promising model hypothesis $\theta_i^*$ as one of the estimated model instances, the remaining good model hypotheses in $\bm{\theta}^*$ include two parts: the model hypothesis set $\bm{\hat{\theta}_i^*}$ generated by the ($p+2$) subsets that are the inliers of $\theta_i^*$, and the model hypothesis set $\bm{\vec{\theta}_i^*}$ generated by the ($p+2$) subsets that are the inliers of the remaining model instances in the data. According to Eq.~(\ref{equ:reduant}), the proposed framework not only removes many bad model hypotheses but also $\bm{\hat{\theta}_i^*}$. Thus, the novel ``fit-and-remove" framework can work well based on the proposed sampling algorithm for challenging multiple-structural data.
\section{The complete method}
\label{sec:completealgorithm}
\begin{algorithm}[t] 
\renewcommand{\algorithmicrequire}{\textbf{Input:}}
\renewcommand\algorithmicensure {\textbf{Output:} }
\caption{The superpixel-guided two-view deterministic fitting method} 
\label{alg:completed} 
\begin{algorithmic}[1] 
\REQUIRE 
An image pair, keypoint correspondences and the number of model instances $l$.
\STATE Generate an initial model hypothesis set $\bm{\theta}$ by Algorithm~\ref{alg:initisampling}.
\STATE Based on $\bm{\theta}$, obtain the updated model hypothesis set $\bm{\theta}^*$ by Algorithm~\ref{alg:hypothesisupdating}.
\FOR {$i=1$ to $l$}
\label{state:modelselection}
\STATE Select a model hypothesis $\theta_i^*$ with the highest weighting value from $\bm{\theta}^*$ as one of the estimated model instances.
\STATE Find the redundant model hypotheses $\bm{\vartheta}_i^*$($=\{\theta_i^j\}_{j=1,2,\ldots}$) with respect to $\theta_i^*$ according to Eq.~(\ref{equ:reduant}).
\STATE Remove $\theta_i^*$ and its redundant model hypotheses $\bm{\vartheta}_i^*$ from $\bm{\theta}^*$, i.e., $\bm{\theta}^*\leftarrow \bm{\theta}^*\setminus \{\bm{\vartheta}_i^*\cup \theta_i^*\}$.
\ENDFOR
\label{state:modelselection2}
\ENSURE The estimated model instances $\{\theta_i^*\}_{i=1}^l$.
\end{algorithmic}
\end{algorithm}
With all the ingredients developed in the previous sections, we summarize the complete SDF fitting method in Algorithm~\ref{alg:completed}. SDF is a superpixel-guided fitting method. We employ the SLIC segmentation algorithm \cite{achanta2012slic} to generate superpixels in Algorithm~\ref{alg:initisampling}, since SLIC adheres well to object boundaries in an image with $O(N)$ complexity ($N$ is the number of image pixels) and it is a deterministic method as well. Although SLIC is an effective superpixel segmentation algorithm, the performance of SDF dose not largely depend on the quality of superpixel segmentation. This is because that a model instance in data often corresponds to two or more model hypotheses based on the grouping cues of keypoint correspondences derived from different superpixels, and the model instance can be estimated from these model hypotheses.

The proposed SDF exploits the relationship between keypoint correspondences and superpixels to deterministically estimate model instances in multi-structure data. SDF includes three main parts, i.e., a deterministic sampling algorithm, a model hypothesis updating strategy and a novel ``fit-and-remove" framework. The proposed deterministic sampling algorithm effectively utilizes the prior information of superpixels to generate initial model hypotheses. The proposed model hypothesis updating strategy further improves the quality of the generated model hypotheses by analyzing the residual information. The proposed ``fit-and-remove" framework takes advantage of the updated high-quality model hypotheses to efficiently select optimal model hypotheses as the estimated model instances in data. The three parts are tightly coupled in SDF, and they jointly lead to efficient and effective deterministic fitting results.

The computational complexity of Algorithm~\ref{alg:initisampling} is approximately proportional to $O(N)$. The step of superpixel segmentation consumes the majority of the computational time of Algorithm~\ref{alg:initisampling}. For Algorithm~\ref{alg:hypothesisupdating}, its computational complexity is approximately proportional to $O(tm)$ $\ll O(t_{max}m)$, where $t$ is the number of iterations, $m$ is the number of the generated model hypotheses and $t_{max}$ is fixed to $50$. For the model selection (i.e., step 3 $\sim$ step 7 in Algorithm~\ref{alg:completed}), the steps take much less time than Algorithm~\ref{alg:initisampling} and Algorithm~\ref{alg:hypothesisupdating}. Therefore, the total complexity of the proposed algorithm approximately amounts to $O(N)+O(tm)$.
%


\section{Experiments}
\label{sec:experiments}
In this section, we perform homography and fundamental matrix estimation on both single-structure and multiple-structure data. We compare the proposed SDF with several state-of-the-art model fitting methods\footnote{We download the codes from their websites.}, including PROSAC~\cite{chum2005matching}, T-linkage~\cite{Magri_2014_CVPR}, RansaCov~\cite{Magri_2016_CVPR} and AStar~\cite{chin2015efficient}. PROSAC is evaluated since it also considers feature appearances as SDF. T-linkage is a representative model fitting method that effectively works on multiple-structure data, but it does not work on single-structure data very well. Thus we only use it as a competing method on multiple-structure datasets in Sec.~\ref{sec:multiplestructures}. RansaCov is compared because it is one of the state-of-the-art model fitting methods. AStar is also one of the state-of-the-art methods for deterministic fitting. However, AStar only works on single-structure data and thus we do not evaluate it in Sec.~\ref{sec:multiplestructures}. In addition, we also run RANSAC as a baseline. 

All experiments are run on MS Windows $10$ with Intel Core i$7$-$3770$ CPU $3.4GHz$ and $16GB$ RAM.
The fitting error is computed as follows~\cite{Magri_2014_CVPR,mittal2012generalized}:
\begin{align}
\label{equ:fittingerror}
error=\frac{\#~mislabeled~data~points}{\#~data~points}\times 100\%.
\end{align}

$\textsc{Datasets.}$ We test all the competing methods on three datasets, i.e., the BLOGS dataset\footnote{\scriptsize\url{http://www.cse.usf.edu/~sarkar/BLOGS/}}, the OXford VGG dataset\footnote{\scriptsize\url{http://www.robots.ox.ac.uk/~vgg/data/}} and the AdelaideRMF dataset\footnote{\scriptsize\url{http://cs.adelaide.edu.au/}}. From the datasets, we select $40$ representative image pairs, where the first $20$ image pairs are single-structure data (see Sec.~\ref{sec:singlestructure}), and the other $20$ image pairs are multiple-structure data (see Sec.~\ref{sec:multiplestructures}).
\begin{figure*}[t!]
\centering
\begin{minipage}[t]{.43\textwidth}
  \centering
\centerline{\includegraphics[width=0.98\textwidth]{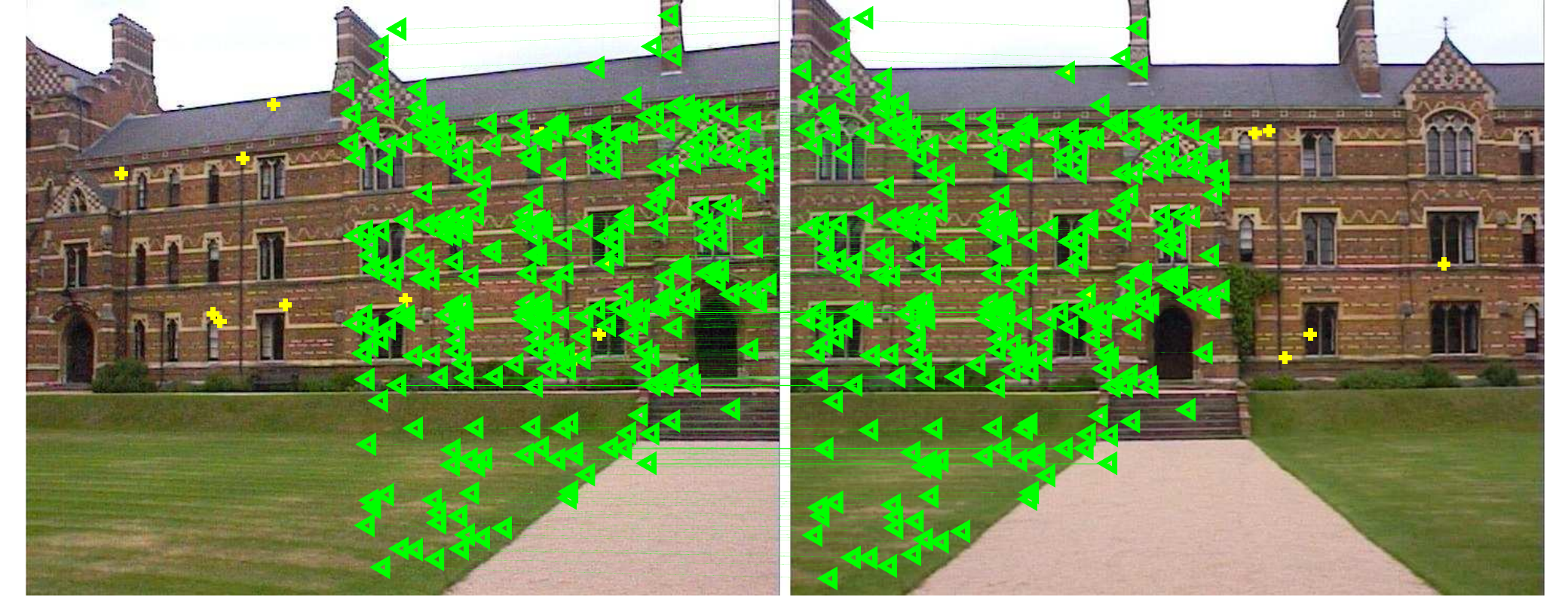}}
  \centerline{(a) Keble}
\centerline{\includegraphics[width=0.98\textwidth]{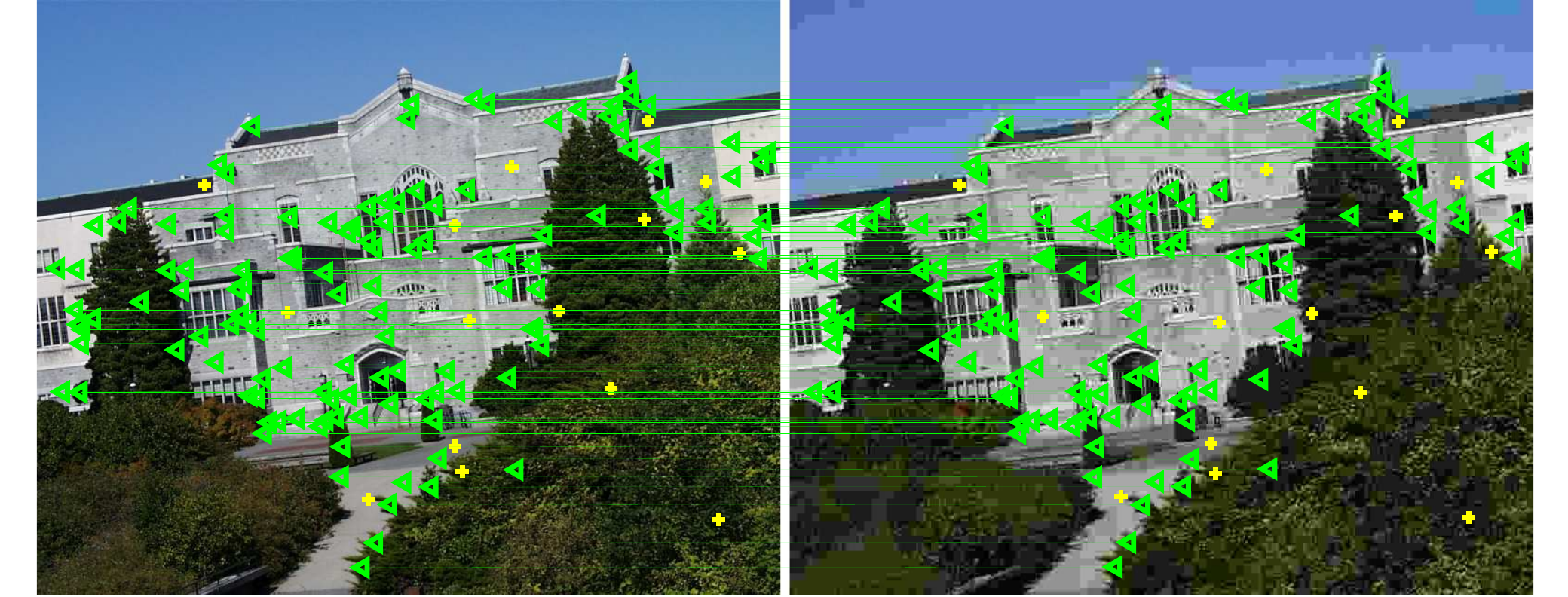}}
  \centerline{(b) UBC}
\centerline{\includegraphics[width=0.98\textwidth]{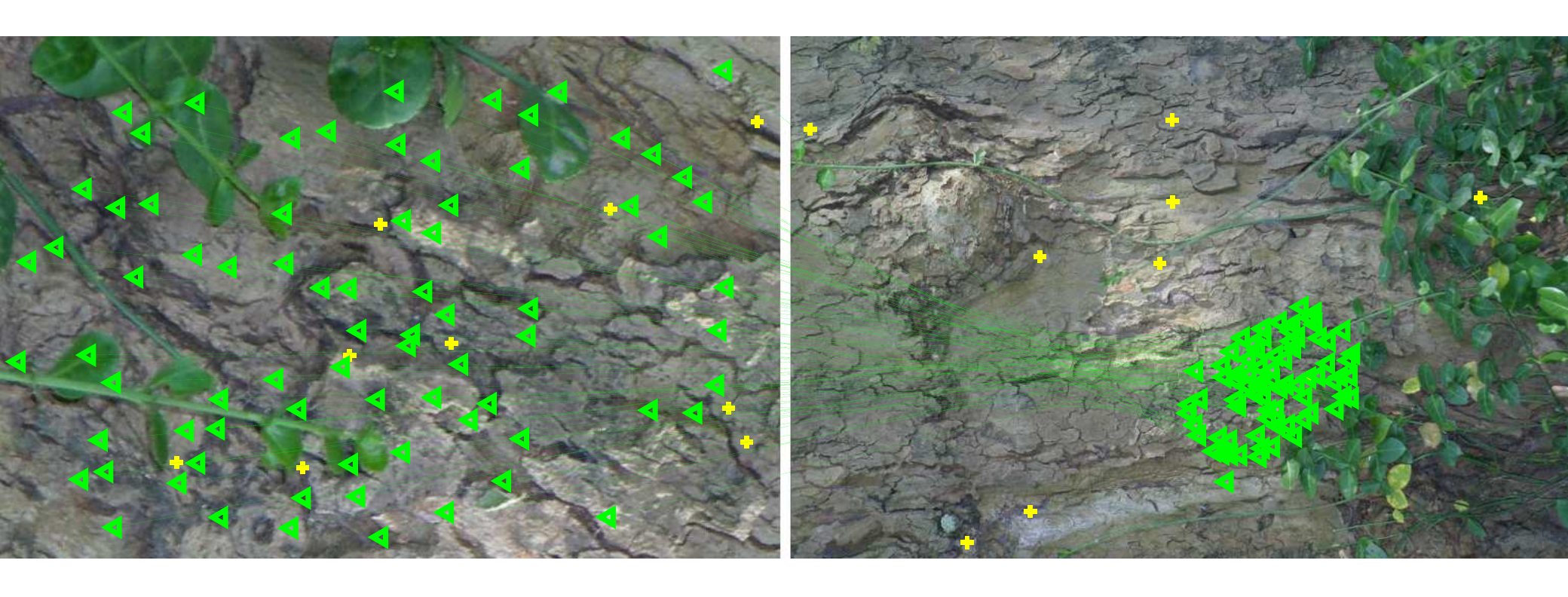}}
  \centerline{(c) Bark}
\centerline{\includegraphics[width=0.98\textwidth]{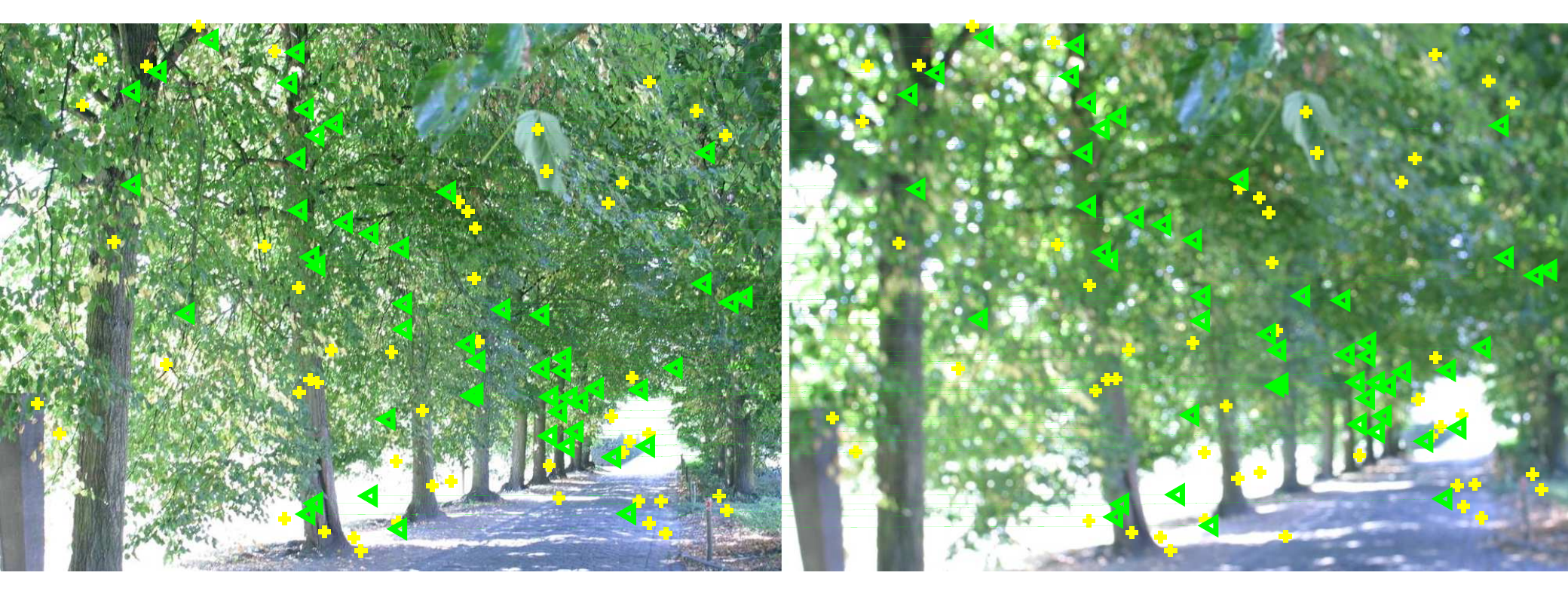}}
  \centerline{(d) Trees}
\centerline{\includegraphics[width=0.98\textwidth]{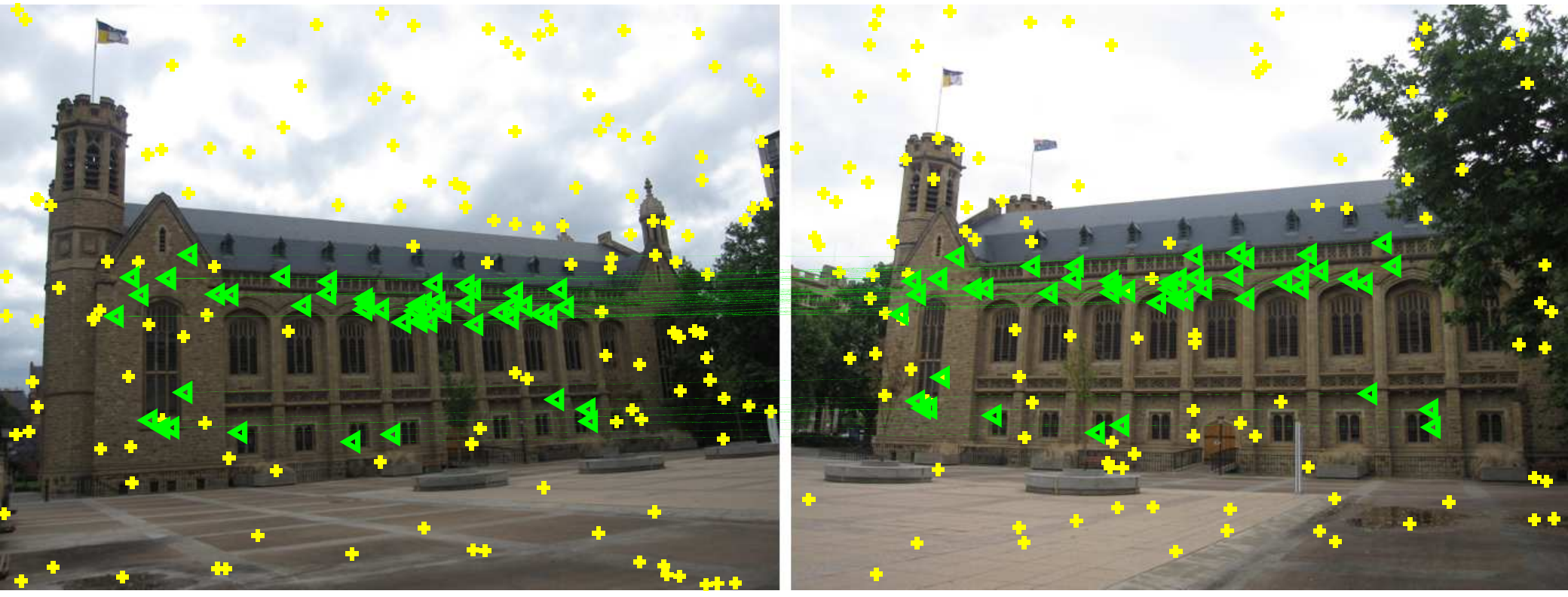}}
  \centerline{(e) Bonython}
\centerline{\includegraphics[width=0.98\textwidth]{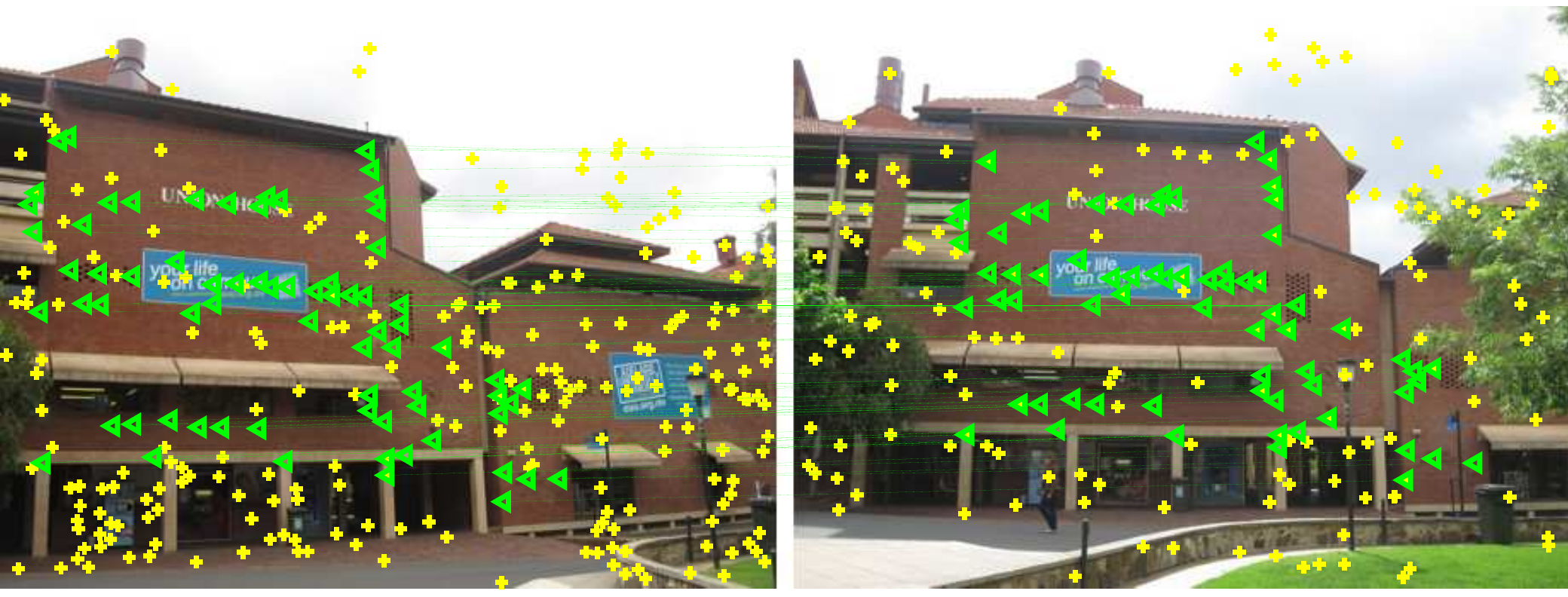}}
  \centerline{(f) Unionhouse}
\end{minipage}
\begin{minipage}{.01\textwidth}
$\qquad$
\end{minipage}
\begin{minipage}[t]{.43\textwidth}
  \centering
\centerline{\includegraphics[width=0.98\textwidth]{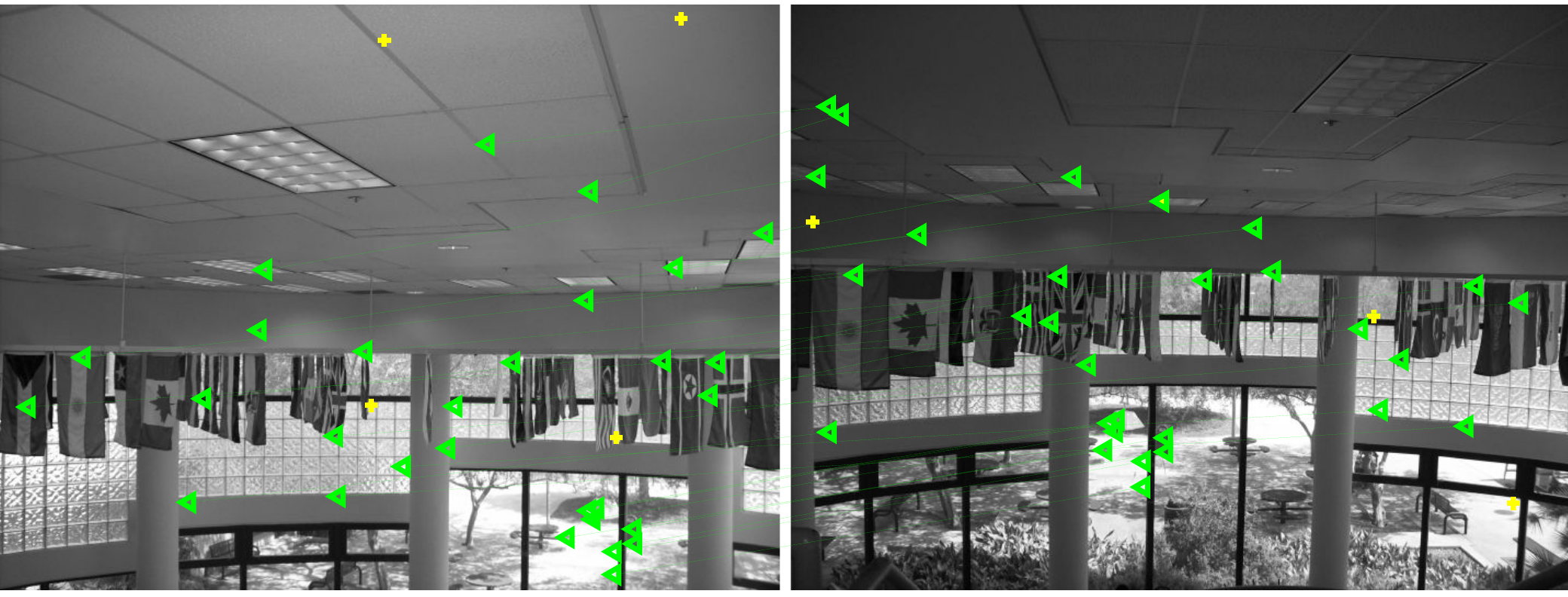}}
  \centerline{(g) Flags}
\centerline{\includegraphics[width=0.98\textwidth]{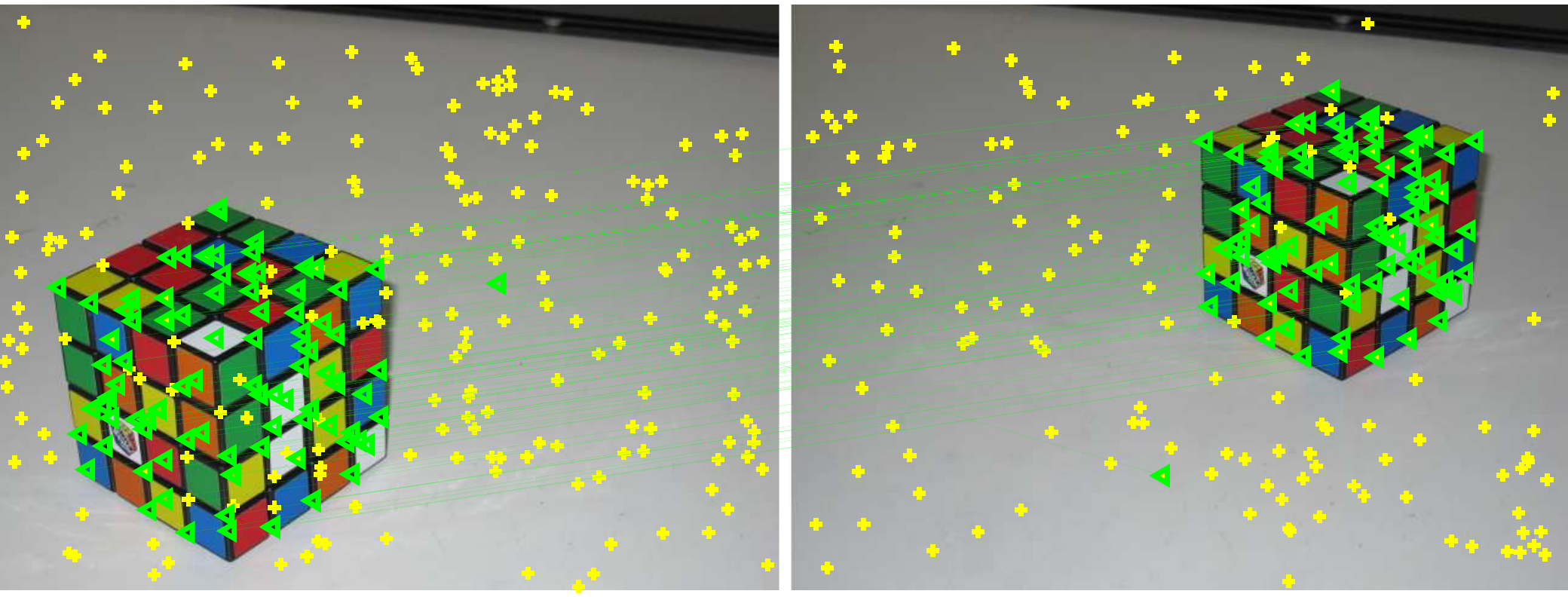}}
  \centerline{(h) Cube}
\centerline{\includegraphics[width=0.98\textwidth]{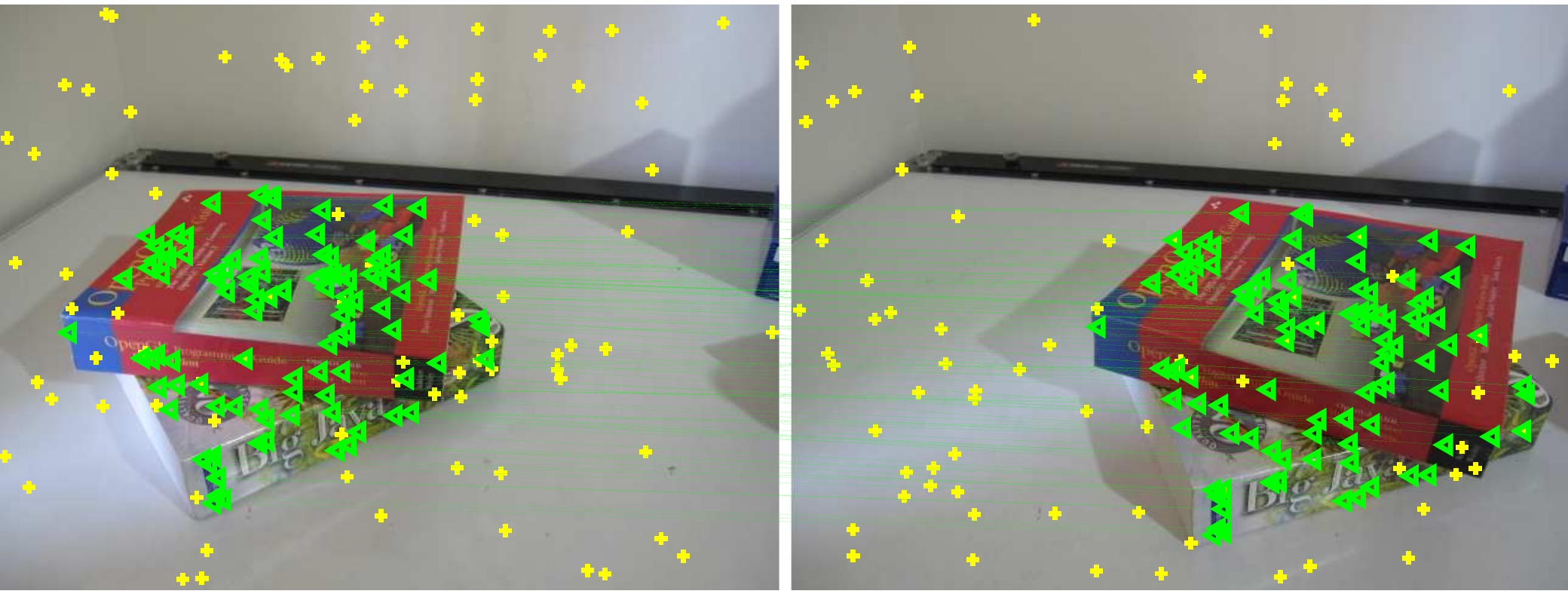}}
  \centerline{(i) Book}
\centerline{\includegraphics[width=0.98\textwidth]{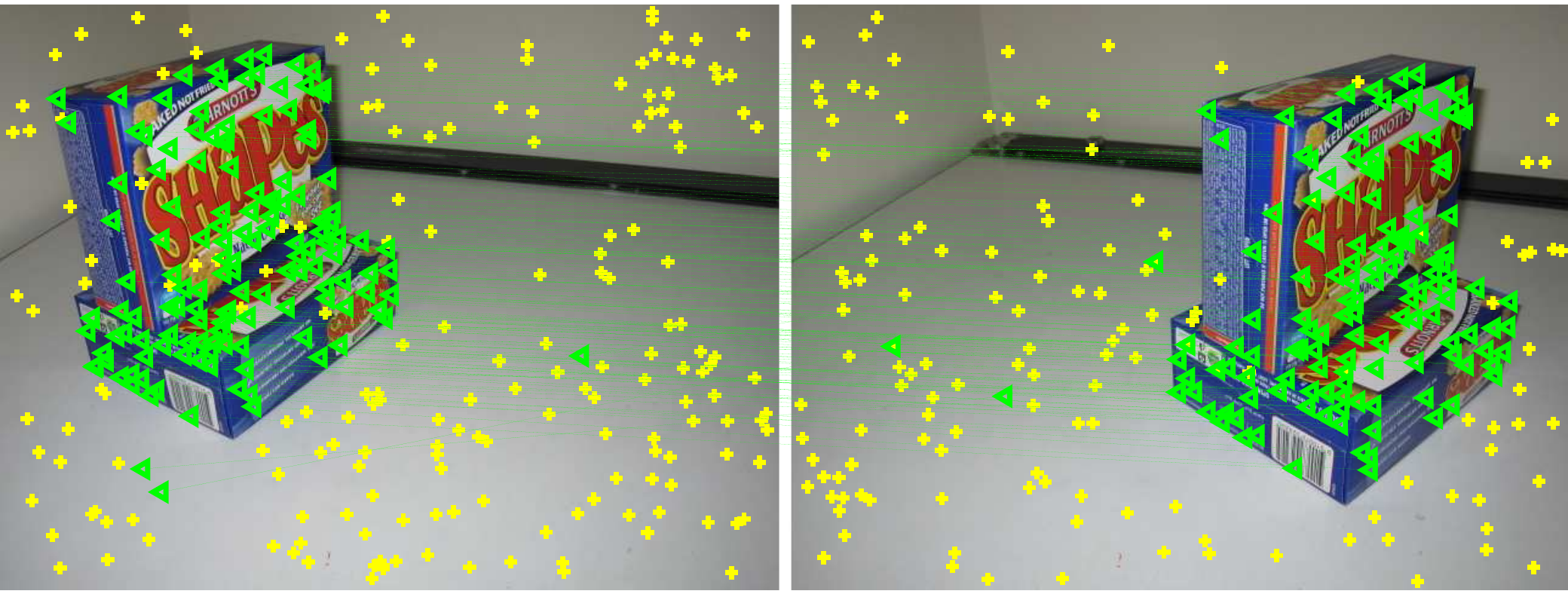}}
  \centerline{(j) Biscuit}
\centerline{\includegraphics[width=0.98\textwidth]{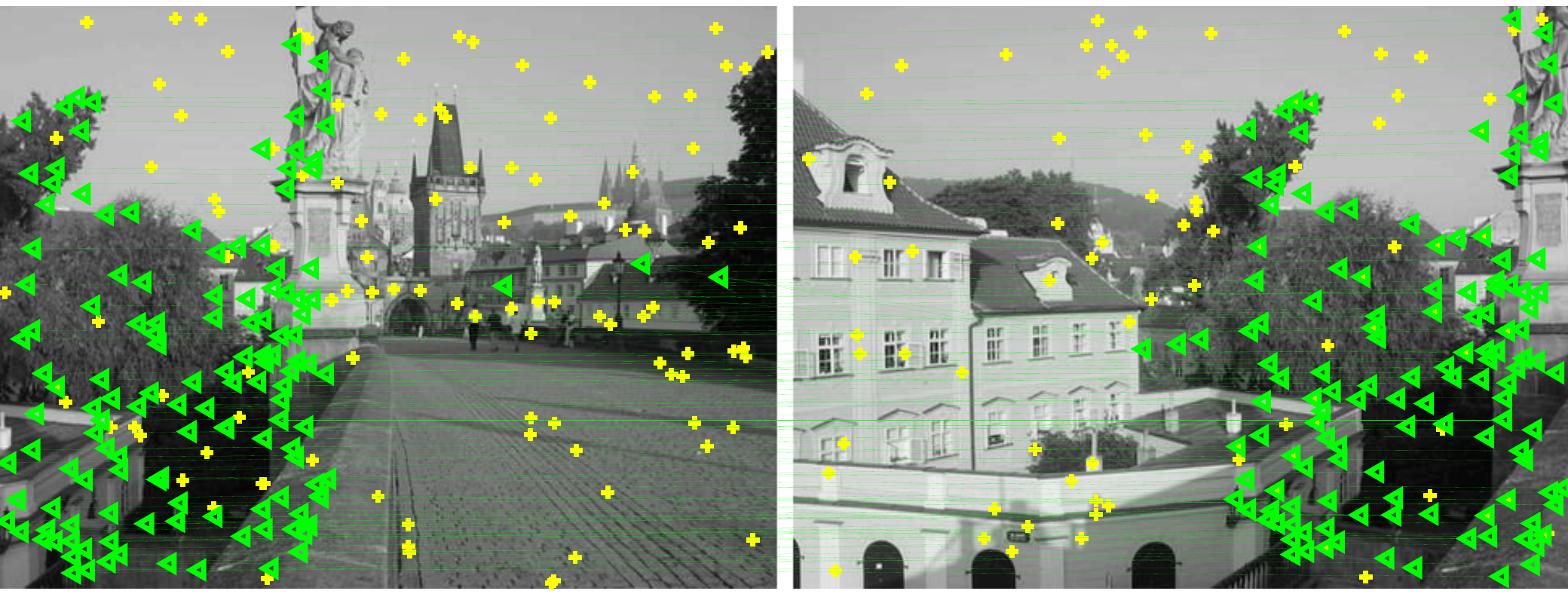}}
  \centerline{(k) Kmsm}
\centerline{\includegraphics[width=0.98\textwidth]{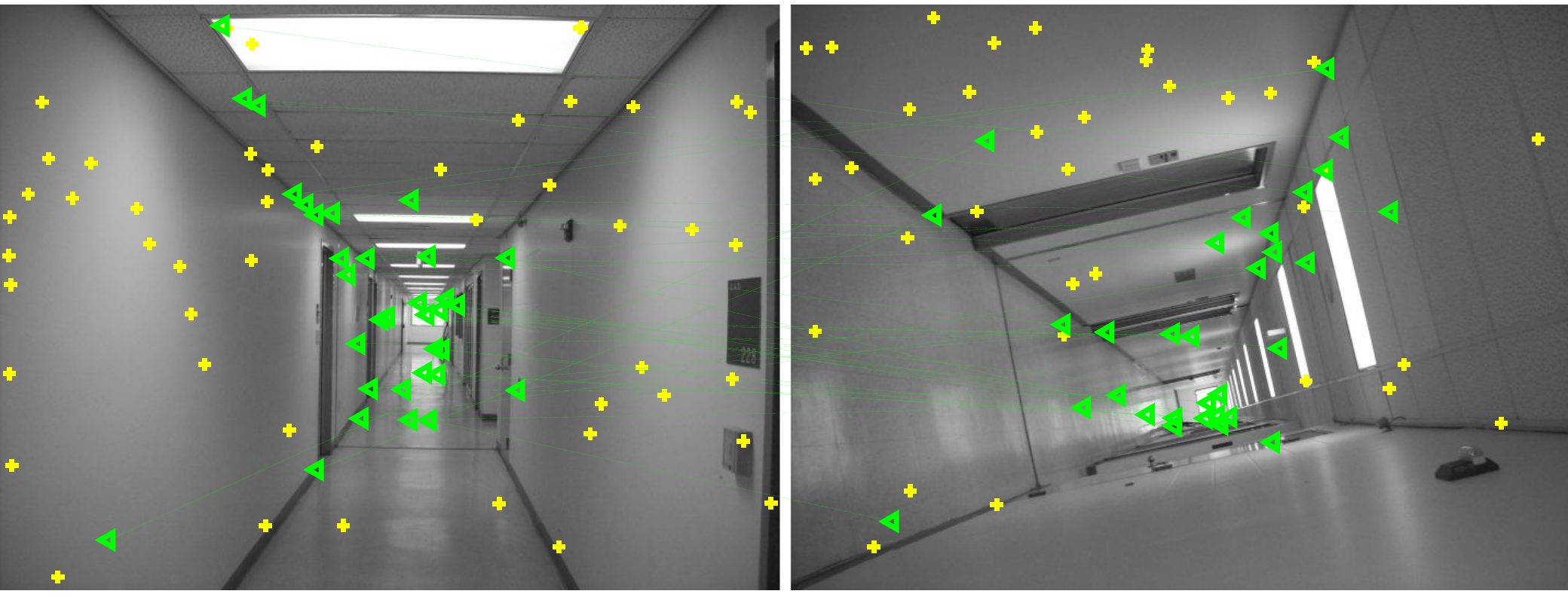}}
  \centerline{(l) Corridor}
\end{minipage}
\caption{Some fitting results obtained by {SDF3} on $12$ image pairs with a single structure. (a) $\sim$ (f) show the results obtained by {SDF3} for homography estimation, and (g) $\sim$ (l) show the results obtained by {SDF3} for fundamental matrix estimation. }
\label{fig:singlestructure}
\end{figure*}
\begin{table}[!t]

\centering
\caption{Quantitative comparison results on single-structure data for homography estimation. }
\scalebox{0.8}{
{\tabcolsep0.02in
\begin{tabular}{llccccccc}
\toprule
Data     &                   & RANSAC&PROSAC&RansaCov&Astar & {SDF1}&SDF2&SDF3\\
\midrule
 &Std.      &{0.42}&{0.97}& {0.01}  &{--}&{--}&{--}&{--}\\
  Keble                  &Avg.      &{6.48}&{6.45}& {\bf2.39}  &{\bf2.39}&{3.07}&{3.07}&{\bf2.39}\\
                      &Time      &{\bf0.01}&{\bf0.01}&35.25&14.73&0.27 &0.46& 0.97\\
            \rowcolor{mygray}
             &Std.      &{4.26}&{3.21}& {0.01}  &{--}&{--}&{--}&{--}\\
             \rowcolor{mygray}
   UBC                  &Avg.     &{9.14}&{9.57}&{\bf4.28}&{\bf4.28}&6.42&{5.00}&{\bf4.28} \\
     \rowcolor{mygray}
                       &Time      &{\bf0.01}&{\bf0.01}&17.64&1.39&0.31&0.51&1.02\\
            &Std.      &{11.34}&{9.81}& {2.31}  &{\#}&{--}&{--}&{--}\\
  Graffiti                 &Avg.      &{14.80}&{23.56}&{9.05}&{\#}&24.05&{16.51}&{\bf4.71} \\
                       &Time      &{\bf0.02}&{\bf0.02}& 15.02&\#&0.35&0.49&0.92\\
         \rowcolor{mygray}
             &Std.      &{0.70}&{7.31}&{0.01}  &{--}&{--}&{--}&{--}\\
             \rowcolor{mygray}
  Bark              &Avg.     &{0.23}&{3.23}&{\bf0.00}&{\bf0.00}&{\bf0.00}&{\bf0.00}&{\bf0.00}\\
     \rowcolor{mygray}
                    &Time      &{\bf0.01}&{\bf0.01}&20.95&5.65&0.25&0.26&0.61\\
             &Std.      &{6.07}&{5.89}& {4.37}  &{\#}&{--}&{--}&{--}\\
  Trees    &Avg.      &{21.96}&{23.46}&{17.62}&{\#}&37.62&{31.68}&{\bf11.88}\\
            &Time      &{\bf0.11}&{\bf0.11}&13.01&{\#}&0.46&0.47&0.82\\
          \rowcolor{mygray}
             &Std.      &{0.49}&{0.35}& {0.13}  &{\#}&{--}&{--}&{--}\\
             \rowcolor{mygray}
    MCI     &Avg.     &{2.53}&{1.06}&{1.45}&{\#}&5.43&{3.12}&{\bf0.46}\\
    \rowcolor{mygray}
                                    &Time      &{\bf0.03}&{\bf0.03}&77.54&{\#}&0.60&0.86&1.62\\
              &Std.      &{1.10}&{1.11}& {0.70}  &{\#}&{--}&{--}&{--}\\
    MCIII      &Avg.      &3.73&1.43&2.41&{\#}&2.56&2.13&{\bf 0.00}\\
                                    &Time      &{\bf0.05}&0.06&57.20&{\#}&0.59&0.76&1.38\\
           \rowcolor{mygray}
               &Std.      &{2.91}&{2.43}& {0.01} &{\#}&{--}&{--}&{--}\\
               \rowcolor{mygray}
Bonython            &Avg.       &{1.51}&{5.60}&{\bf0.00}&{\#}&5.55&{3.53}&{\bf0.00}\\
  \rowcolor{mygray}
                         &Time      &0.88&{4.30}&17.49&{\#}&{\bf0.26}&{0.42}&0.82\\
            &Std.      &{3.24}&{6.46}& {0.29}  &{\#}&{--}&{--}&{--}\\
     Physics   &Avg.       &{3.20}&{5.00}&{2.92}&{\#}&{2.83}&{2.83}&{\bf0.00}\\
                         &Time      &{\bf0.05} &0.18&17.39&{\#}&0.22&0.33&0.69\\
           \rowcolor{mygray}
            &Std.      &{14.64}&{10.04}& {0.01} &{\#}&{--}&{--}&{--}\\
            \rowcolor{mygray}
    Unionhouse                   &Avg.        &{5.00}&{4.66}&{\bf0.30}&{\#}&42.77&{38.55}&{\bf0.30} \\
     \rowcolor{mygray}
                         &Time      &1.14&{4.49} & 17.93&{\#}&{\bf0.15}&{0.32}&0.83\\
         \bottomrule
\end{tabular}}}
\\
\medskip
\raggedright
(`\#' denotes that the fitting methods do not obtain a solution within $1$ hour. '--' denotes the results obtained by the deterministic fitting methods, which are not implemented repeatedly. The best results are boldfaced.)
 \label{table:homographytable}
\end{table}
\begin{table}[!t]
\caption{Quantitative comparison results on single-structure data for fundamental matrix estimation. }
\centering
\scalebox{0.8}{
{\tabcolsep0.03in
\begin{tabular}{llccccccc}
\toprule
Data     &                  & RANSAC&PROSAC&RansaCov&Astar &{SDF1}&SDF2&SDF3\\
\midrule
 &Std.      &{5.05}&{4.16}& {4.09}  &{--}&{--}&{--}&{--}\\
Twocars                     &Avg.      &{8.98}&{7.50}& {2.65}  &{\bf1.56}&4.68&{\bf1.56}&{\bf1.56}\\
                       &Time      &{\bf0.01}&{\bf0.01}&21.25&20.34&2.21&2.25&2.51\\
           \rowcolor{mygray}
             &Std.      &{4.20}&{5.99}& {0.01}  &{--}&{--}&{--}&{--}\\
              \rowcolor{mygray}
    Library                 &Avg.     &{18.51}&{14.81}&{5.92}&{\bf1.88}&{5.66}&{5.66}&{\bf1.88} \\
      \rowcolor{mygray}
                       &Time       &{\bf0.01}&{\bf0.01}&3.82&21.34&2.12&2.23&2.56\\
            &Std.      &{0.95}&{6.51}& {0.01}  &{--}&{--}&{--}&{--}\\
Flags                 &Avg.      &{0.30}&{5.45}&{12.12}&{\bf0.00}&6.06&{3.03}&{\bf0.00} \\
                       &Time      &{\bf0.01}&{0.02}&4.09&213.23& 0.09&0.10&0.31\\
            \rowcolor{mygray}
             &Std.      &{0.83}&{0.55}& {0.63}  &{\#}&{--}&{--}&{--}\\
              \rowcolor{mygray}
Cube               &Avg.     &{5.16}&{2.84}&{4.23}&{\#}&7.28&8.94&{\bf1.32}\\
      \rowcolor{mygray}
                    &Time      &7.95&{6.43}&38.76&\#& {\bf0.25}&0.29&0.80\\
             &Std.      &{0.77}&{1.95}& {0.88}  &\#&{--}&{--}&{--}\\
Book               &Avg.      &{2.19}&{3.52}&{2.62}&\#&3.20&{2.67}&{\bf0.53}\\
                    &Time      &0.34&{0.52}&36.39&\#&{\bf0.20}&{0.23}&0.65\\
            \rowcolor{mygray}
             &Std.      &{0.40}&0.73& {0.55}  &{\#}&{--}&{--}&{--}\\
              \rowcolor{mygray}
Biscuit     &Avg.     &{1.33}&{1.84}&{2.21}&{\#}&{5.45}&{5.45}&{\bf0.90}\\
     \rowcolor{mygray}
                                    &Time      &2.68&2.83&46.19&\#&{\bf0.28}&{0.34}&0.78\\
              &Std.      &{0.80}&{1.17}& {0.38}  &{\#}&{--}&{--}&{--}\\
   Kmsm     &Avg.      &4.04&3.22&5.18&\#&3.67&3.67&{\bf 1.63}\\
                                    &Time      &0.40&0.53&19.66&\#&{\bf0.10}&0.14&0.47\\
           \rowcolor{mygray}
               &Std.      &{0.15}&{0.58}& {0.98}  &{\#}&{--}&{--}&{--}\\
                \rowcolor{mygray}
   Path                     &Avg.       &{5.55}&{5.60}&{5.38}&{\#}&3.39&{\bf1.98}&{\bf1.98}\\
     \rowcolor{mygray}
                         &Time      &108.48&{252.60}&49.62&\#&{\bf0.28}&0.31&0.79\\

            &Std.      &{6.10}&{5.01}& {6.45}  &{\#}&{--}&{--}&{--}\\
Corridor                    &Avg.        &{22.59}&{21.72}&{48.14}&{\#}&11.11&{9.87}&{\bf3.70} \\
                         &Time      &1.17&{1.09} & 6.38&\#& {\bf0.10}&0.13&0.36\\
            \rowcolor{mygray}
            &Std.      &{4.61}&{3.95}& {0.01}  &{\#}&{--}&{--}&{--}\\
             \rowcolor{mygray}
    Parking  &Avg.       &{17.93}&{17.46}&{30.15}&{\#}&11.11&{9.52}&{\bf7.93}\\
      \rowcolor{mygray}
                         &Time      &21.81&{4.52}&5.50&\#&{\bf0.12}&0.14&0.34\\
   \bottomrule
\end{tabular}}}
\\
\medskip
\raggedright
(`\#' denotes that the fitting methods do not obtain a solution within $1$ hour. '--' denotes the results obtained by the deterministic fitting methods, which are not implemented repeatedly. The best results are boldfaced.)
 \label{table:fundtable}
\end{table}

\subsection{Single-structure Data}
\label{sec:singlestructure}
{To evaluate the improvements of the proposed SDF in this paper over the original version of SDF~\cite{ECCVXiao2016}, we test three versions of SDF: SDF1 (the original version), SDF2 which uses the improved model selection algorithm and SDF3 which uses both the improved model selection algorithm and the model hypothesis updating strategy.}
Then, we compare the performance of the {seven} competing fitting methods (i.e., RANSAC, PROSAC, RansaCov, Astar and the proposed {SDF1/SDF2/SDF3}) on the 20 image pairs with single-structure data for homography and fundamental matrix estimation. We report the standard variances, the average fitting errors and the average computational time (i.e., the CPU time used in seconds)  in Table~\ref{table:homographytable} and Table~\ref{table:fundtable} for homography and fundamental matrix estimation, respectively (we only show the results obtained within $1$ hour as \cite{chin2015efficient}). For RANSAC, PROSAC and RansaCov, we show the average results (we repeat the experiments $50$ times) due to their {randomness}. For Astar and {SDF1/SDF2/SDF3}, we do not repeat the experiments due to their deterministic nature. Some fitting results obtained by SDF3 are also shown in Fig.~\ref{fig:singlestructure} (we do not show the fitting results obtained by the other competing methods due to the limit of space).

$\textsc{Homography~Estimation.}$ From Fig.~\ref{fig:singlestructure}(a) $\sim$ \ref{fig:singlestructure}(f) and Table~\ref{table:homographytable}, we can see that SDF3 achieves good fitting results on image pairs with a single structure for homography estimation, i.e., it achieves the lowest average fitting errors for all the $10$ image pairs, and it obtains the results within $2$ seconds for all the $10$ image pairs. For the standard variances, the deterministic fitting methods (i.e., Astar and {SDF1/SDF2/SDF3}) show significant superiority over random fitting methods (i.e., RANSAC, PROSAC and RansaCov). For the fitting errors, RansaCov and SDF3 achieve good performance, where they achieve the lowest average fitting errors for $5$ and $10$ out of the $10$ image pairs, respectively. {SDF1/SDF2 achieve low average fitting errors for most of the $10$ image pairs, but both fail in the image pairs with the wide spatial distribution (i.e., ``Graffiti", ``Trees" and ``Unionhouse"). Astar achieves the lowest average fitting errors for the image pairs with low outlier percentages (i.e., ``Keble", ``Ubc" and ``Graffiti")}. RANSAC and PROSAC achieve the worst performance among all the seven competing fitting methods for most image pairs due to their instability. For the computational time, RANSAC, PROSAC and {SDF1/SDF2/SDF3} quickly yield the fitting results for all the $10$ image pairs. In contrast, RansaCov and Astar are slower than the other competing methods. RansaCov includes an efficient model selection algorithm, but most of its computational time is used to sample minimal subsets of data points for generating model hypotheses. Astar takes much time to find globally optimal solutions and it cannot yield fitting results for the image pairs with a high percentage of outliers within $1$ hour.

$\textsc{Fundamental~Matrix~Estimation.}$ From Fig.~\ref{fig:singlestructure}(g) $\sim$ \ref{fig:singlestructure}(l) and Table~\ref{table:fundtable}, we can see that SDF3 also achieves good fitting results on the image pairs with a single structure for fundamental matrix estimation, i.e., it achieves the lowest average fitting errors for all the $10$ image pairs, and it obtains the results within $3$ seconds for all the $10$ image pairs. For the fitting errors, SDF3 achieves much better than the other competing methods, especially for the image pairs with high outlier percentages (i.e., ``Corridor" and ``Parking"). {SDF1/SDF2 achieve} low fitting errors for most of the $10$ image pairs due to their effectiveness, but both achieve the worst fitting errors for two image pairs (i.e., ``Cube" and ``Biscuit"). This is because {SDF1/SDF2 may sample} subsets with small spans to generate model hypotheses. Astar achieves reliable fitting errors for the image pairs with high inlier percentages. RansaCov achieves low fitting errors for $8$ out of the $10$ image pairs but it achieves the worst fitting errors for three image pairs (i.e., ``Flag", ``Corridor" and ``Parking"). For the image pairs with a single structure, RansaCov does not show its effectiveness for segmenting data points belonging to different model instances, and its performance largely depends on the quality of the generated model hypotheses. RANSAC and PROSAC do not achieve stable fitting errors due to their {randomness}, especially for the image pairs with high outlier percentages (i.e., ``Corridor" and ``Parking"). For the computational time, {SDF1/SDF2/SDF3} achieve good performance for all the $10$ image pairs. Astar is computationally inefficient and it does not yield solutions within $1$ hour for $7$ out of the $10$ image pairs. RansaCov is slower than RANSAC and PROSAC for most image pairs because we use the proximity sampling \cite{Magri_2014_CVPR} (which is slower than RANSAC and PROSAC for most cases) for RansaCov.

\begin{figure}[h!]
\centering
\begin{minipage}[t]{.23\textwidth}
  \centerline{\includegraphics[width=1.06\textwidth]{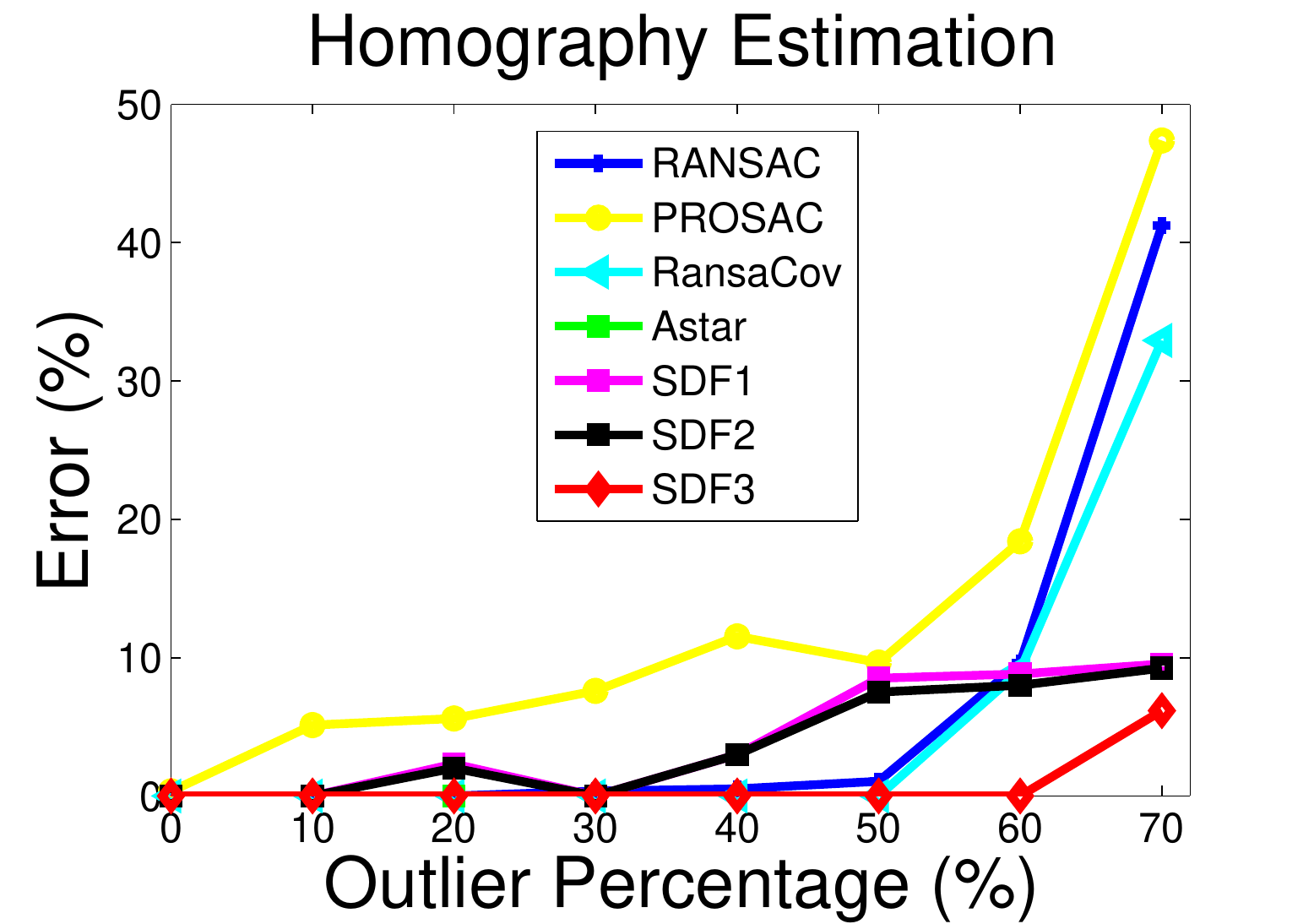}}
 \centerline{(a)}
  \centerline{\includegraphics[width=1.06\textwidth]{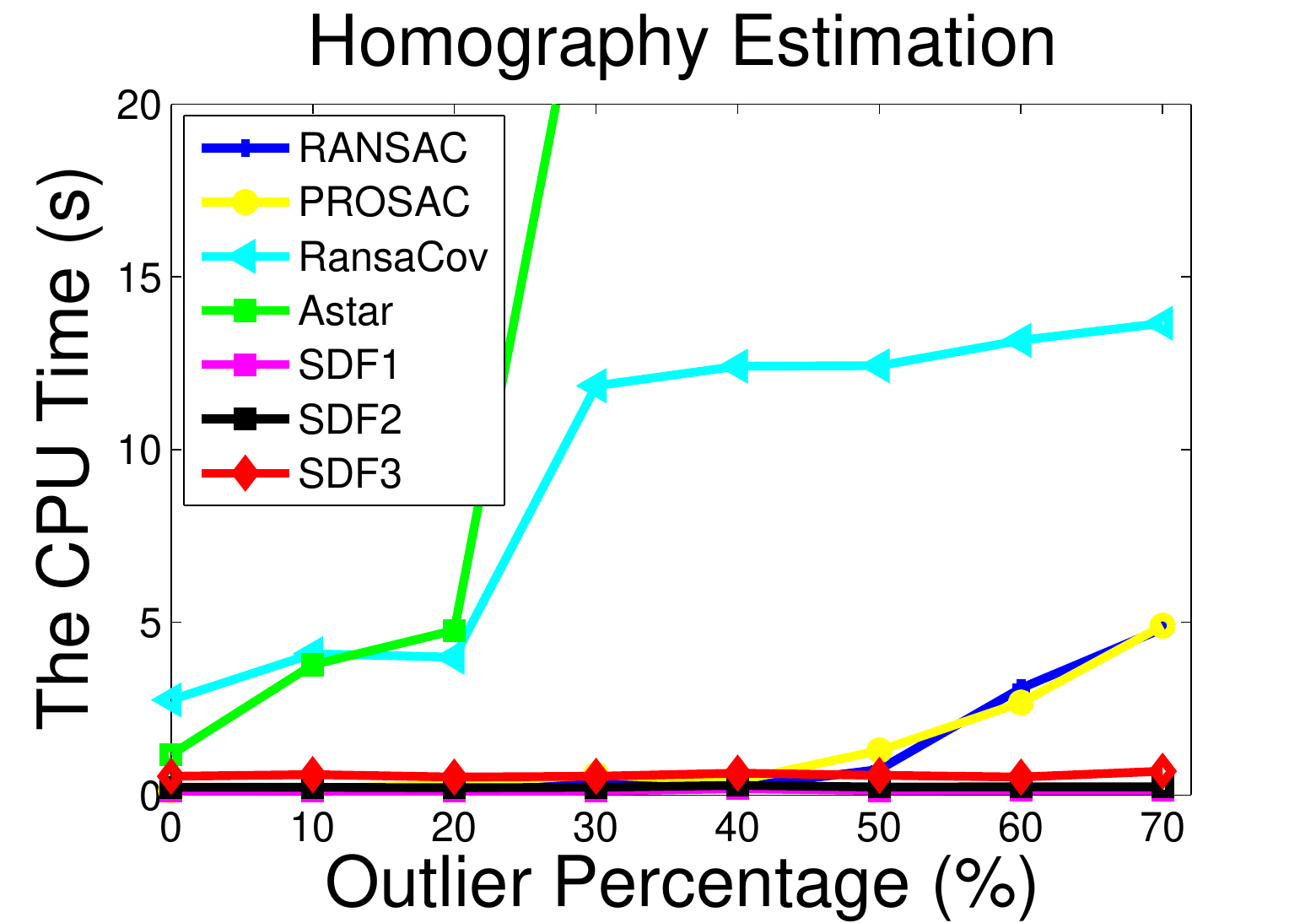}}
  \centerline {(c)}
\end{minipage}
\begin{minipage}[t]{.23\textwidth}
  \centerline{\includegraphics[width=1.06\textwidth]{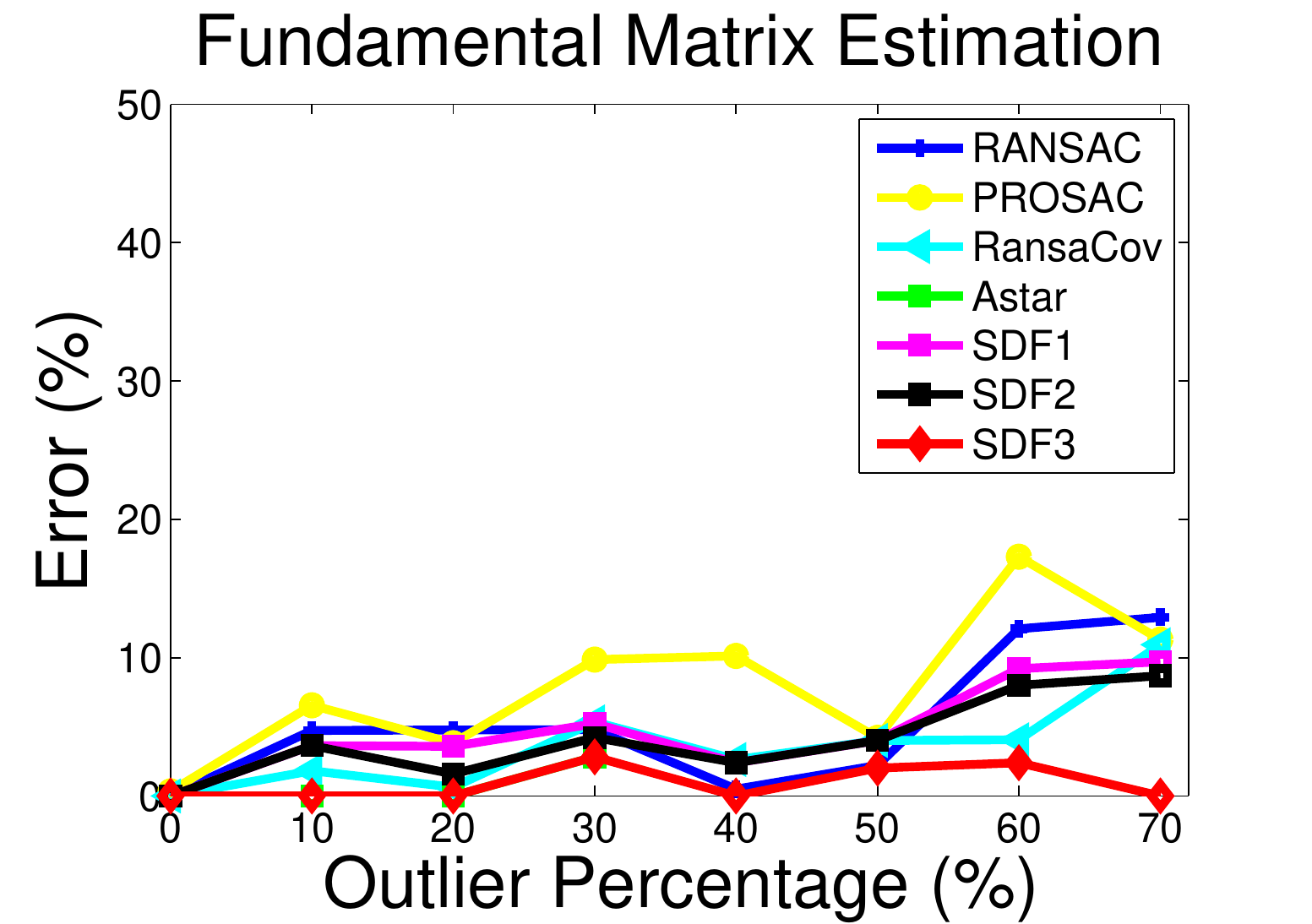}}
 \centerline{(b)}
  \centerline{\includegraphics[width=1.06\textwidth]{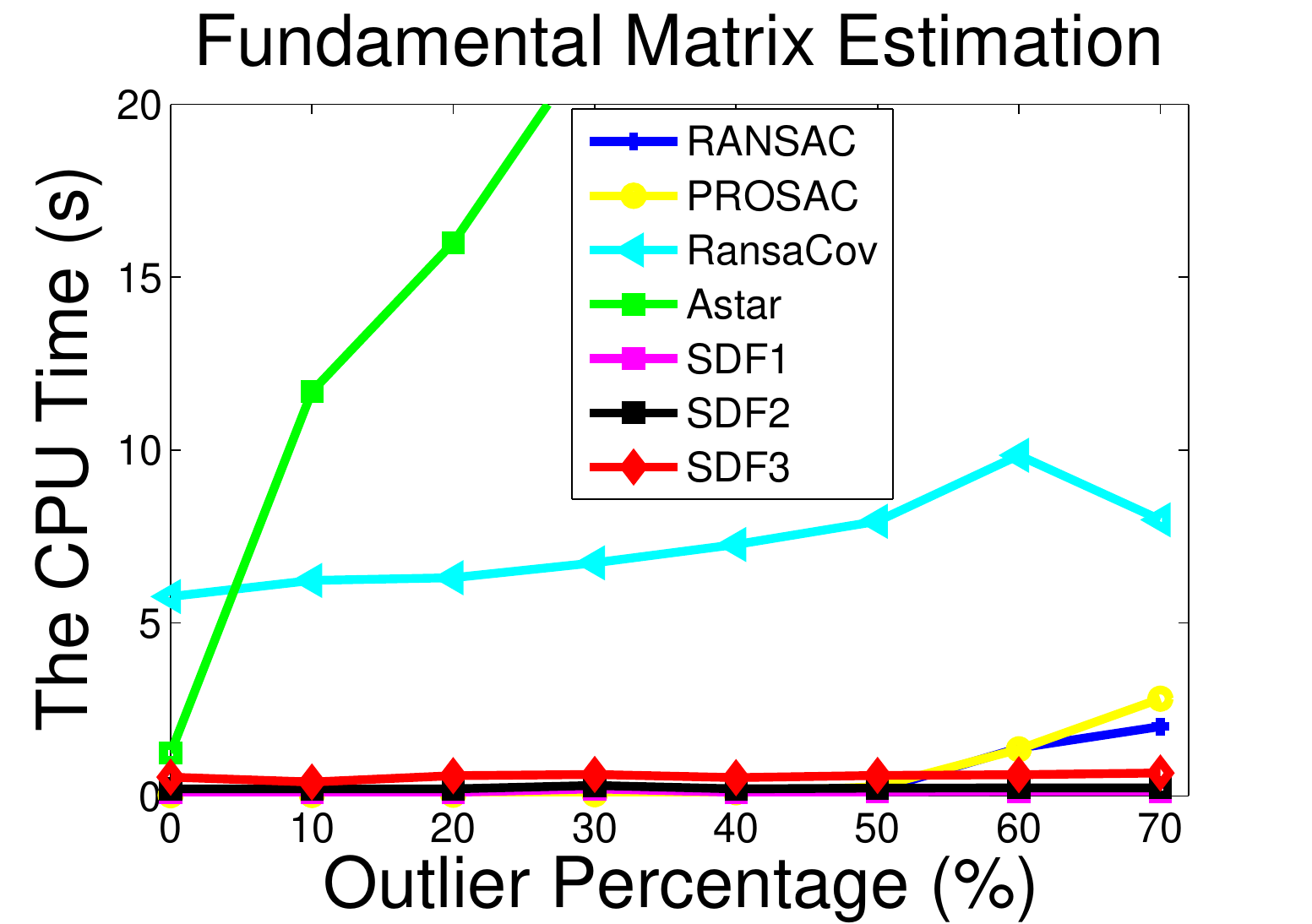}}
 \centerline {(d)}
\end{minipage}
\caption{The fitting errors and computational time of {the seven competing methods} on two image pairs with different outlier percentages: (a) and (c) show the performance comparison on the ``Physics'' image pair for homography estimation; (b) and (d) show the performance comparison on the ``Book" image pair for fundamental matrix estimation.}
\label{fig:differentoutliers}
\end{figure}
$\textsc{Influence~of~Outlier~Percentages.}$ We also evaluate the performance of all the seven fitting methods on image pairs with different outlier percentages. As shown in Fig.~\ref{fig:differentoutliers}, we report the fitting errors and computational time of the seven competing methods on two image pairs with different outlier percentages. We can see that, for the fitting error, SDF3 significantly outperforms the other six competing methods when the outlier percentage is larger than $50\%$. {SDF1/SDF2} achieve low fitting errors for the image pairs with different outlier percentages. Astar achieves low fitting errors within $1$ hour when the outlier percentage is smaller than $30\%$.
RANSAC, PROSAC and RansaCov work well when the outlier percentage is small, but they break down when the outlier percentage is larger than 60\% for homography estimation.

The CPU time used by Astar is much higher than that used by the other six competing methods, especially when the outlier percentage is larger than $30\%$. The CPU time used by RansaCov is higher than that used by RANSAC, PROSAC and {SDF1/SDF2/SDF3}. The CPU time used by RANSAC and PROSAC increases substantially when the outlier percentage is larger than $40\%$ (for homography estimation)  and $50\%$ (for fundamental matrix estimation). This is because they suffer from the influence of outliers during the process of sampling all-inlier subsets. In contrast, the CPU time used by {SDF1/SDF2/SDF3} has no significant change when the outlier percentage increases, which shows the robustness of {SDF1/SDF2/SDF3} to outliers with regard to the CPU time.

For the performance obtained by the three different versions of SDF, we can see that, SDF1 is faster than SDF2/SDF3 in most cases since that SDF1 directly computes the number of inliers to select significant model hypotheses instead of using weighting scores, and it does not employ the model hypothesis updating strategy. However, SDF2/SDF3 achieve lower fitting errors than SDF1, and SDF3 shows significant superiority over SDF1/SDF2, from which we can see that the model hypothesis updating strategy used in SDF3 plays an important role in the proposed method. For this reason, in next subsection, we only evaluate SDF3 which uses both the improved model selection algorithm and the model hypothesis updating strategy, on multiple-structure data.

\begin{figure*}[t!]
\begin{minipage}{.19\textwidth}
\centerline{\includegraphics[width=1.06\textwidth]{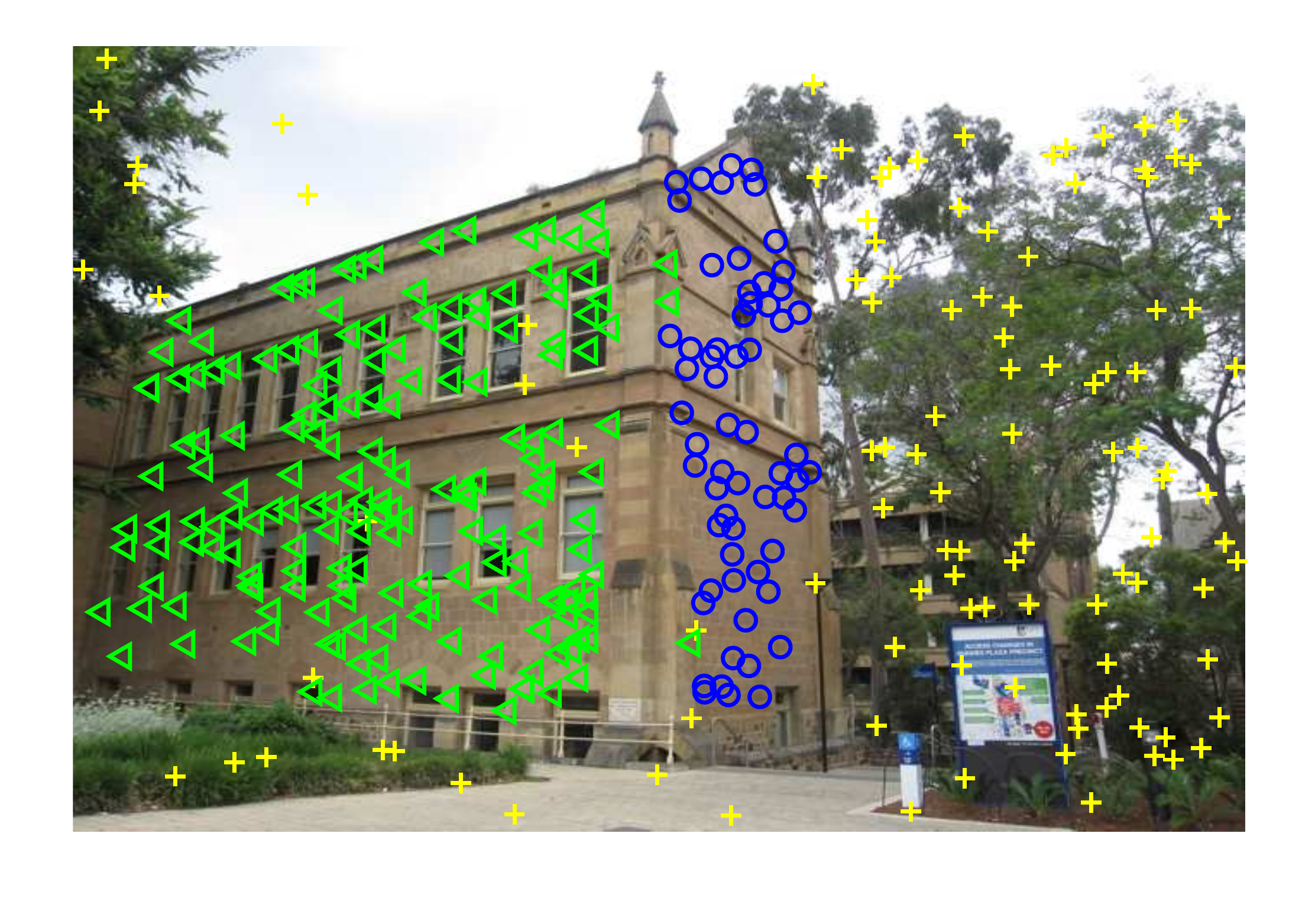}}
  \centerline{(a) Oldclassicswing}
  \centerline{\includegraphics[width=1.06\textwidth]{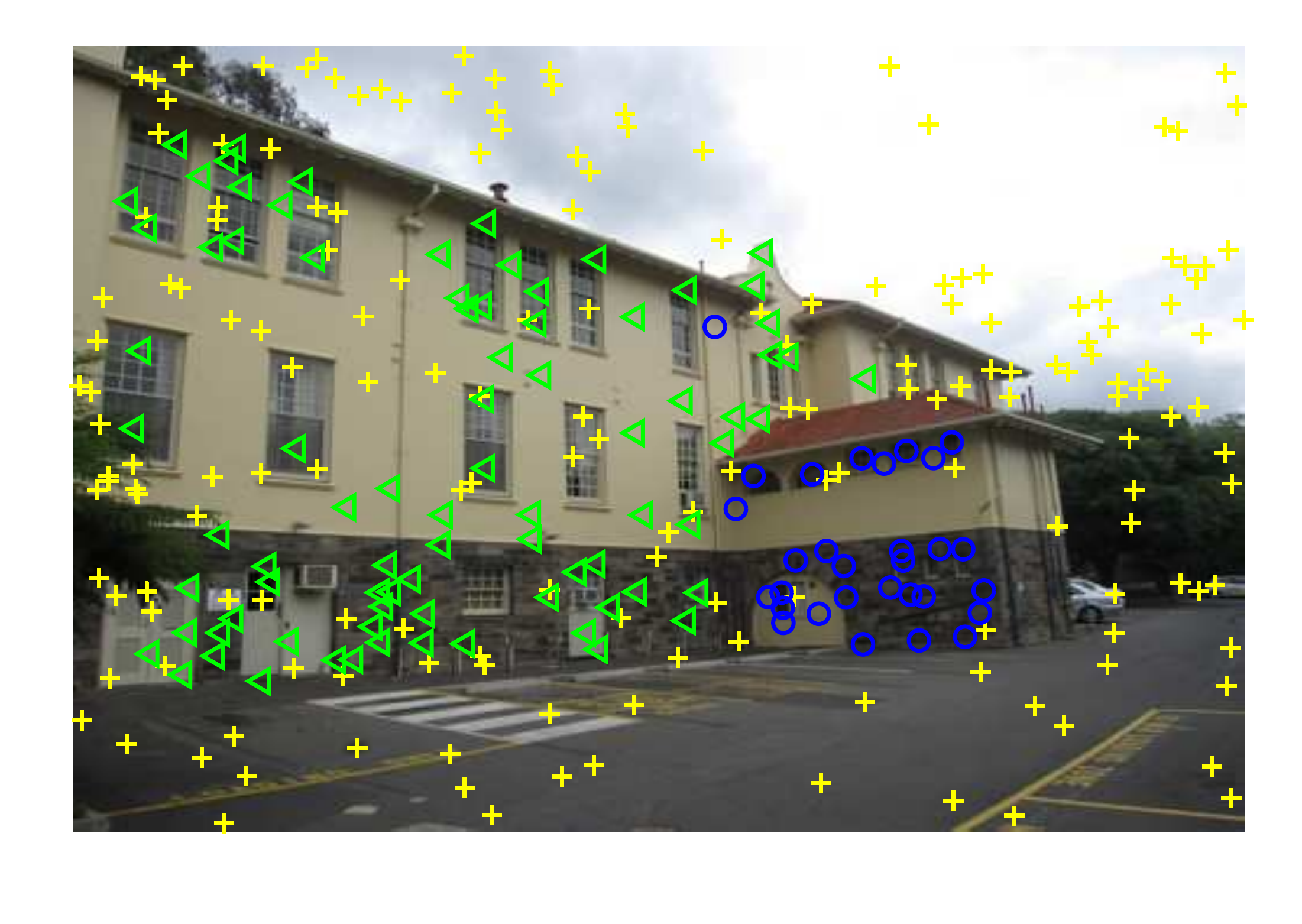}}
  \centerline{(f) Hartley}
\end{minipage}
\begin{minipage}{.19\textwidth}
\centerline{\includegraphics[width=1.06\textwidth]{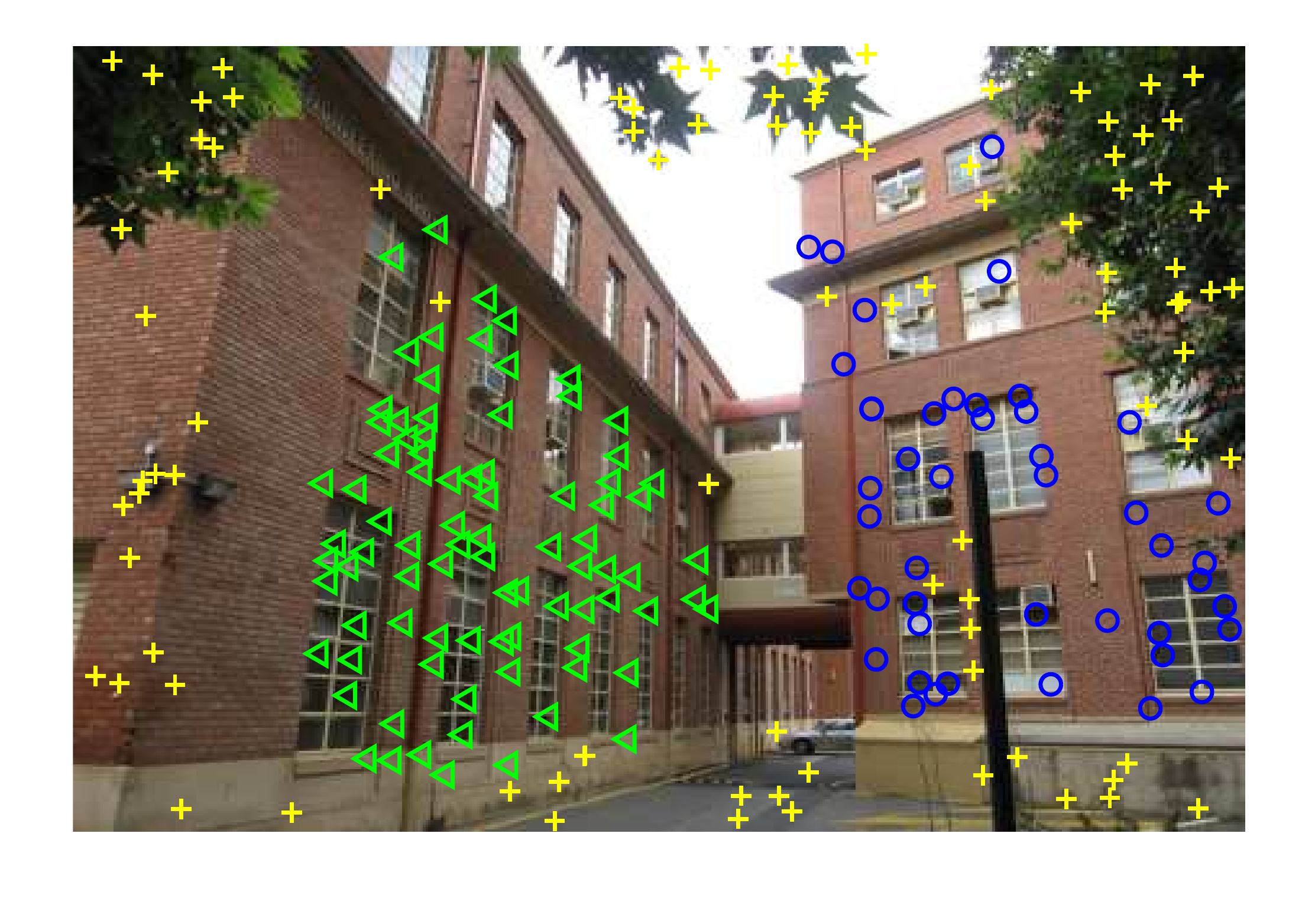}}
  \centerline{(b) Sene}
  \centerline{\includegraphics[width=1.06\textwidth]{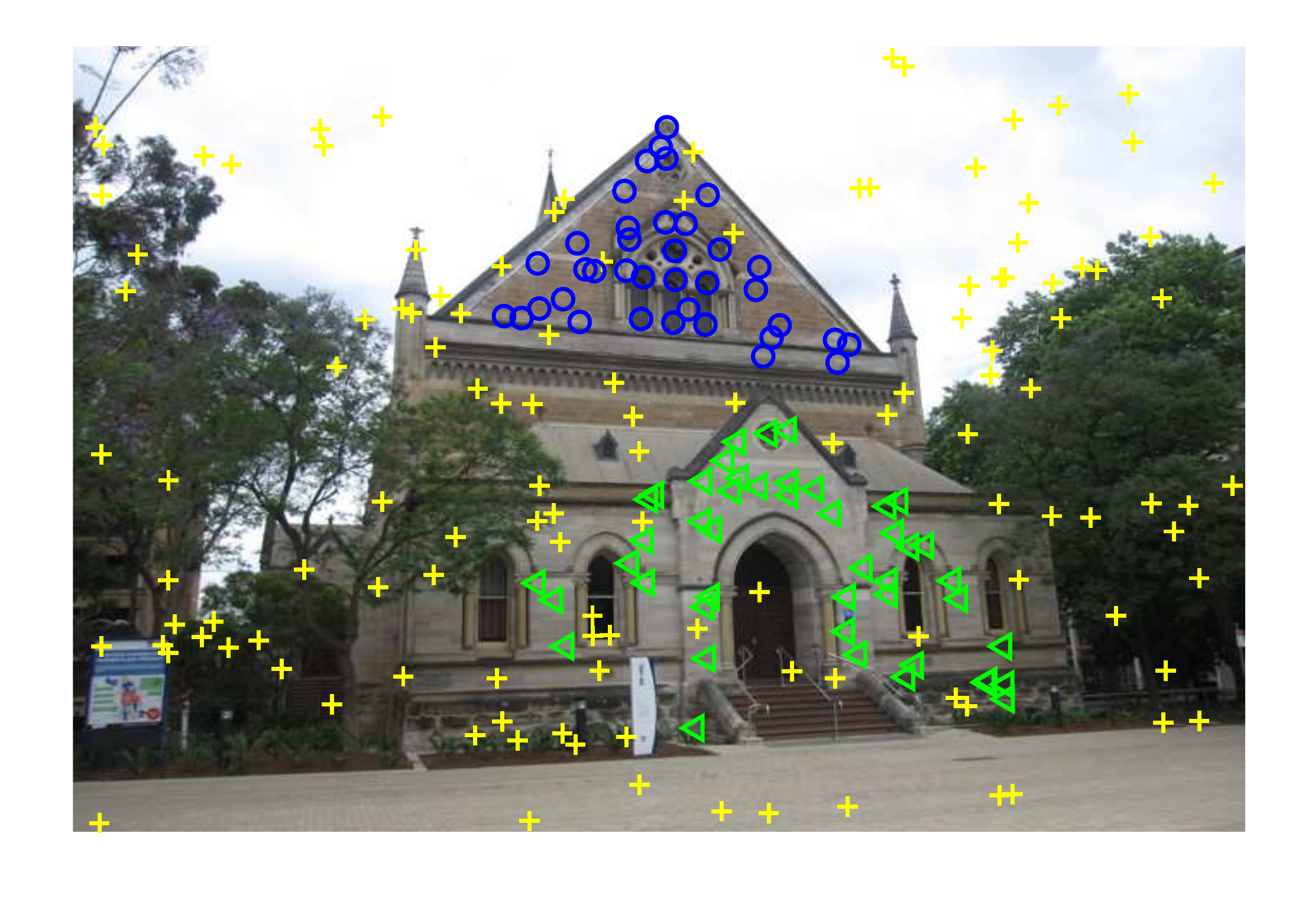}}
  \centerline{(g) Elderhalla}
\end{minipage}
\begin{minipage}{.19\textwidth}
\centerline{\includegraphics[width=1.06\textwidth]{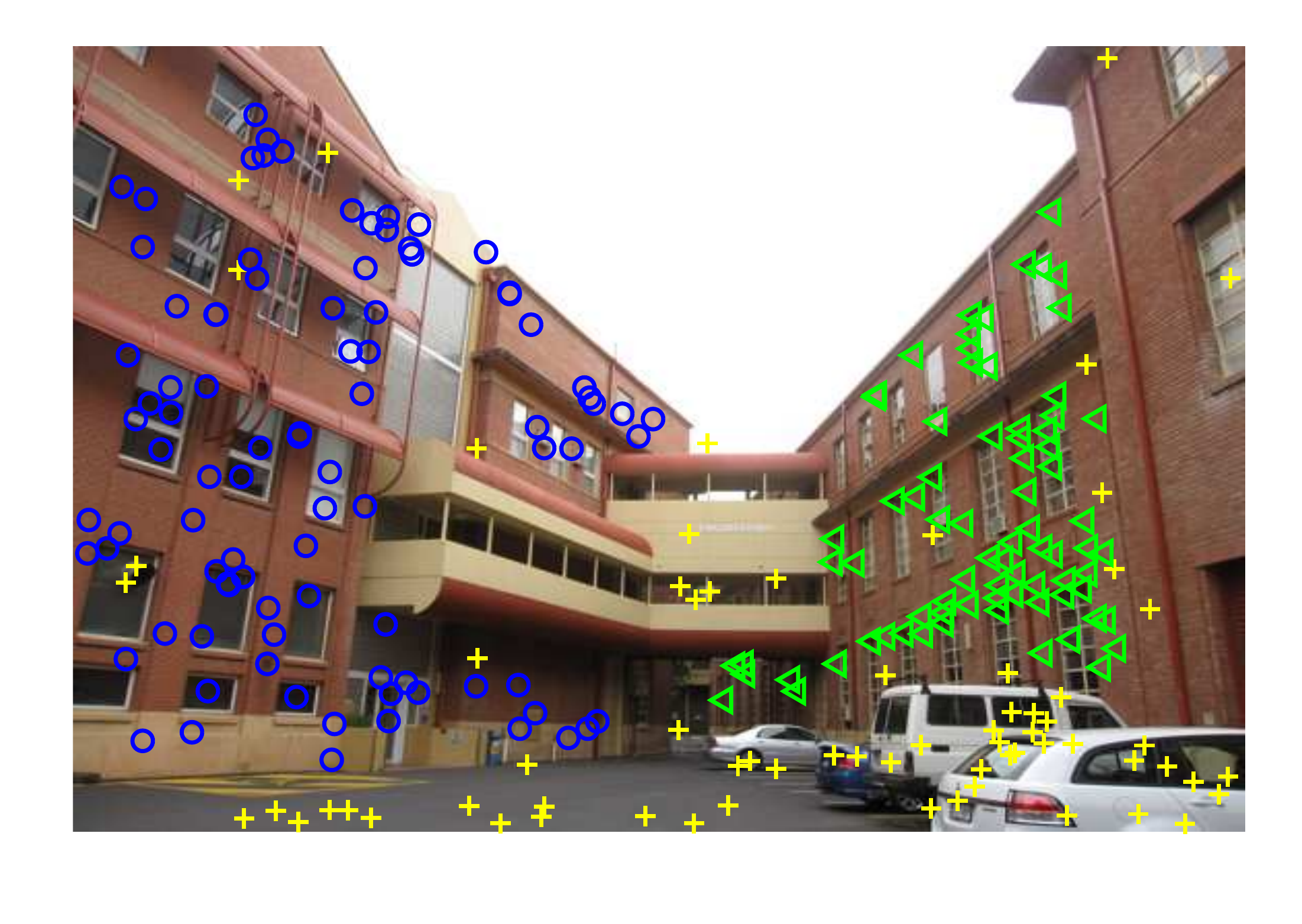}}
  \centerline{(c) Nese}
  \centerline{\includegraphics[width=1.06\textwidth]{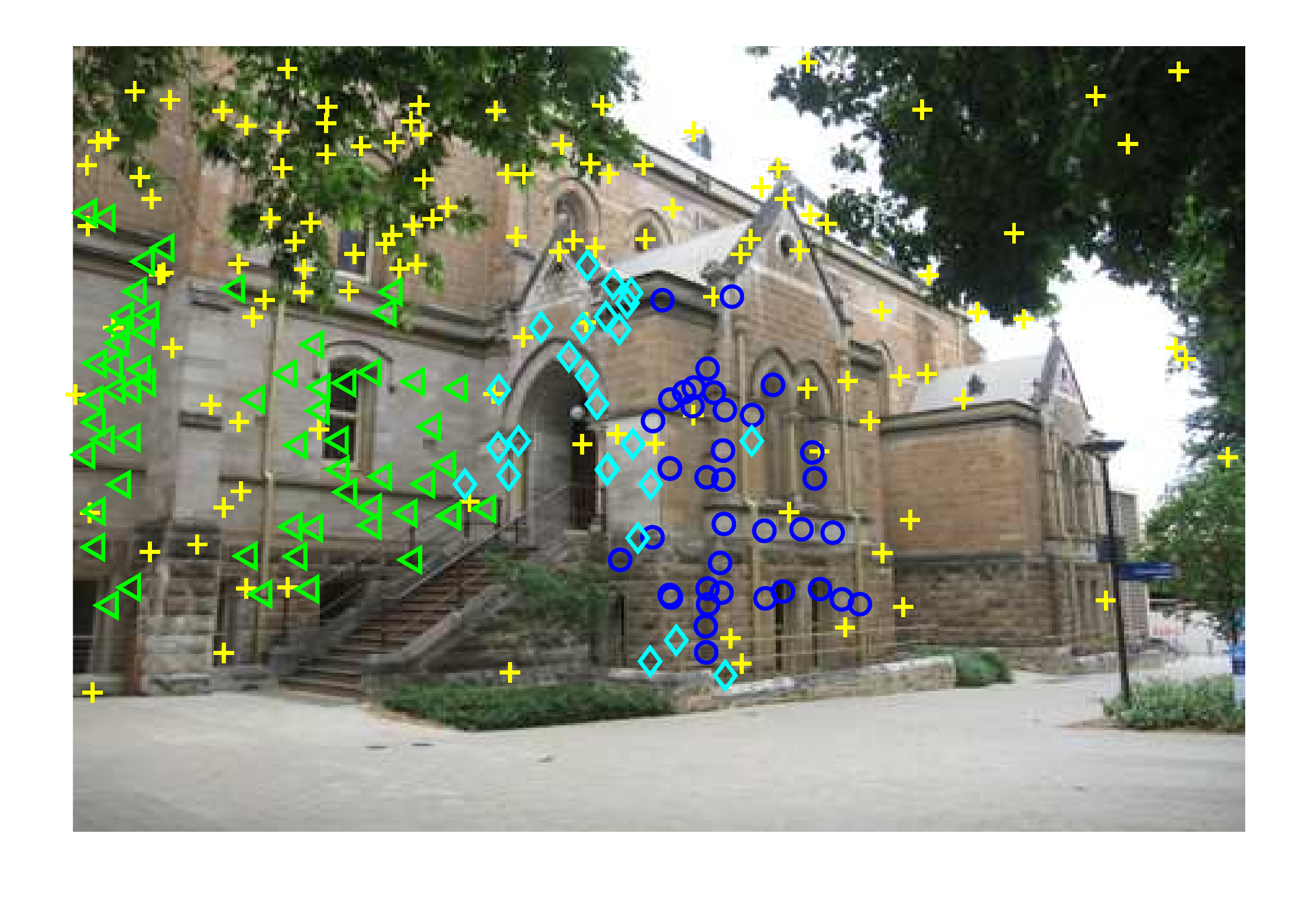}}
  \centerline{(h) Elderhallb}
\end{minipage}
\begin{minipage}{.19\textwidth}
\centerline{\includegraphics[width=1.06\textwidth]{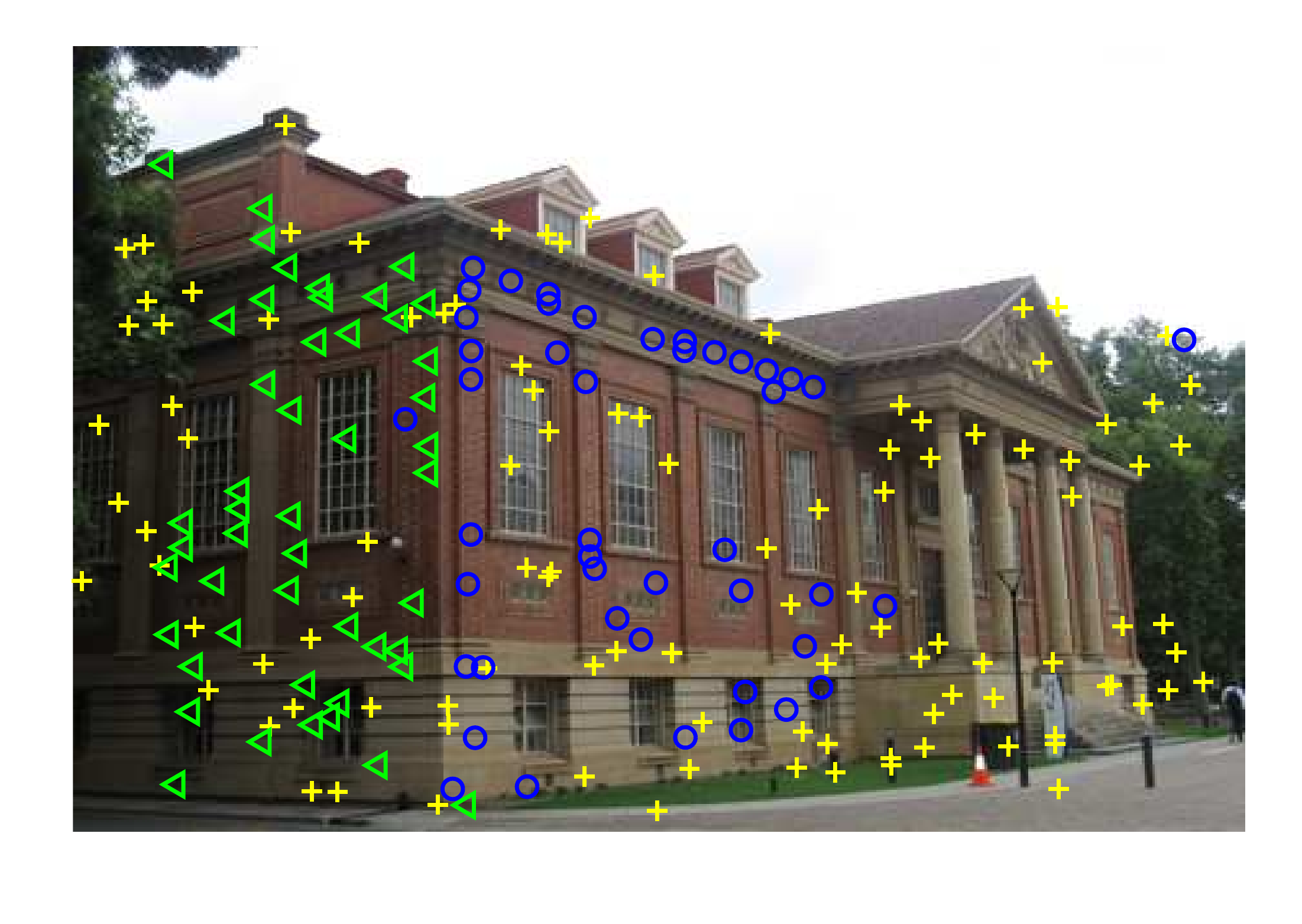}}
  \centerline{(d) Library}
  \centerline{\includegraphics[width=1.06\textwidth]{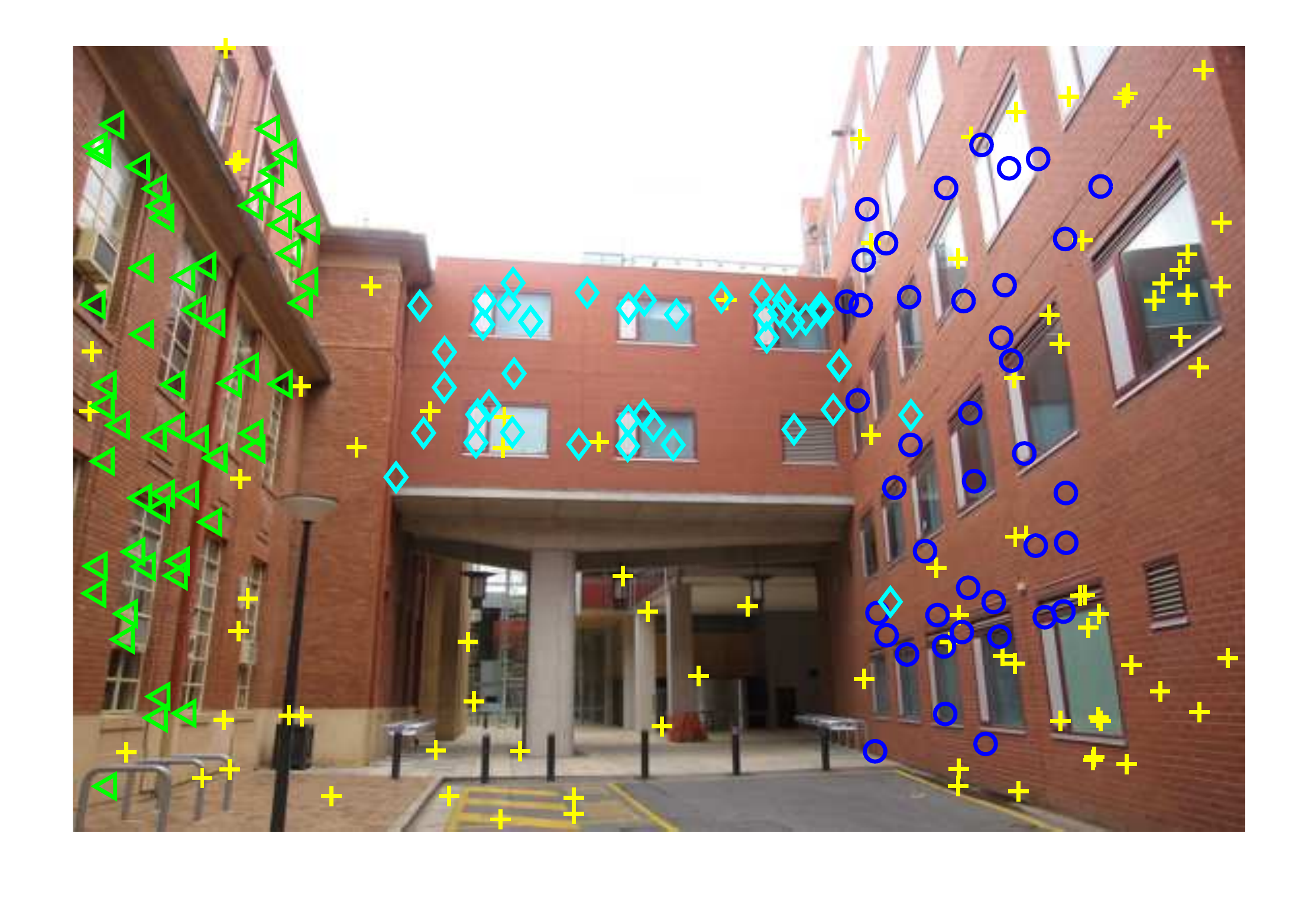}}
  \centerline{(i) Neem}
\end{minipage}
\begin{minipage}{.19\textwidth}
\centerline{\includegraphics[width=1.06\textwidth]{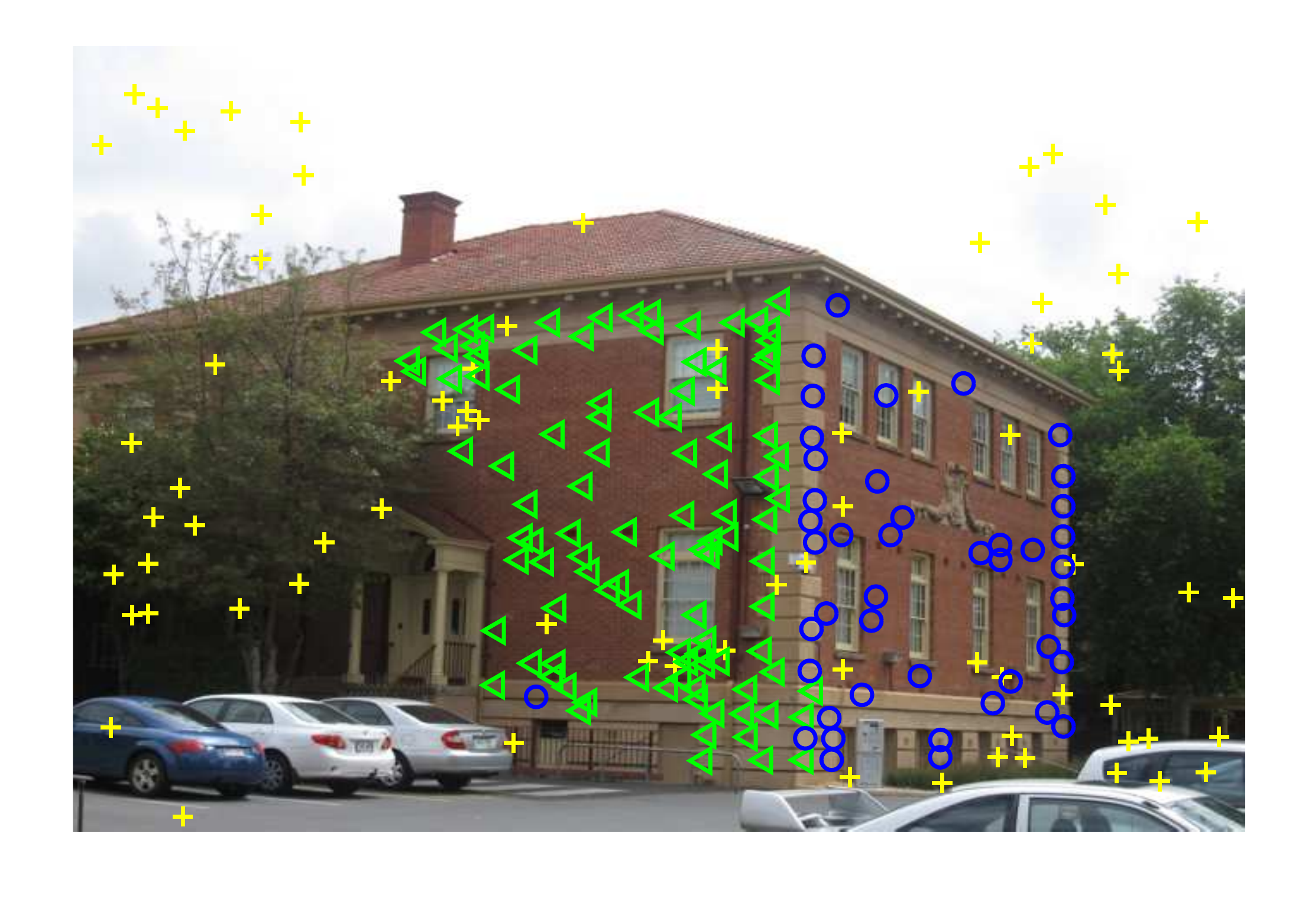}}
  \centerline{(e) Ladysymon}
  \centerline{\includegraphics[width=1.06\textwidth]{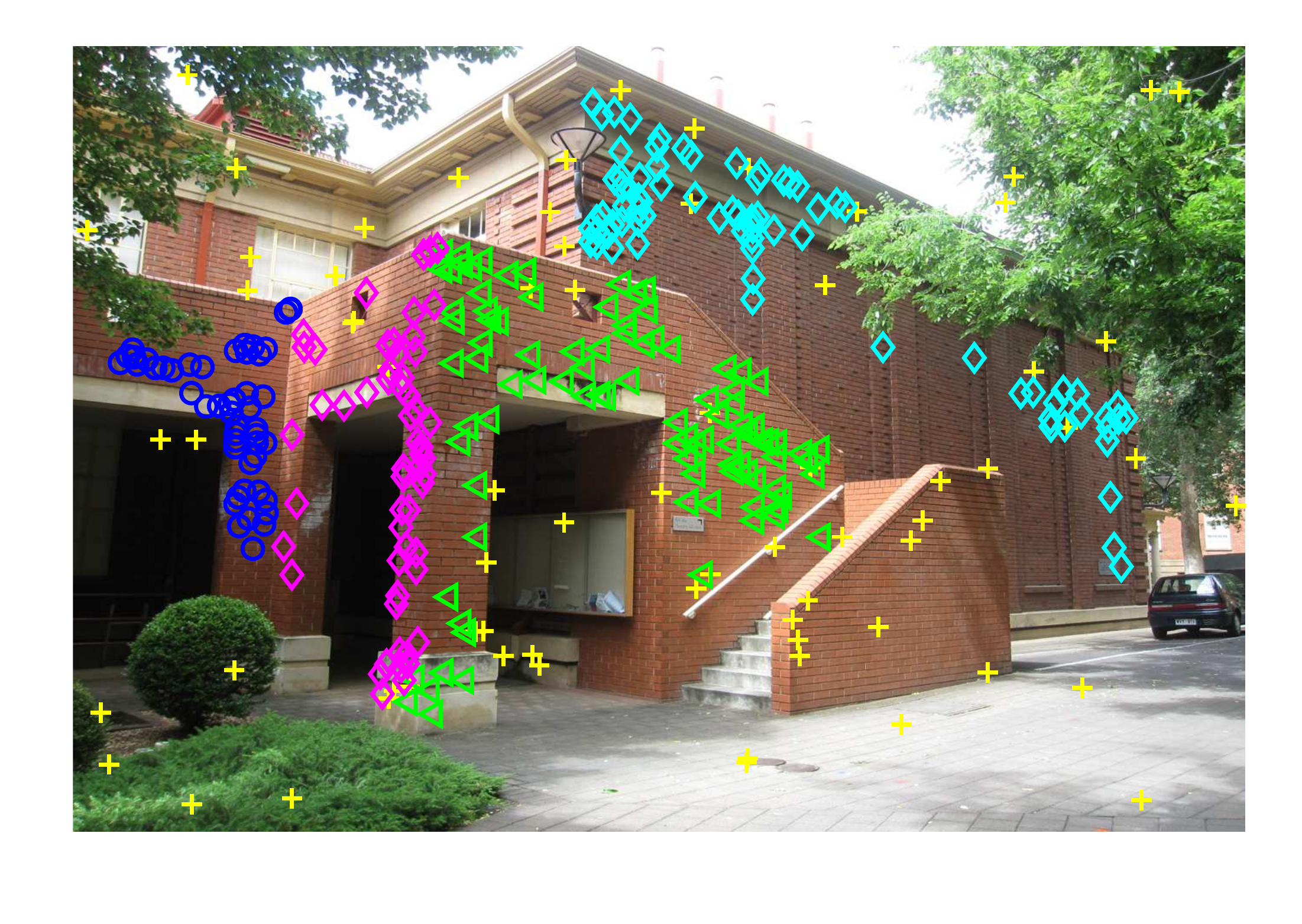}}
  \centerline{(j) Johnsona}
\end{minipage}
\caption{The fitting results obtained by SDF3 on ten image pairs with multiple structures for homography estimation. Only one of the two views is shown for each case.}
\label{fig:multistructure_homography}
\end{figure*}
\begin{figure*}[t!]
\begin{minipage}{.19\textwidth}
\centerline{\includegraphics[width=1.06\textwidth]{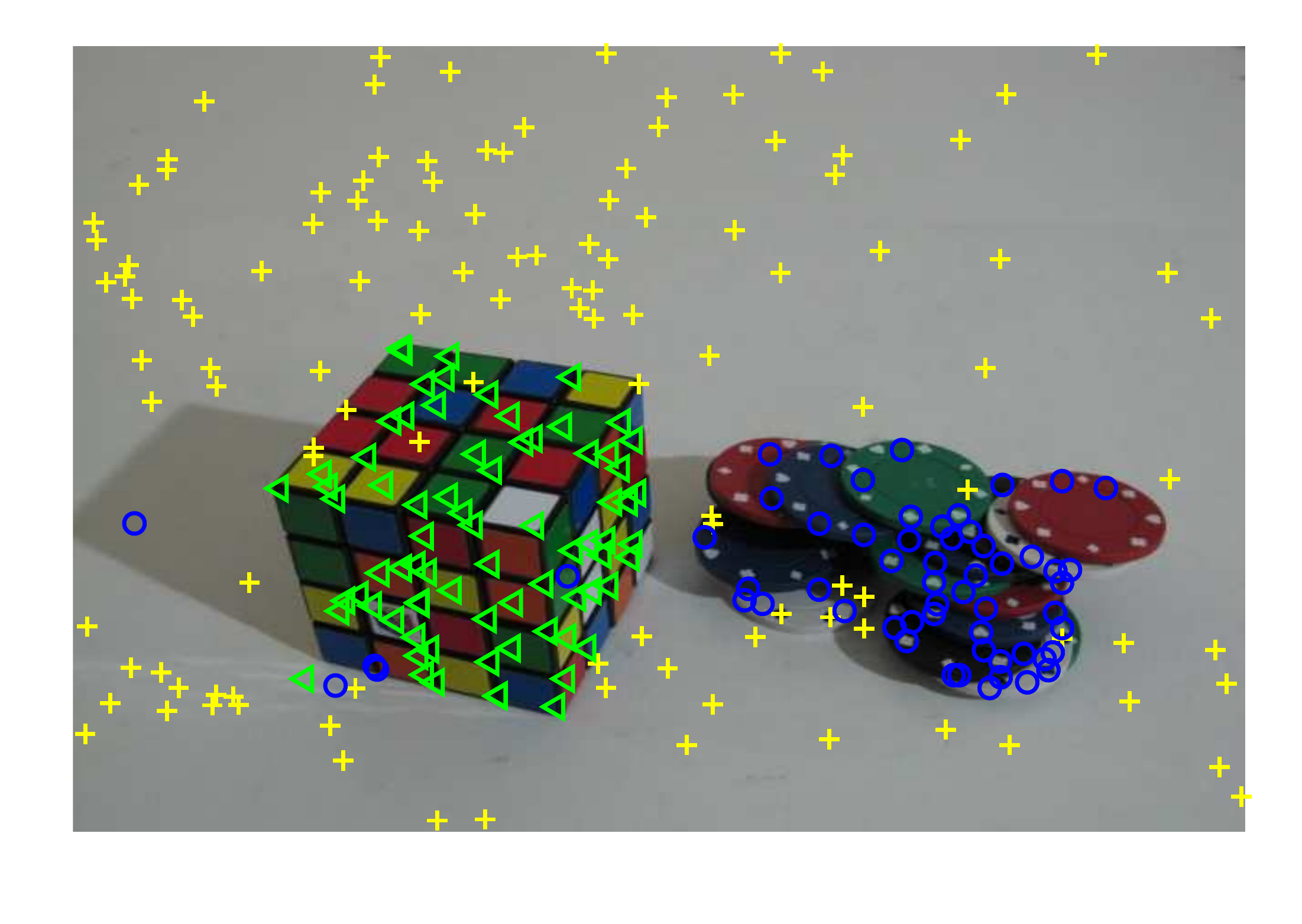}}
  \centerline{(a) }
  \centerline{\includegraphics[width=1.06\textwidth]{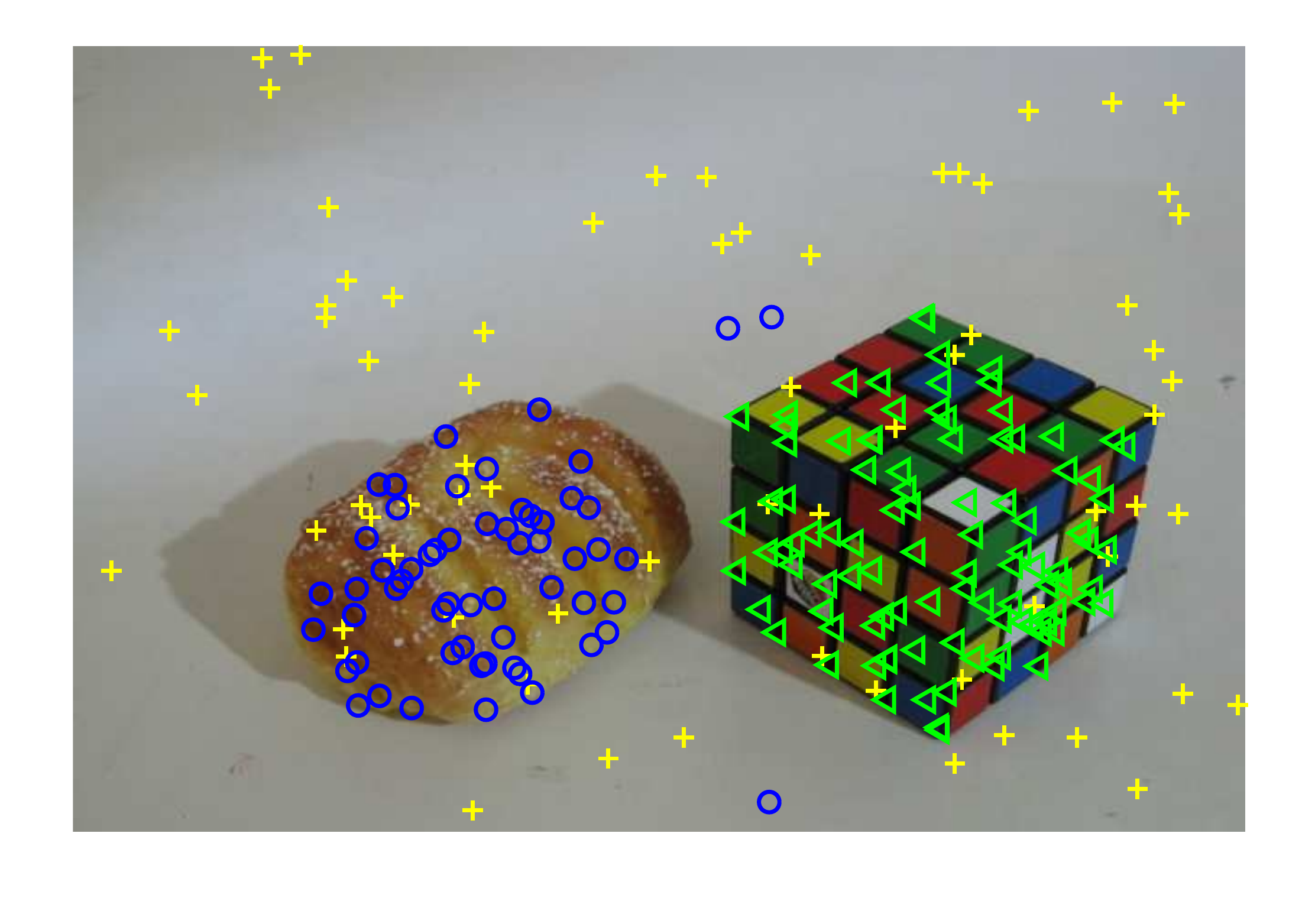}}
  \centerline{(f) }
\end{minipage}
\begin{minipage}{.19\textwidth}
\centerline{\includegraphics[width=1.06\textwidth]{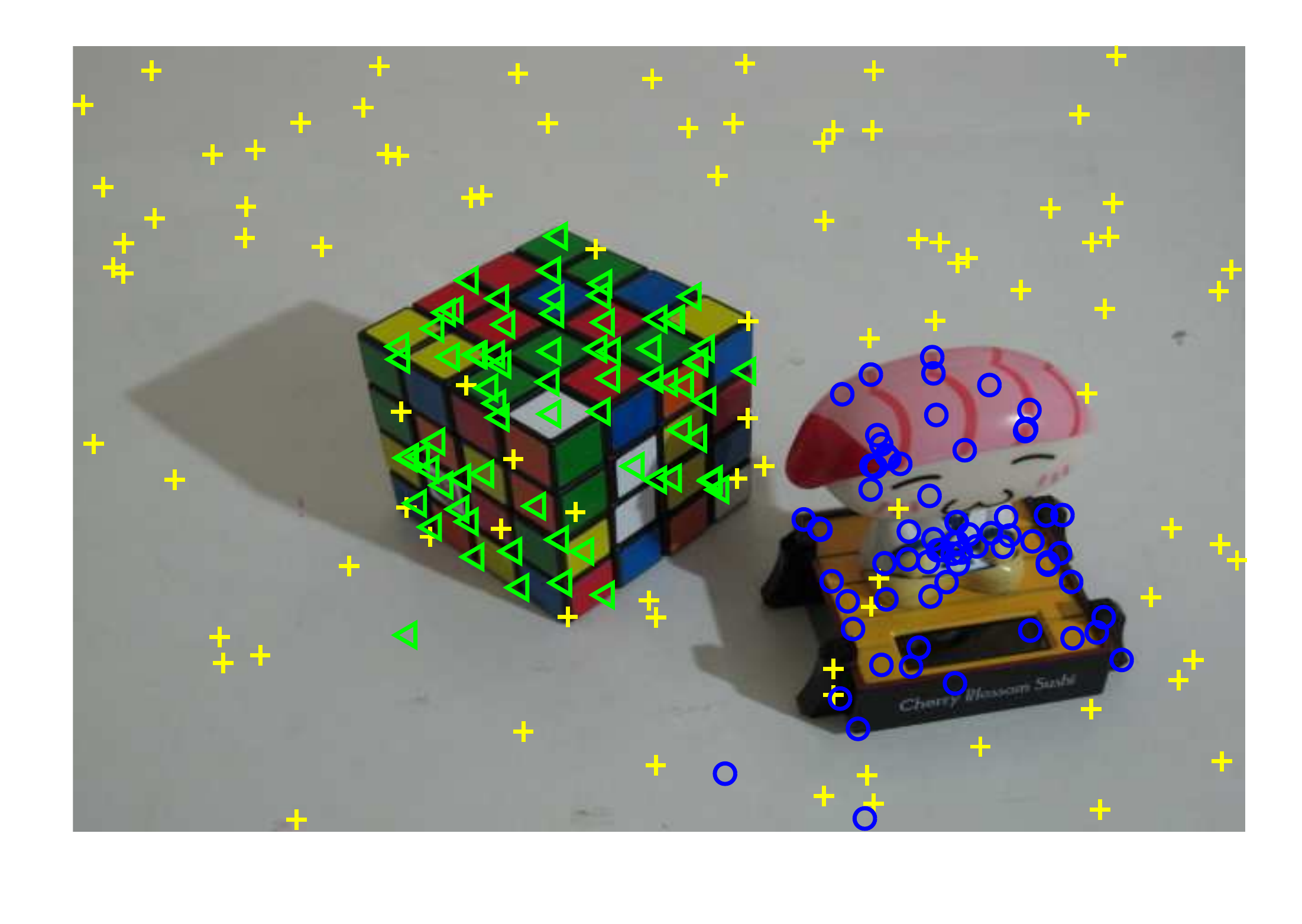}}
  \centerline{(b) }
  \centerline{\includegraphics[width=1.06\textwidth]{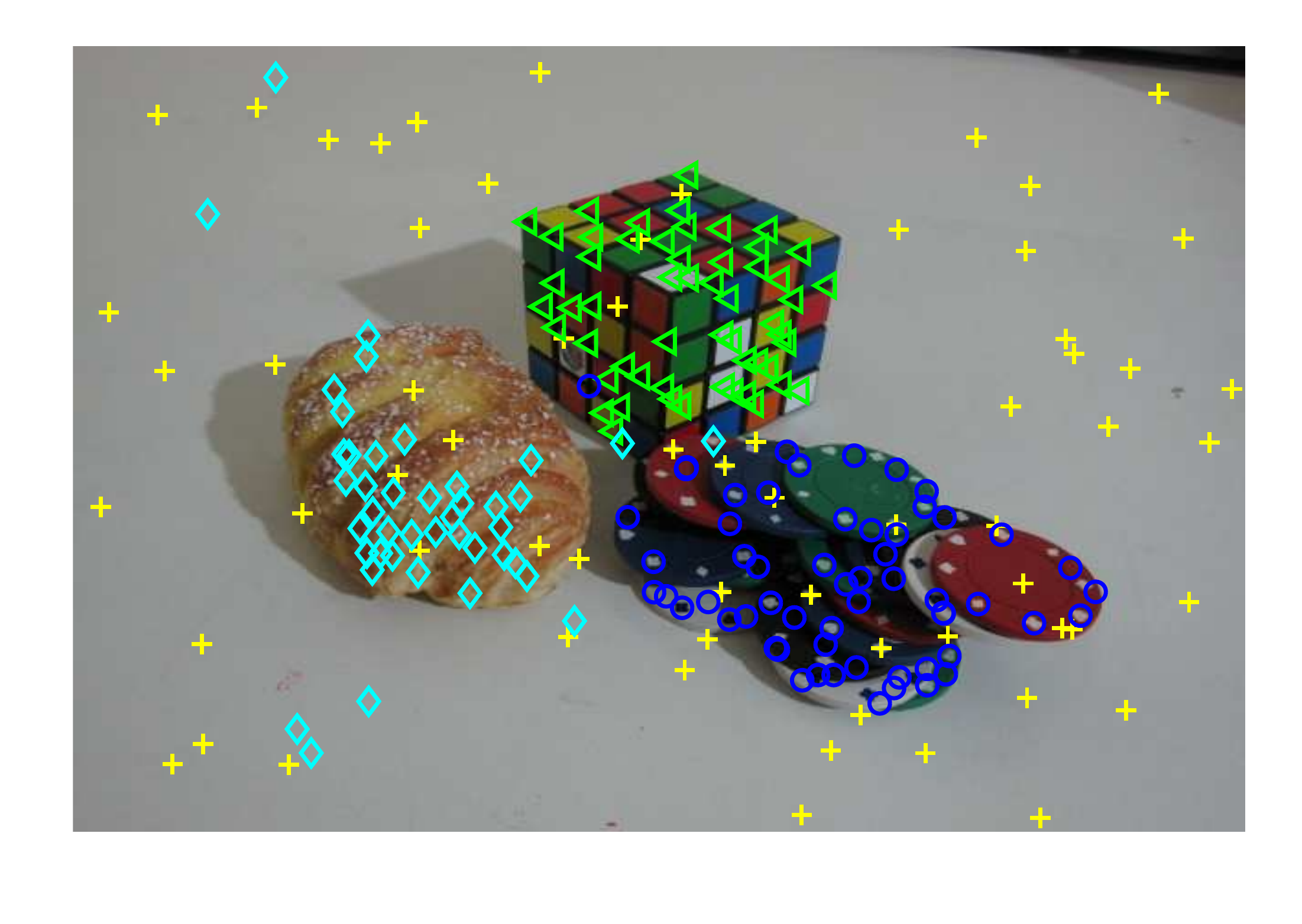}}
  \centerline{(g) }
\end{minipage}
\begin{minipage}{.19\textwidth}
\centerline{\includegraphics[width=1.06\textwidth]{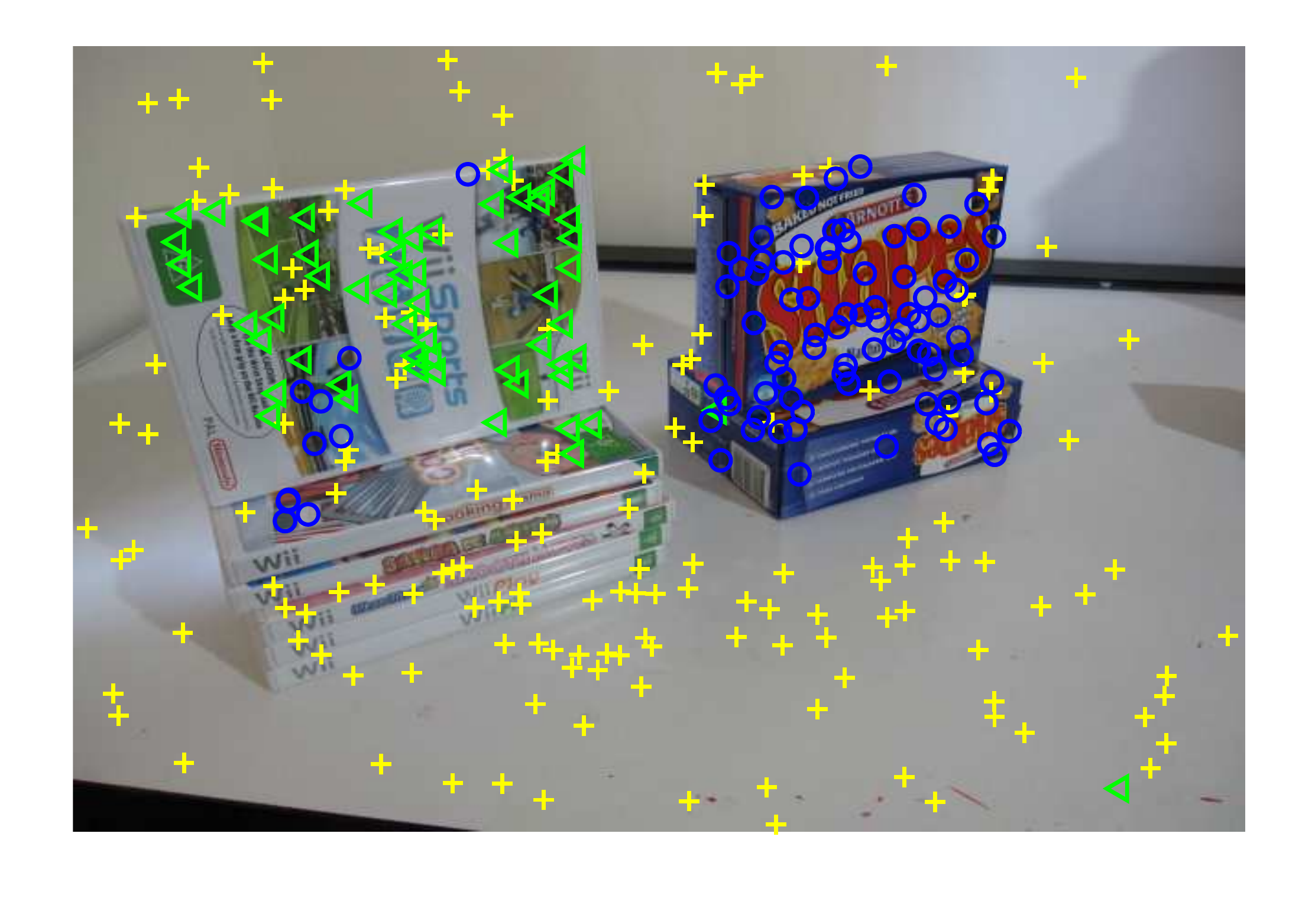}}
  \centerline{(c) }
  \centerline{\includegraphics[width=1.06\textwidth]{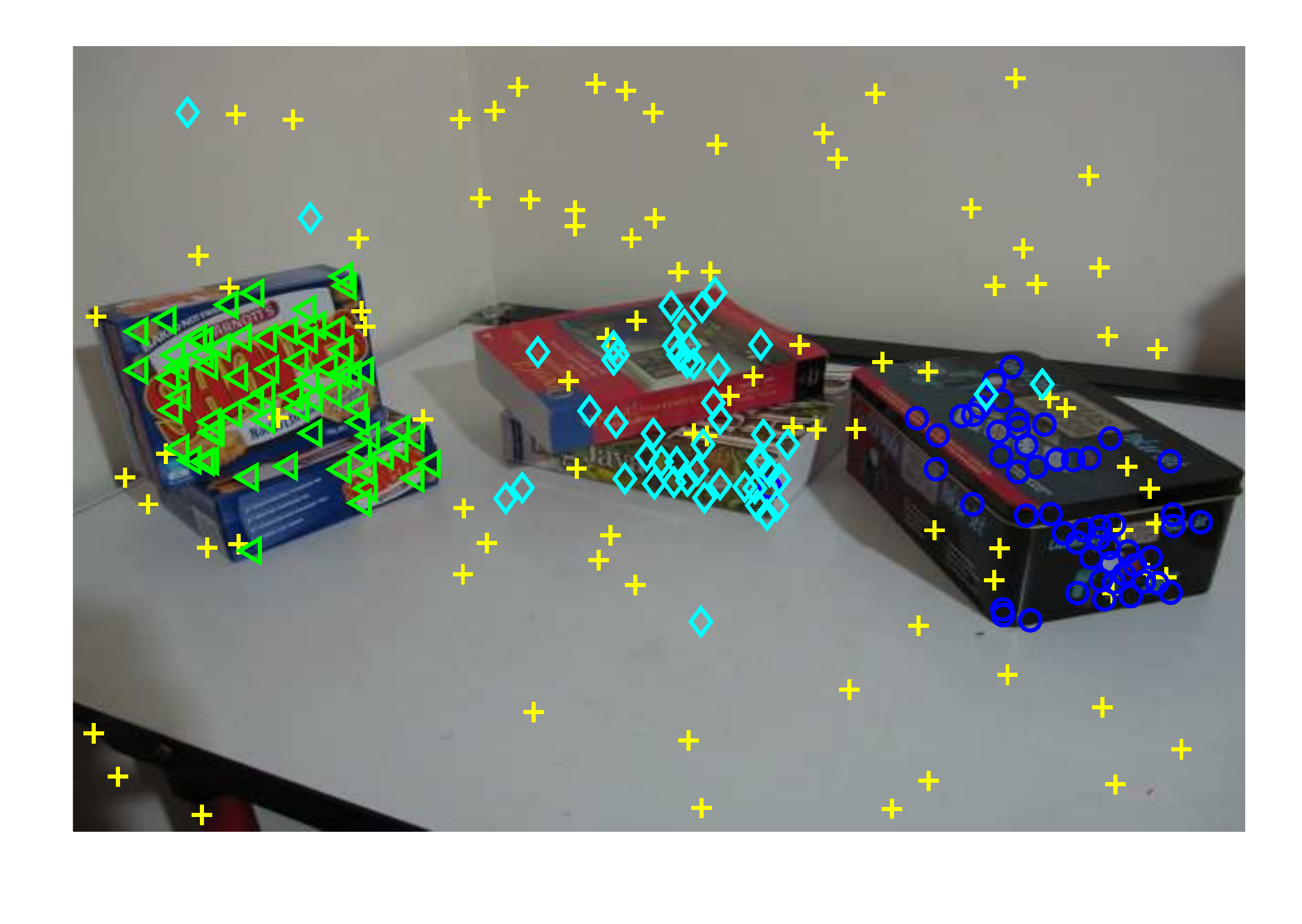}}
  \centerline{(h) }
\end{minipage}
\begin{minipage}{.19\textwidth}
\centerline{\includegraphics[width=1.06\textwidth]{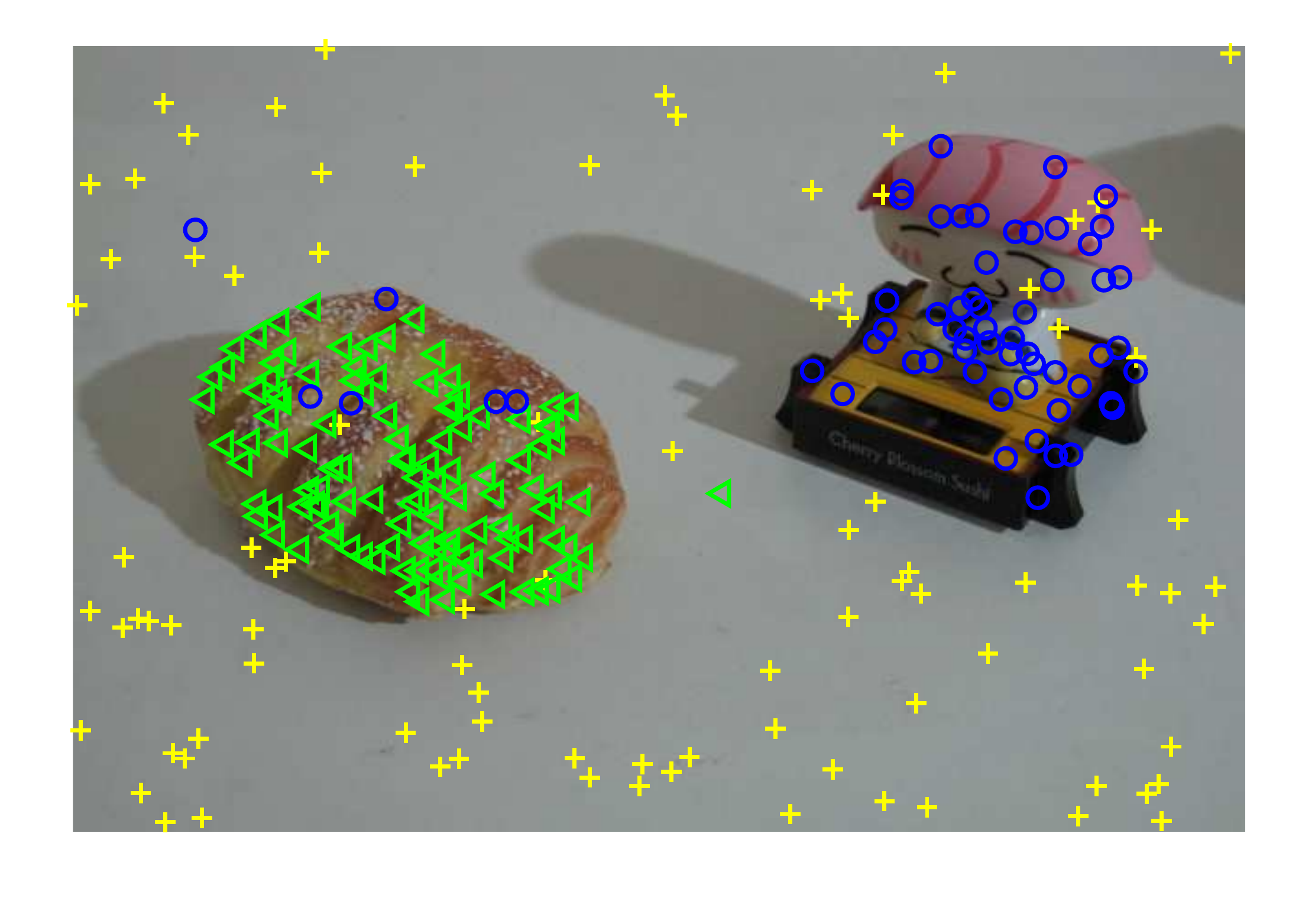}}
  \centerline{(d) }
  \centerline{\includegraphics[width=1.06\textwidth]{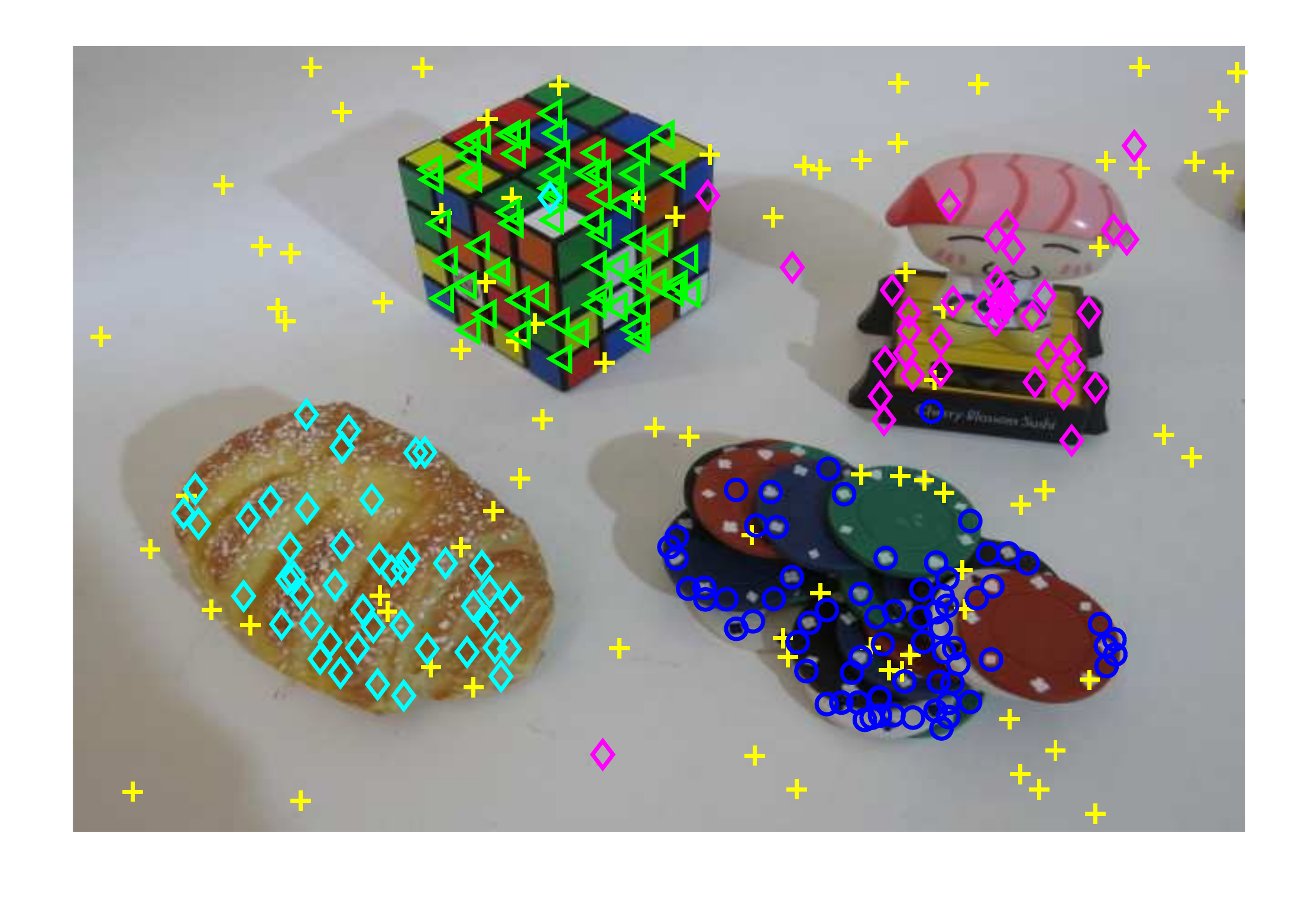}}
  \centerline{(i) }
\end{minipage}
\begin{minipage}{.19\textwidth}
\centerline{\includegraphics[width=1.06\textwidth]{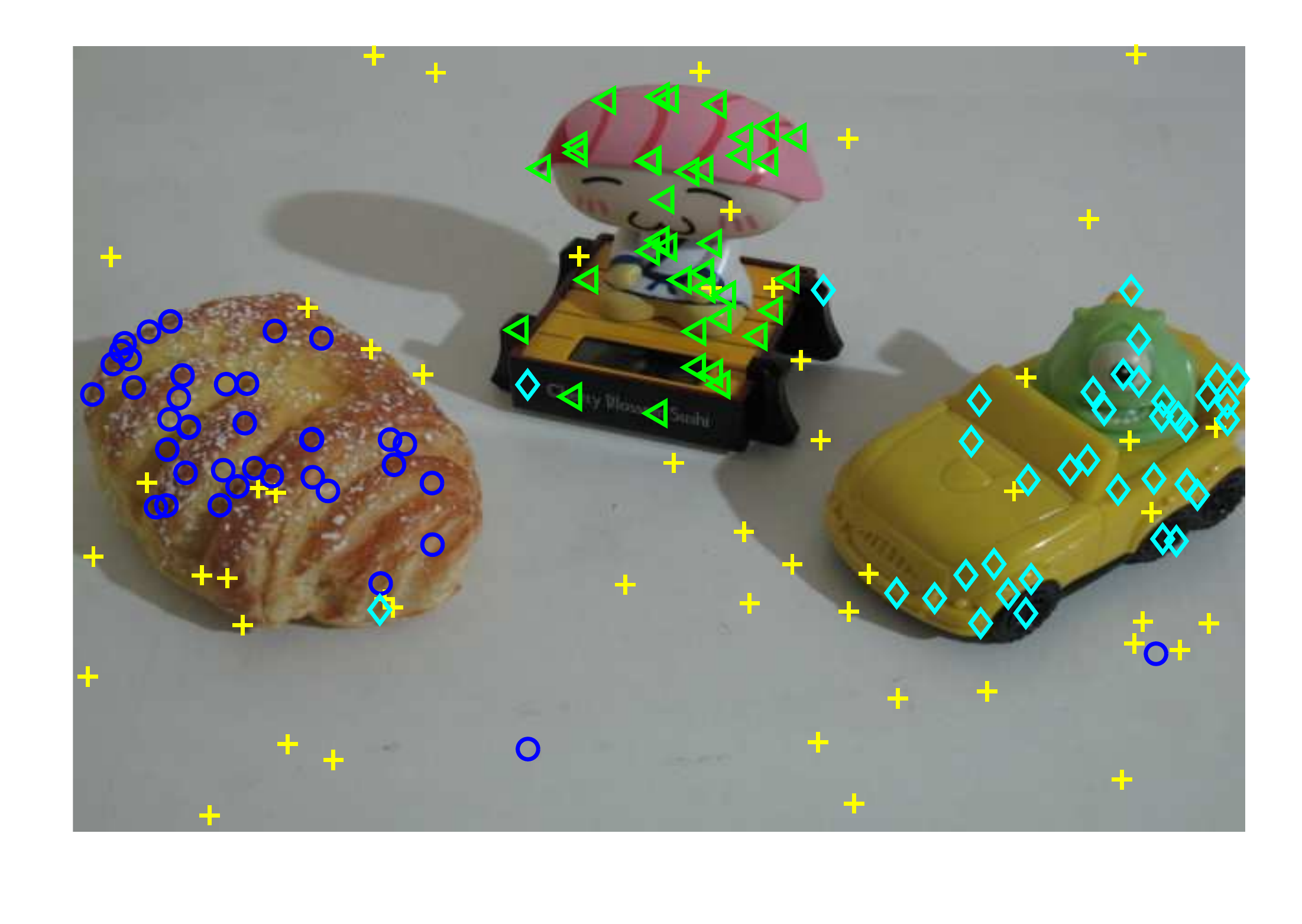}}
  \centerline{(e) }
  \centerline{\includegraphics[width=1.06\textwidth]{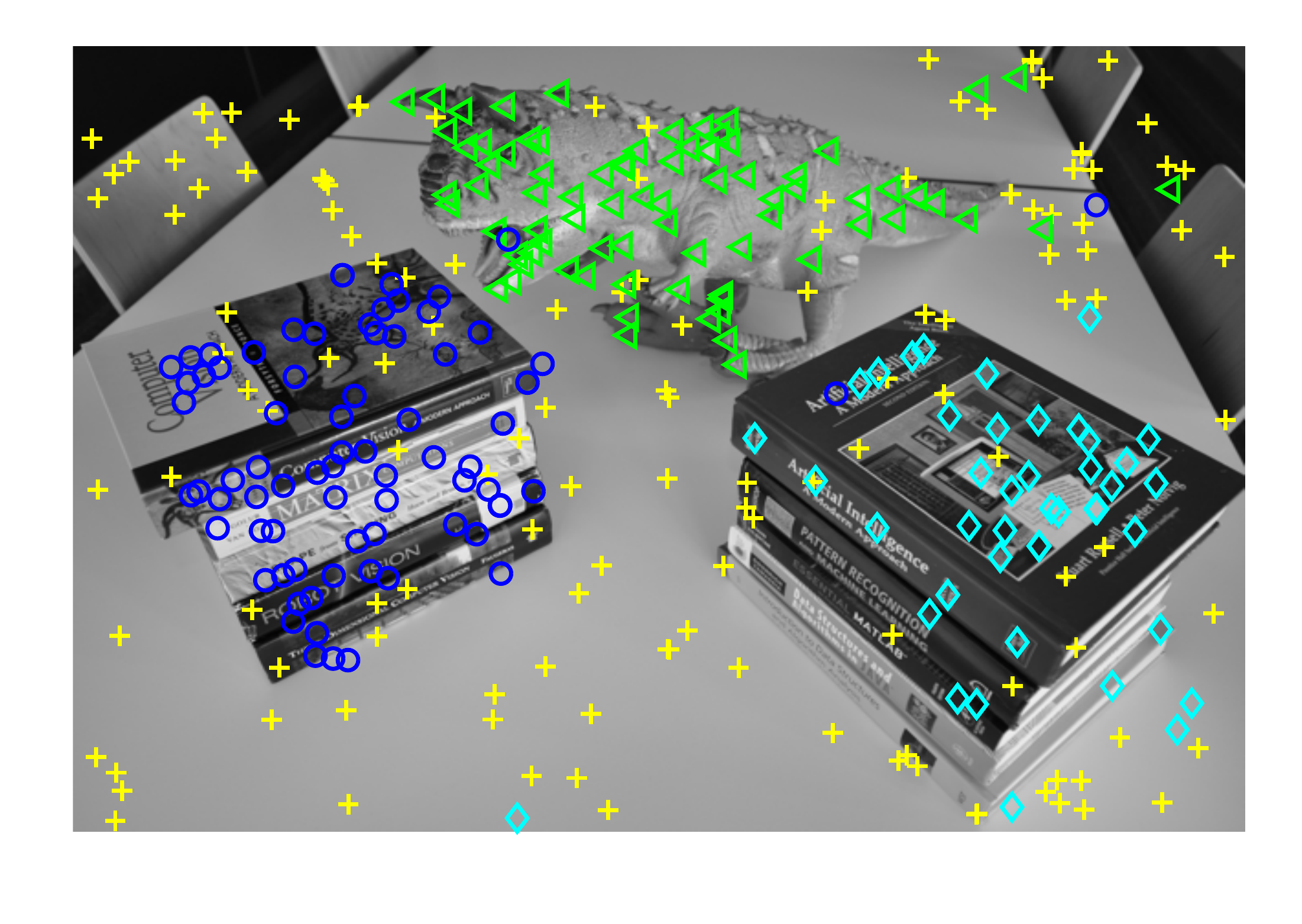}}
  \centerline{(j) }
\end{minipage}
\caption{The fitting results obtained by SDF3 on ten image pairs (namely (a) Cubechips, (b) Cubetoy, (c) Gamebiscuit, (d) Breadtoy, (e) Breadtoycar, (f) Breadcube, (g) Breadcubechips, (h) Biscuitbookboxj, (i) Cubebreadtoychips and (j) Dinobooks) with multiple structures for fundamental matrix estimation. Only one of the two views is shown for each case.}
\label{fig:multistructure_fundamental}
\end{figure*}
\begin{table}[t]
\caption{Quantitative comparison results on multiple-structure data for homography estimation.}
\centering
\scalebox{0.8}{
{\tabcolsep0.04in
\begin{tabular}{llccccc}
\toprule
Data     &                   & RANSAC&PROSAC&T-linkage&RansaCov&SDF3\\
\midrule
 &Std.      &{0.34}&{0.39}& {0.70}  &{4.70}&{--}\\
Oldclassicswing                  &Avg.      &{3.15}&{3.42}& {2.08}  &{4.93}&{\bf0.82}\\
                       &Time      &16.24&{16.40}&69.72&29.38 &{\bf2.21}\\
            \rowcolor{mygray}
             &Std.      &{0.01}&{0.01}& {0.19}  &{0.12}&{--}\\
             \rowcolor{mygray}
Sene                &Avg.     &{\bf0.41}&{\bf0.41}&{\bf0.41}&{1.68}&{\bf0.41} \\
     \rowcolor{mygray}
                       &Time      &16.52&{16.72}&35.44&19.80&{\bf0.94}\\

                         &Std.      &{8.40}&{8.45}& {0.58}  &{0.35}&{--}\\
   Nese                 &Avg.     &{8.04}&{8.37}&{1.46}&{3.69}&{\bf0.00} \\
                       &Time      &15.98&{16.24}&35.72&17.69&{\bf1.27}\\

          \rowcolor{mygray}
            &Std.      &{2.04}&{2.11}& {0.57}  &{1.28}&{--}\\
            \rowcolor{mygray}
  Library                &Avg.      &{7.82}&{8.51}&{6.08}&{6.38}&{\bf1.49} \\
     \rowcolor{mygray}
                       &Time      &15.89&{15.96}& 25.46&17.30&{\bf0.86}\\
             &Std.      &{0.79}&{0.43}& {1.91}  &{0.76}&{--}\\
 Ladysymon               &Avg.     &{10.08}&{10.04}&{7.30}&{4.30}&{\bf2.24}\\
                    &Time      &15.45&{15.75}&29.44&17.39&{\bf1.30}\\

            \rowcolor{mygray}
             &Std.      &{6.62}&{4.91}& {0.46}  &{0.29}&{--}\\
             \rowcolor{mygray}
  Hartley               &Avg.      &{11.36}&{14.64}&{2.86}&{3.10}&{\bf1.02}\\
    \rowcolor{mygray}
                    &Time      &16.04&{16.38}&45.15&15.45&{\bf8.36}\\

             &Std.      &{0.23}&{0.85}& {1.26}  &{0.01}&{--}\\
    Elderhalla     &Avg.     &{1.29}&{1.48}&{1.96}&{1.43}&{\bf0.96}\\
                                    &Time      &15.51&{15.86}&26.13&17.24&{\bf1.07}\\

           \rowcolor{mygray}
              &Std.      &{10.04}&{9.76}& {1.54}  &{1.82}&{--}\\
              \rowcolor{mygray}
   Elderhallb      &Avg.      &26.83&24.79&12.62&11.54&{\bf 3.33}\\
    \rowcolor{mygray}
                                    &Time      &23.57&{24.01}&32.86&15.62&{\bf1.58}\\

               &Std.      &{3.89}&{5.74}& {9.01}  &{2.31}&{--}\\

Neem                    &Avg.       &{25.15}&{18.66}&{14.53}&{11.24}&{\bf0.89}\\

                         &Time      &23.59&{23.81}&29.33&15.25&{\bf0.98}\\

        \rowcolor{mygray}
            &Std.      &{4.57}&{4.36}& {1.60}  &{0.88}&{--}\\
            \rowcolor{mygray}
Johnsona   &Avg.       &{21.87}&{23.77}&{5.90}&{5.87}&{\bf1.15}\\
     \rowcolor{mygray}
                         &Time      &32.58 &{32.29}&62.14&17.15&{\bf2.75}\\
\bottomrule
\end{tabular}}}
\\
\medskip
\raggedright
('--' denotes the results obtained by the deterministic fitting methods, which are not implemented repeatedly. The best results are boldfaced.)
 \label{table:homographytable2}
\end{table}
\subsection{Multiple-structure Data}
\label{sec:multiplestructures}
In this subsection, we evaluate the performance of the five fitting methods (i.e., RANSAC, PROSAC, T-linkage,
RansaCov and SDF3) on 20 image pairs with multiple structures for homography and fundamental matrix estimation. Note that we do not compare deterministic fitting methods with SDF3. This is because most of existing deterministic fitting methods only work for single-structure data (\cite{lee2013deterministic,Enqvist2015} are two exceptions as we known but the original codes are not provided and some implementation details are missing). We report the standard variances, the average fitting errors and the average computational time (i.e., the CPU time used in seconds)  in Table~\ref{table:homographytable2} and Table~\ref{table:fundtable2} for homography and fundamental matrix estimation, respectively. For RANSAC, PROSAC, T-linkage and RansaCov, we show the average results (we repeat the experiments $50$ times) due to their {randomness}. For SDF3, we do not repeat the experiments due to its deterministic nature. The fitting results obtained by SDF3 for homography and fundamental matrix estimation are also shown in Figs.~\ref{fig:multistructure_homography} and \ref{fig:multistructure_fundamental}, respectively.

$\textsc{Homography~Estimation.}$ From Fig.~\ref{fig:multistructure_homography} and Table~\ref{table:homographytable2}, we can see that SDF3 shows significant superiority over the other four competing methods on the performance of the average fitting errors and the average computational time. RansaCov and T-linkage achieve low fitting errors for $8$ out of $10$ image pairs. RansaCov and T-linkage are clustering-based fitting methods, and they cannot effectively deal with the data points in the intersection of two model instances for some image pairs, e.g., ``Elderhallb" and ``Neem". They use the same sampling strategy, while RansaCov is much faster than T-linkage. RANSAC and PROSAC achieve low fitting errors for some easy image pairs (i.e., ``Oldclassicswing", ``Sene" and ``Elderhalla"), but both often fail for {the} other image pairs. They are also much slower than SDF3 because they use the conventional ``fit and remove" framework, which requires to repeatedly generate model hypotheses for each model instance, for multiple-structure data. In contrast, SDF3 achieves the lowest fitting errors and the fastest computational speed among the five fitting methods for all the $10$ image pairs, which shows that SDF3 is an effective and efficient fitting method.

\begin{table}[t]
\caption{Quantitative comparison results on multiple-structure data for fundamental matrix estimation. }
\centering
\scalebox{0.8}{{\tabcolsep0.04in
\begin{tabular}{llccccc}
\toprule
Data     &                  & RANSAC&PROSAC&T-linkage&RansaCov&SDF3\\
\midrule
 &Std.      &{12.17}&{12.64}& {3.24}  &{9.52}&{--}\\
 Cubechips                    &Avg.      &{13.21}&{16.03}& {3.76}  &{9.80}&{\bf3.14}\\
                       &Time      &28.67&{29.49}&82.32&39.43 &{\bf0.88}\\

           \rowcolor{mygray}
             &Std.      &{6.87}&{10.54}& {0.80}  &{1.16}&{--}\\
             \rowcolor{mygray}
Cubetoy                  &Avg.     &{34.08}&{33.78}&{5.62}&{11.78}&{\bf2.61} \\
     \rowcolor{mygray}
                       &Time      &29.27&{30.13}&67.85&30.37&{\bf0.82}\\

            &Std.      &{3.46}&{4.39}& {1.80}  &{1.21}&{--}\\
   Gamebiscuit                 &Avg.      &{25.66}&{22.72}&{7.33}&{5.12}&{\bf5.06} \\
                       &Time      &34.26&{34.60}&114.39&45.65&{\bf1.03}\\

           \rowcolor{mygray}
             &Std.      &{1.38}&{1.84}& {0.88}  &{3.32}&{--}\\
             \rowcolor{mygray}
Breadtoy               &Avg.     &{10.07}&{10.85}&{3.54}&{28.47}&{\bf2.61}\\
     \rowcolor{mygray}
                    &Time      &29.56&{30.47}&84.15&38.10&{\bf1.05}\\

             &Std.      &{4.05}&{2.66}& {2.10}  &{3.91}&{--}\\
Breadtoycar                &Avg.      &{36.38}&{37.60}&{\bf4.42}&{17.66}&{6.74}\\
                    &Time      &29.54&{29.99}&40.81&30.21&{\bf0.85}\\

            \rowcolor{mygray}
             &Std.      &{2.53}&{1.56}& {1.32}  &{5.66}&{--}\\
             \rowcolor{mygray}
Breadcube     &Avg.     &{13.25}&{12.29}&{4.96}&{18.98}&{\bf1.76}\\
    \rowcolor{mygray}
                                    &Time      &27.38&{28.17}&64.67&31.30&{\bf 0.81}\\

              &Std.      &{8.06}&{9.56}& {11.92}  &{5.62}&{--}\\
Breadcubechips      &Avg.      &37.16&35.14&{ 5.61}&28.28&{\bf 5.00}\\
                                    &Time      &62.18&{63.92}&105.91&34.28&{\bf1.24}\\

            \rowcolor{mygray}
               &Std.      &{6.68}&{6.97}& {1.61}  &{12.60}&{--}\\
               \rowcolor{mygray}
Biscuitbookbox                 &Avg.       &{23.32}&{26.21}&{7.50}&{23.35}&{\bf4.69}\\
    \rowcolor{mygray}
                         &Time      &41.52&{43.06}&113.07&34.74&{\bf1.05}\\

            &Std.      &{8.20}&{7.58}& {13.18}  &{3.21}&{--}\\
    Cubebreadtoychips  &Avg.       &{32.07}&{40.72}&{16.13}&{27.52}&{\bf4.95}\\
                         &Time      &59.87 &{62.62}&101.85&32.54&{\bf1.06}\\

            \rowcolor{mygray}
            &Std.      &{4.10}&{3.64}& {1.80}  &{2.56}&{--}\\
            \rowcolor{mygray}
Dinobooks                   &Avg.        &{23.58}&{23.43}&{18.73}&{33.28}&{\bf8.43} \\
     \rowcolor{mygray}
                         &Time      &49.91&{50.98} &123.35&38.13&{\bf0.85}\\
\bottomrule
\end{tabular}}}
\\
\medskip
\raggedright
('--' denotes the results obtained by the deterministic fitting methods, which are not implemented repeatedly. The best results are boldfaced.)
 \label{table:fundtable2}
\end{table}
$\textsc{Fundamental~Matrix~Estimation.}$ From Fig.~\ref{fig:multistructure_fundamental} and Table~\ref{table:fundtable2}, we can see that SDF3 shows significant superiority over the other four competing methods on the performance of the average fitting errors and computational time. And SDF3 achieves the lowest average fitting errors for $9$ out of $10$ image pairs and second lowest average fitting errors for one image pair (i.e., ``Breadtoycar"). RansaCov is sensitive to the inlier noise scale. Note that the residual value for fundamental matrix estimation is very small (i.e., the discrimination between the residual values is not very obvious), thus RansaCov is hard to achieve low fitting errors. T-linkage achieves low fitting errors for $8$ out of $10$ image pairs, but it has typically high computational cost due to the use of the agglomerative clustering procedure.
Both RANSAC and PROSAC are slow for all $10$ image pairs because that it is difficult for them to sample all-inlier subsets by using the random sampling technique when the data contain multiple structures with a high outlier percentage and the estimation needs to sample a large number of subsets.
However, SDF3 achieves good {performance} of fitting errors for all the $10$ image pairs, and it is much faster than the other four fitting methods (about $28.15$$\sim$$58.72$ times faster than RANSAC, $29.02$$\sim$$59.98$ times faster than PROSAC, $48.01$$\sim$$145.12$ times faster than T-linkage, and $27.65$$\sim$$44.86$ times faster than RansaCov). The results show the effectiveness and efficiency of SDF3 for fitting multiple-structure data. 

\section{Discussion}
\label{sec:limitations}
\begin{figure}
\centering
\subfigure{
\begin{minipage}{.23\textwidth}
\centerline{\includegraphics[width=1.06\textwidth]{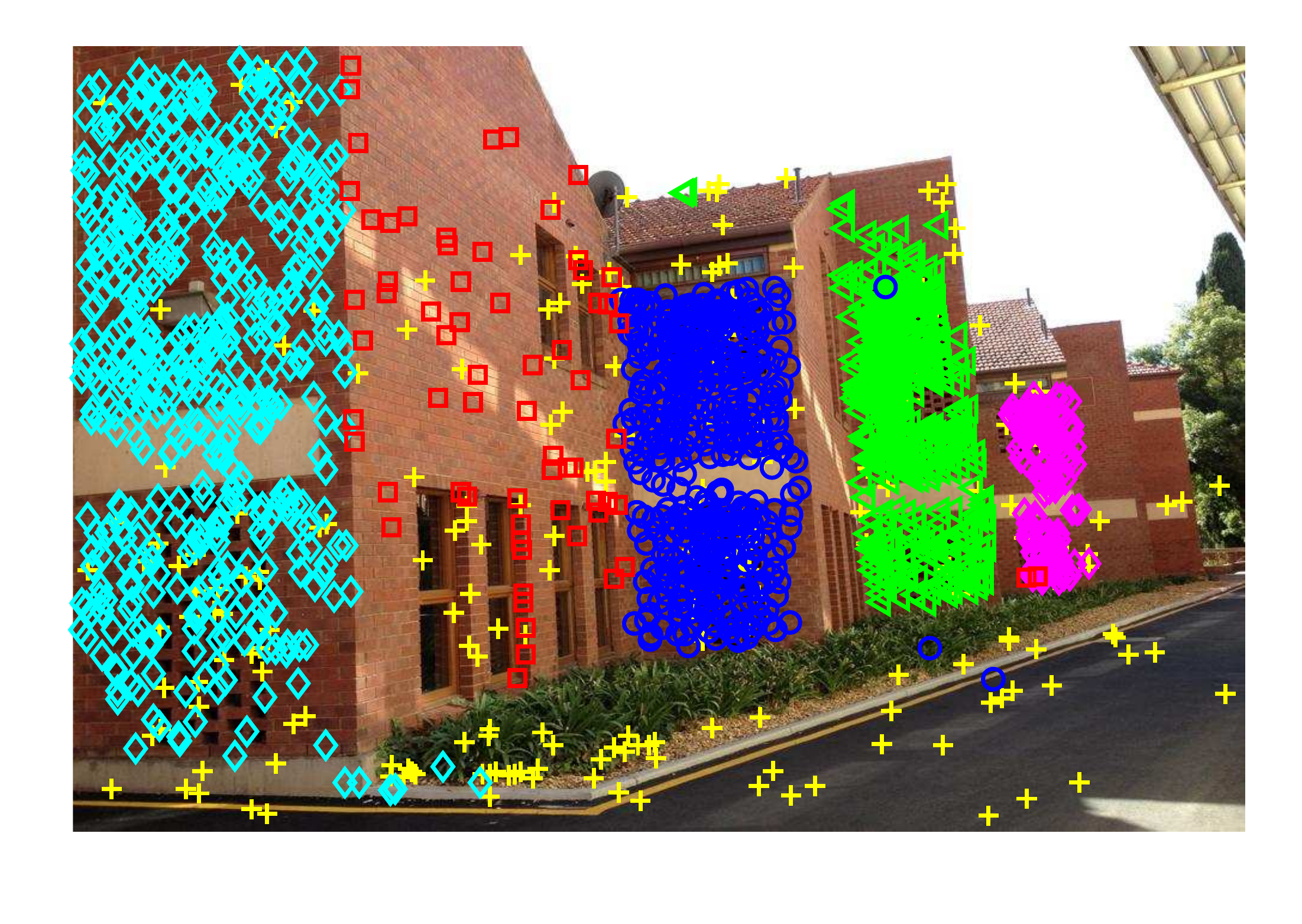}}
  \centerline{(a1) Unihouse}
\end{minipage}
\begin{minipage}{.23\textwidth}
\centerline{\includegraphics[width=1.06\textwidth]{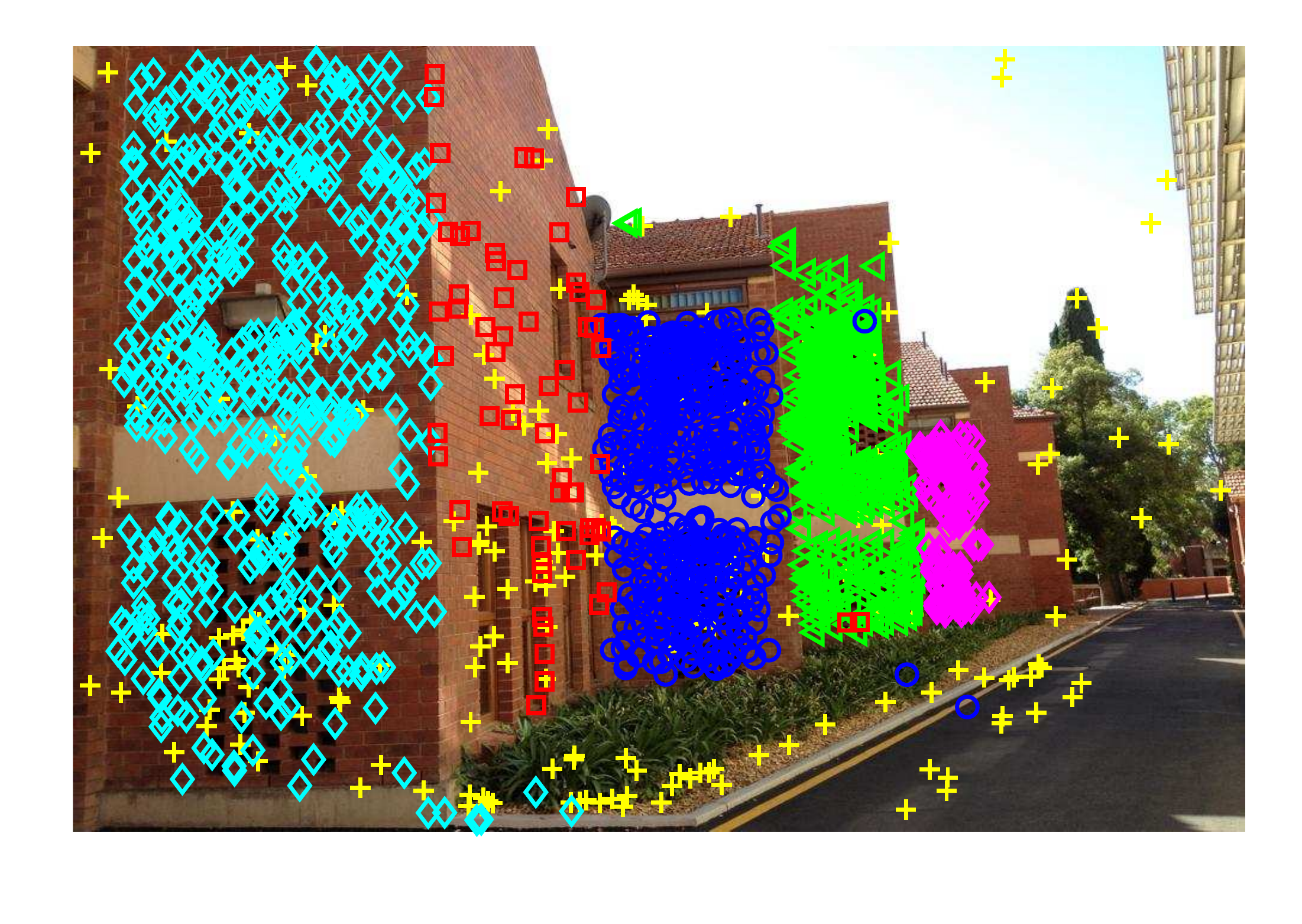}}
  \centerline{(a2) Unihouse}
\end{minipage}
}
\subfigure{
\begin{minipage}{.23\textwidth}
\centerline{\includegraphics[width=1.06\textwidth]{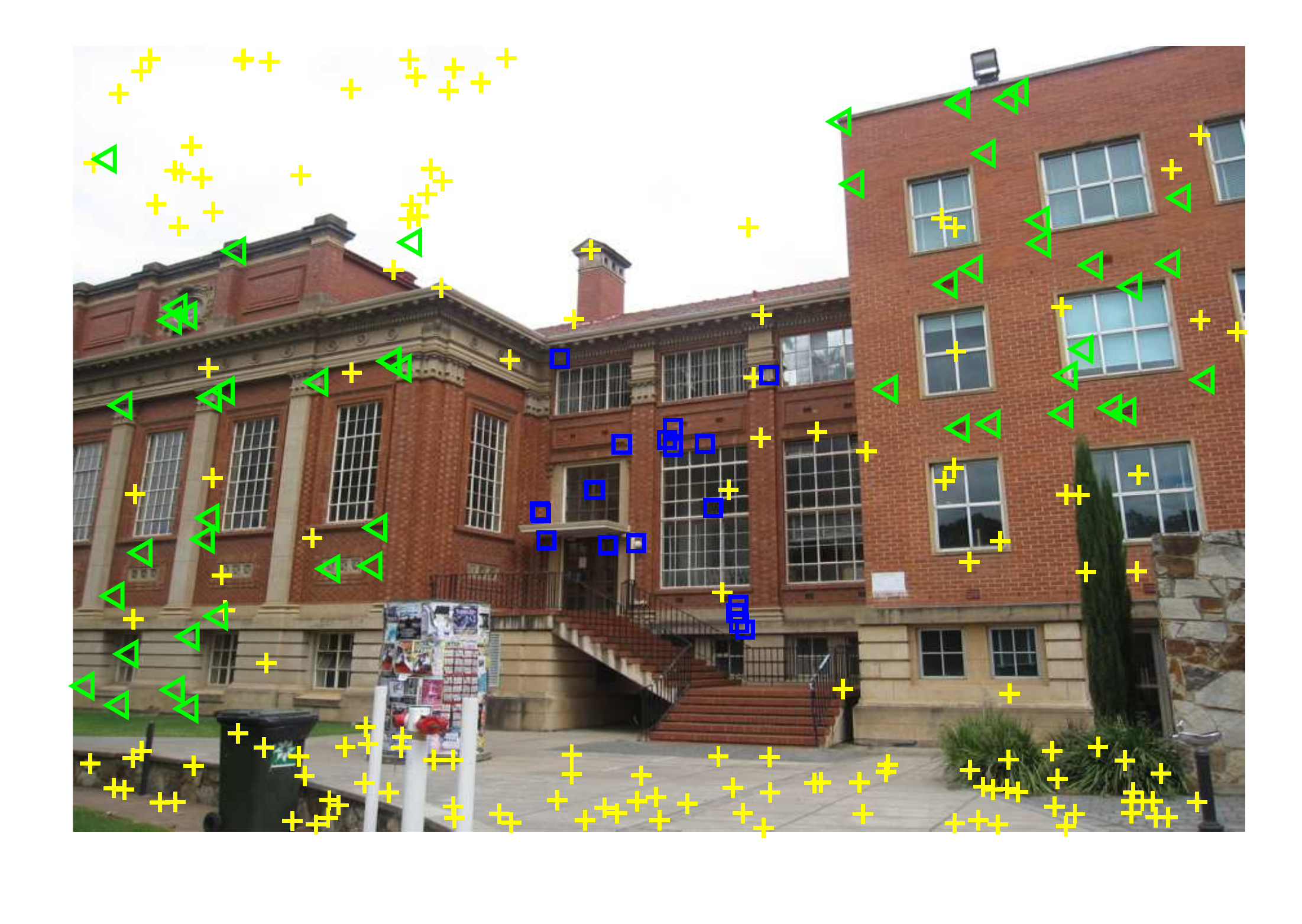}}
  \centerline{(b1) Barrsmith}
\end{minipage}
\begin{minipage}{.23\textwidth}
\centerline{\includegraphics[width=1.06\textwidth]{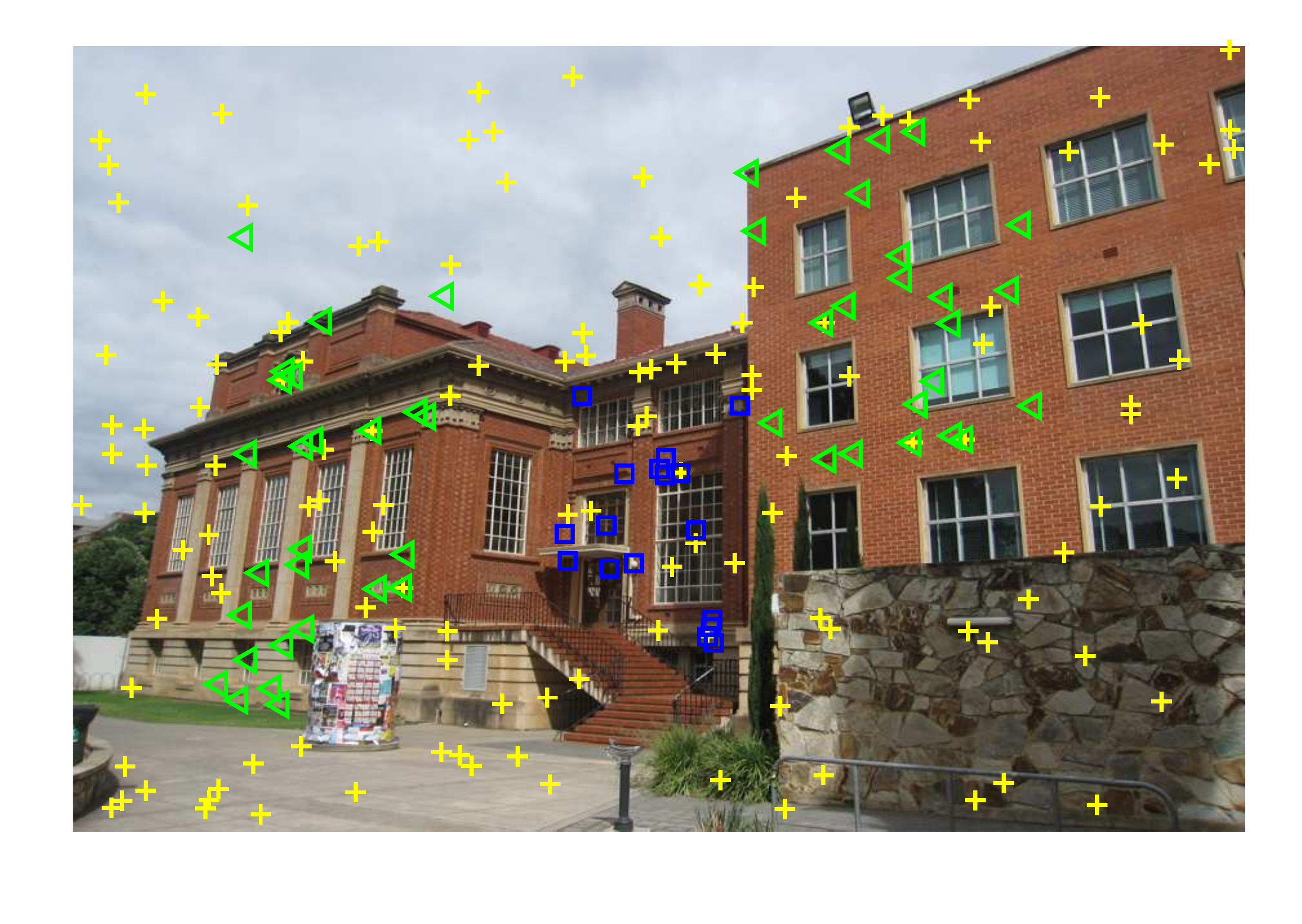}}
  \centerline{(b2) Barrsmith}
\end{minipage}
}
\caption{The fitting results obtained by SDF on two highly challenging image pairs. Each row shows the two views of each image pair. }
\label{fig:challengingdata}
\end{figure}

This paper presents an efficient and effective model fitting method (SDF), which achieves consistent and reliable results on two-view images. SDF requires very few user-adjusted parameters (i.e., the number of model instances, the number of superpixels and the support size ${\hat{n}}$ for Algorithm~\ref{alg:hypothesisupdating}). In practice, the fitting errors obtained by SDF does not change significantly with these three parameters within a wide range of values. During the experiments, we fix the number of superpixels to $100$$\sim$$200$, and fix ${\hat{n}}=10\%*n$, where $n$ is the number of input data points, for most image pairs.

It is worth pointing out that SDF can achieve very good performance on some highly challenging data, where most state-of-the-art fitting methods break down. As shown in Fig.~\ref{fig:challengingdata}, we test on the ``Unihouse" and ``Barrsmith" image pairs, and the fitting errors obtained by SDF is $1.14\%$ and $0.89\%$, respectively. 
The model instance (with red color in Fig.~\ref{fig:challengingdata}(a1) and \ref{fig:challengingdata}(a2)) on ``Unihouse" only includes $4.17\%$ inliers, and most or many methods would be challenged to detect. The inliers of the model instance (with green color in Fig.~\ref{fig:challengingdata}(b1) and \ref{fig:challengingdata}(b2)) on ``Barrsmith" have a wide spatial distribution and they are divided into two groups. Most state-of-the-art fitting methods usually estimate two model instances corresponding to the same model instance. In contrast, SDF succeeds in fitting all model instances on these two challenging data with very low fitting errors. The good fitting results are obtained mainly because of the three main components of SDF, i.e., the deterministic sampling algorithm is proposed to generate high-quality model hypotheses, the model hypothesis updating strategy is employed to further improve the quality of the generated model hypotheses and a novel model selection algorithm is developed to find all the model instances in {the data}.

\begin{figure}[t]
\centering
\subfigure{
\begin{minipage}{.24\textwidth}
\centerline{\includegraphics[width=1.06\textwidth]{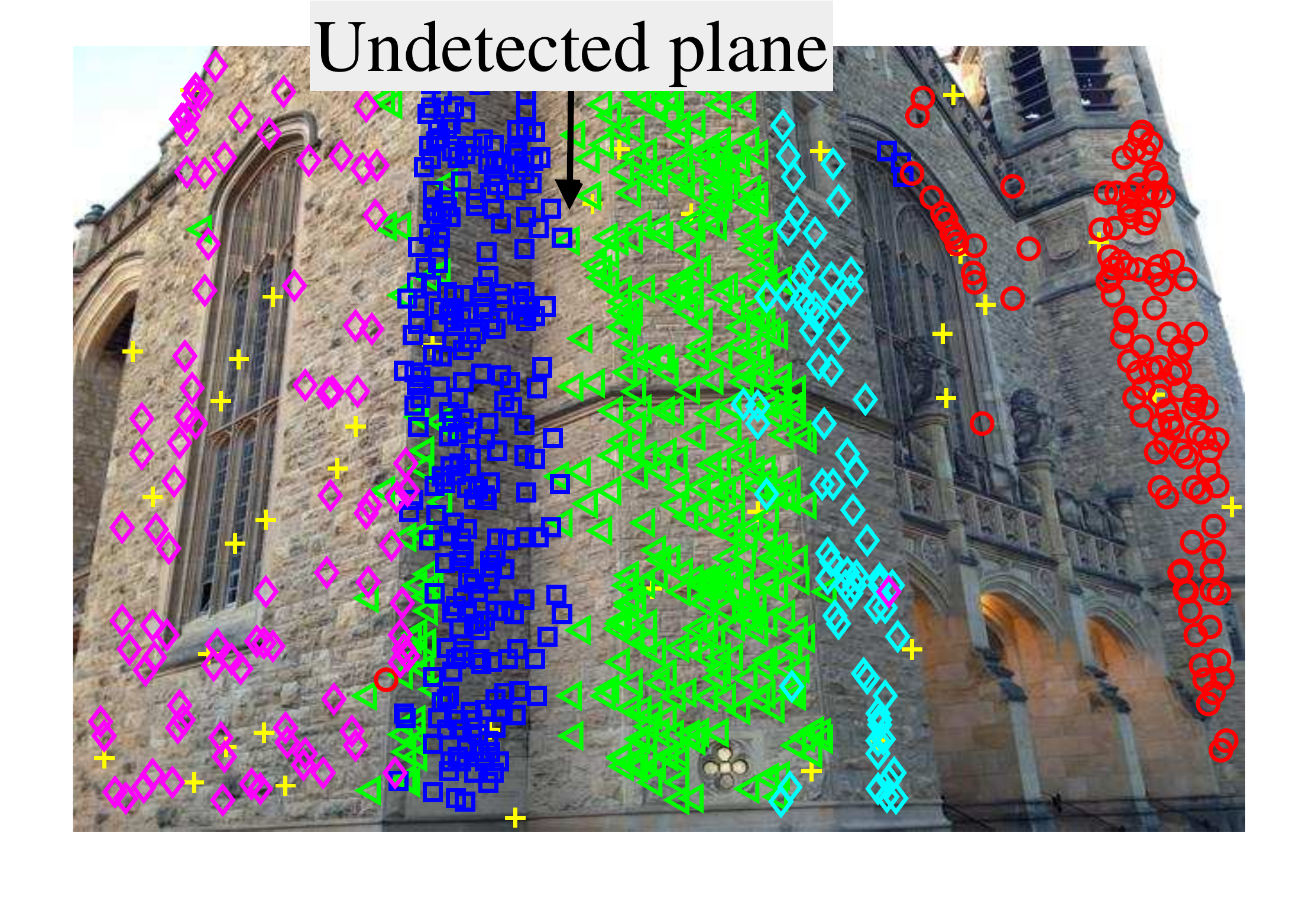}}
  \centerline{(a) }
\end{minipage}
\begin{minipage}{.24\textwidth}
\centerline{\includegraphics[width=1.06\textwidth]{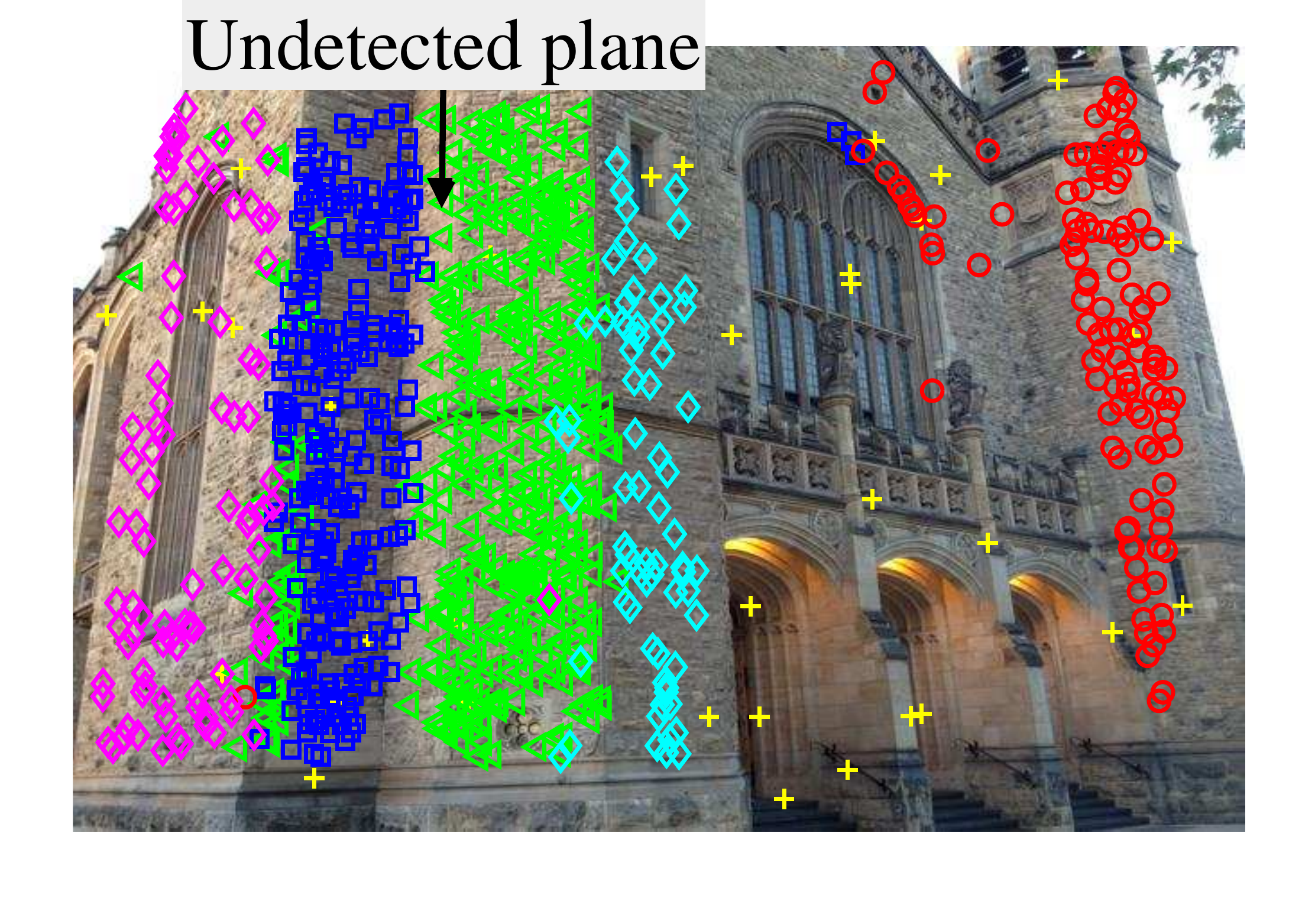}}
  \centerline{(b)}
\end{minipage}
}
\caption{An example showing that SDF fails to fit all the model instances. (a) and (b) show the two views of the ``Bonhall" image pairs, respectively. }
\label{fig:faildata}
\end{figure}
Of course, the proposed fitting method cannot successfully deal with all cases for the model fitting problem. For example, as shown in Fig.~\ref{fig:faildata}, SDF fails to estimate one model instance, and the undetected plane is long and narrow (pointed by an arrow). The model instance corresponding to the undetected plane only includes $6.20\%$ inliers, and {it is hard to generate significant model hypotheses by using both the proposed deterministic sampling algorithm and the other sampling algorithms}. However, this problem also affects most of the other state-of-the-art fitting methods.
\section{Conclusions}
\label{sec:conclusion}
We propose a novel deterministic fitting method (SDF) which utilizes the prior information of superpixels. In contrast to the general model fitting methods (e.g., RANSAC, T-linkage and RansaCov), the proposed method provides consistent and reliable solutions and it also significantly improves the computational efficiency and accuracy for most cases. Compared to existing deterministic fitting methods (e.g., BnB~\cite{li2009consensus} and Astar~\cite{chin2015efficient}), SDF is much faster and yields good solutions within a reasonable time. Moreover, SDF is able to effectively deal with both single-structure and multiple-structure data. Experimental results on a number of real images have demonstrated that SDF outperforms several other start-of-the-art fitting methods.

\begin{acknowledgements}
This work was supported by the National Natural Science Foundation of China under Grants U1605252, 61702431, 61472334 and 61571379, by the China Postdoctoral Science Foundation 2017M620272 and by the Fujian Province Education-Science Project for Middle-aged and Young Teachers JAT170024. This work was carried out when David Suter was with The University of Adelaide. This work was partially supported by ARC grant DP130102524.
\end{acknowledgements}

\bibliographystyle{spmpsci}      
\bibliography{ijcv2017}

\begin{thebibliography}{10}
\providecommand{\url}[1]{{#1}}
\providecommand{\urlprefix}{URL }
\expandafter\ifx\csname urlstyle\endcsname\relax
  \providecommand{\doi}[1]{DOI~\discretionary{}{}{}#1}\else
  \providecommand{\doi}{DOI~\discretionary{}{}{}\begingroup
  \urlstyle{rm}\Url}\fi

\bibitem{achanta2012slic}
Achanta, R., Shaji, A., Smith, K., Lucchi, A., Fua, P., Susstrunk, S.: Slic
  superpixels compared to state-of-the-art superpixel methods.
\newblock IEEE Trans. Pattern Anal. Mach. Intell. \textbf{34}(11), 2274--2282
  (2012)

\bibitem{brahmachari2013hop}
Brahmachari, A.S., Sarkar, S.: Hop-diffusion monte carlo for epipolar geometry
  estimation between very wide-baseline images.
\newblock IEEE Trans. Pattern Anal. Mach. Intell. \textbf{35}(3), 755--762
  (2013)

\bibitem{chin2015efficient}
Chin, T.J., Purkait, P., Eriksson, A., Suter, D.: Efficient globally optimal
  consensus maximisation with tree search.
\newblock IEEE Trans. Pattern Anal. Mach. Intell. \textbf{1}(1), 1--14 (2016)

\bibitem{chin2012accelerated}
Chin, T.J., Yu, J., Suter, D.: Accelerated hypothesis generation for
  multistructure data via preference analysis.
\newblock IEEE Trans. Pattern Anal. Mach. Intell. \textbf{34}(4), 625--638
  (2012)

\bibitem{chum2005matching}
Chum, O., Matas, J.: Matching with prosac-progressive sample consensus.
\newblock In: Proc. IEEE Conf. Comput. Vis. Pattern Recog., pp. 220--226 (2005)

\bibitem{chum2003locally}
Chum, O., Matas, J., Kittler, J.: Locally optimized ransac.
\newblock Pattern Recog. \textbf{25}(1), 236--243 (2003)

\bibitem{Enqvist2015}
Enqvist, O., Ask, E., Kahl, F., {\AA}str{\"o}m, K.: Tractable algorithms for
  robust model estimation.
\newblock Int. J. Comput. Vis. \textbf{112}(1), 115--129 (2015)

\bibitem{fischler1981random}
Fischler, M.A., Bolles, R.C.: Random sample consensus: a paradigm for model
  fitting with applications to image analysis and automated cartography.
\newblock Comm. ACM \textbf{24}(6), 381--395 (1981)

\bibitem{Fragoso_2013_ICCV}
Fragoso, V., Sen, P., Rodriguez, S., Turk, M.: Evsac: Accelerating hypotheses
  generation by modeling matching scores with extreme value theory.
\newblock In: Proc. IEEE Int. Conf. Comput. Vis., pp. 2472--2479 (2013)

\bibitem{Fragoso_2013_CVPR}
Fragoso, V., Turk, M.: Swigs: A swift guided sampling method.
\newblock In: Proc. IEEE Conf. Comput. Vis. Pattern Recog., pp. 2770--2777
  (2013)

\bibitem{fredriksson2015practical}
Fredriksson, J., Larsson, V., Olsson, C.: Practical robust two-view translation
  estimation.
\newblock In: Proc. IEEE Conf. Comput. Vis. Pattern Recog., pp. 2684--2690
  (2015)

\bibitem{hart1968formal}
Hart, P.E., Nilsson, N.J., Raphael, B.: A formal basis for the heuristic
  determination of minimum cost paths.
\newblock IEEE Trans. Syst., Man, Cybern. B, Cybern. \textbf{4}(2), 100--107
  (1968)

\bibitem{isack2014energy}
Isack, H., Boykov, Y.: Energy based multi-model fitting \& matching for 3d
  reconstruction.
\newblock In: Proc. IEEE Conf. Comput. Vis. Pattern Recog., pp. 1146--1153
  (2014)

\bibitem{kanazawa2004detection}
Kanazawa, Y., Kawakami, H.: Detection of planar regions with uncalibrated
  stereo using distributions of feature points.
\newblock In: Proc. Bri. Mach. Vis. Conf., pp. 247--256 (2004)

\bibitem{Lai2017152}
Lai, T., Wang, H., Yan, Y., Xiao, G., Suter, D.: Efficient guided hypothesis
  generation for multi-structure epipolar geometry estimation.
\newblock Comput. Vis. Image Und. \textbf{154}, 152 -- 165 (2017)

\bibitem{lee2013deterministic}
Lee, K.H., Lee, S.W.: Deterministic fitting of multiple structures using
  iterative maxfs with inlier scale estimation.
\newblock In: Proc. IEEE Int. Conf. Comput. Vis., pp. 41--48 (2013)

\bibitem{li2009consensus}
Li, H.: Consensus set maximization with guaranteed global optimality for robust
  geometry estimation.
\newblock In: Proc. IEEE Int. Conf. Comput. Vis., pp. 1074--1080 (2009)

\bibitem{litman2015inverting}
Litman, R., Korman, S., Bronstein, A., Avidan, S.: Inverting ransac: Global
  model detection via inlier rate estimation.
\newblock In: Proc. IEEE Conf. Comput. Vis. Pattern Recog., pp. 5243--5251
  (2015)

\bibitem{lowe2004distinctive}
Lowe, D.G.: Distinctive image features from scale-invariant keypoints.
\newblock Int. J. Comput. Vis. \textbf{60}(2), 91--110 (2004)

\bibitem{Magri_2014_CVPR}
Magri, L., Fusiello, A.: T-linkage: A continuous relaxation of j-linkage for
  multi-model fitting.
\newblock In: Proc. IEEE Conf. Comput. Vis. Pattern Recog., pp. 3954--3961
  (2014)

\bibitem{Magri_2016_CVPR}
Magri, L., Fusiello, A.: Multiple model fitting as a set coverage problem.
\newblock In: Proc. IEEE Conf. Comput. Vis. Pattern Recog., pp. 3318--3326
  (2016)

\bibitem{mittal2012generalized}
Mittal, S., Anand, S., Meer, P.: Generalized projection-based m-estimator.
\newblock IEEE Trans. Pattern Anal. Mach. Intell. \textbf{34}(12), 2351--2364
  (2012)

\bibitem{pham2014interacting}
Pham, T.T., Chin, T.J., Schindler, K., Suter, D.: Interacting geometric priors
  for robust multimodel fitting.
\newblock IEEE Trans. Image Process. \textbf{23}(10), 4601--4610 (2014)

\bibitem{Poling2014}
Poling, B., Lerman, G.: A new approach to two-view motion segmentation using
  global dimension minimization.
\newblock Int. J. Comput. Vis. \textbf{108}(3), 165--185 (2014)

\bibitem{serradell2010combining}
Serradell, E., {\"O}zuysal, M., Lepetit, V., Fua, P., Moreno-Noguer, F.:
  Combining geometric and appearance priors for robust homography estimation.
\newblock In: Proc. Eur. Conf. Comput. Vis., pp. 58--72 (2010)

\bibitem{shen2014lazy}
Shen, J., Du, Y., Wang, W., Li, X.: Lazy random walks for superpixel
  segmentation.
\newblock IEEE Trans. Image Process. \textbf{23}(4), 1451--1462 (2014)

\bibitem{tennakoon2016robust}
Tennakoon, R., Bab-Hadiashar, A., Cao, Z., Hoseinnezhad, R., Suter, D.: Robust
  model fitting using higher than minimal subset sampling.
\newblock IEEE Trans. Pattern Anal. Mach. Intell. \textbf{38}(2), 350--361
  (2016)

\bibitem{Tordoff2005}
Tordoff, B.J., Murray, D.W.: Guided-mlesac: faster image transform estimation
  by using matching priors.
\newblock IEEE Trans. Pattern Anal. Mach. Intell. \textbf{27}(10), 1523--1535
  (2005)

\bibitem{Tran2014}
Tran, Q.H., Chin, T.J., Chojnacki, W., Suter, D.: Sampling minimal subsets with
  large spans for robust estimation.
\newblock Int. J. Comput. Vis. \textbf{106}(1), 93--112 (2014)

\bibitem{wand1994kernel}
Wand, M., Jones, M.: Kernel Smoothing.
\newblock Crc Press (1994)

\bibitem{wang2012simultaneously}
Wang, H., Chin, T.J., Suter, D.: Simultaneously fitting and segmenting
  multiple-structure data with outliers.
\newblock IEEE Trans. Pattern Anal. Mach. Intell. \textbf{34}(6), 1177--1192
  (2012)

\bibitem{Wang_2015_ICCV}
Wang, H., Xiao, G., Yan, Y., Suter, D.: Mode-seeking on hypergraphs for robust
  geometric model fitting.
\newblock In: Proc. IEEE Int. Conf. Comput. Vis., pp. 2902--2910 (2015)

\bibitem{wong2011dynamic}
Wong, H.S., Chin, T.J., Yu, J., Suter, D.: Dynamic and hierarchical
  multi-structure geometric model fitting.
\newblock In: Proc. IEEE Int. Conf. Comput. Vis., pp. 1044--1051 (2011)

\bibitem{Woodford2014}
Woodford, O.J., Pham, M.T., Maki, A., Perbet, F., Stenger, B.: Demisting the
  hough transform for 3d shape recognition and registration.
\newblock Int. J. Comput. Vis. \textbf{106}(3), 332--341 (2014)

\bibitem{xiao2016HF}
Xiao, G., Wang, H., Lai, T., Suter, D.: Hypergraph modelling for geometric
  model fitting.
\newblock Pattern Recog. \textbf{60}(1), 748--760 (2016)

\bibitem{ECCVXiao2016}
Xiao, G., Wang, H., Yan, Y., Suter, D.: Superpixel-based two-view deterministic
  fitting for multiple-structure data.
\newblock In: Proc. Eur. Conf. Comput. Vis., pp. 517--533 (2016)

\end{thebibliography}

\end{document}